\documentclass[10pt,twocolumn,letterpaper]{article}

\usepackage{cvpr}
\usepackage{times}
\usepackage{epsfig}
\usepackage{graphicx}
\usepackage{amsmath}
\usepackage{amssymb}
\usepackage{caption} %
\usepackage{soul} %
\usepackage{algorithmic} %
\usepackage{algorithm} %
\usepackage{ifthen} %
\usepackage[T1]{fontenc} %
\usepackage{microtype}
\usepackage{xcolor}
\renewcommand\fbox{\fcolorbox{red}{white}}
\usepackage{array}
\usepackage{arydshln} %
\usepackage{multirow,array,rotating}

\usepackage{booktabs}
\newcommand{\ra}[1]{\renewcommand{\arraystretch}{#1}}

\newbox\jsavebox
\newcommand{\jsubfig}[2]{%
	\sbox\jsavebox{#1}%
	\parbox[t]{\wd\jsavebox}{\centering\usebox\jsavebox\\#2}%
	}

\DeclareMathOperator*{\argmin}{arg\,min}

\newcommand{\mbf}{\mathbf}

\newcommand{\SDF}{\mathtt{SDF}}
\newcommand{\muj}{\pmb{\mu}_j}
\newcommand{\sigmaj}{\pmb{\sigma}_j}

\usepackage[pagebackref=true,breaklinks=true,letterpaper=true,colorlinks,bookmarks=false]{hyperref}

\cvprfinalcopy %

\def\cvprPaperID{418} %

\ifcvprfinal\fi
\begin{document}

\title{DualSDF: Semantic Shape Manipulation using a Two-Level Representation}

\author{Zekun Hao, Hadar Averbuch-Elor, Noah Snavely, Serge Belongie \\
Cornell Tech, Cornell University
}
\twocolumn[{%
\renewcommand\twocolumn[1][]{#1}%
\maketitle
\thispagestyle{empty}
\begin{center}
\vspace{-4pt}
  \centering
  \setlength{\fboxrule}{2.0pt}
\includegraphics[trim={2.9cm 3.6cm 3.3cm 3.5cm},clip,width=0.135\textwidth]{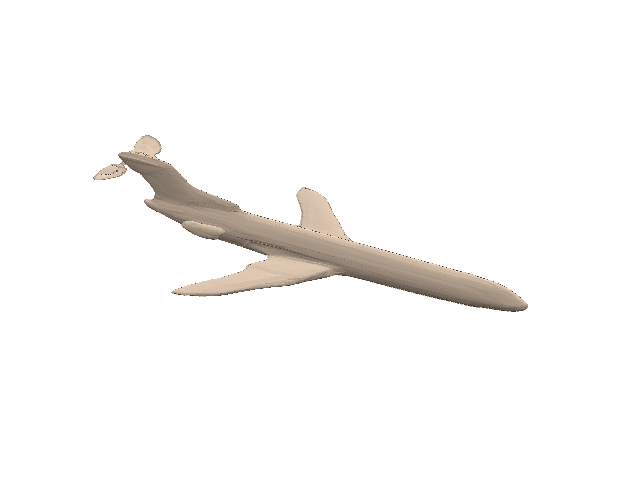}
\includegraphics[trim={2.9cm 3.6cm 3.3cm 3.5cm},clip,width=0.135\textwidth]{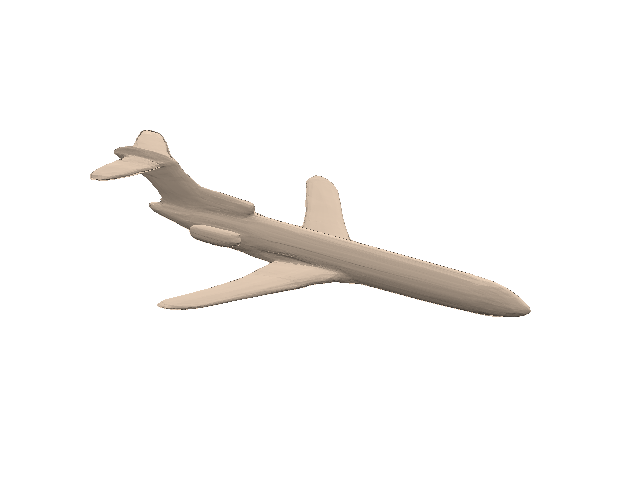}
\includegraphics[trim={2.9cm 3.6cm 3.3cm 3.5cm},clip,width=0.135\textwidth]{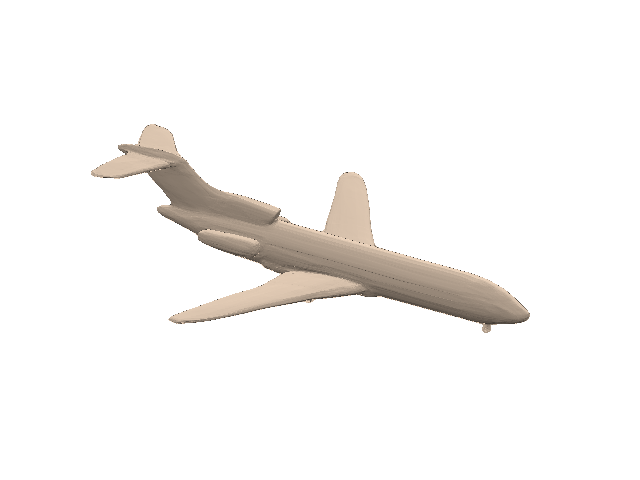}
\fbox{{\includegraphics[trim={2.9cm 3.6cm 3.3cm 3.5cm},clip,width=0.135\textwidth]{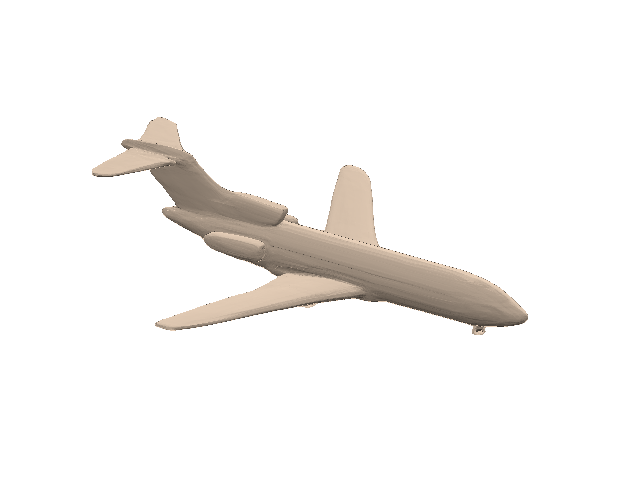}}}
\includegraphics[trim={2.9cm 3.6cm 3.3cm 3.5cm},clip,width=0.14\textwidth]{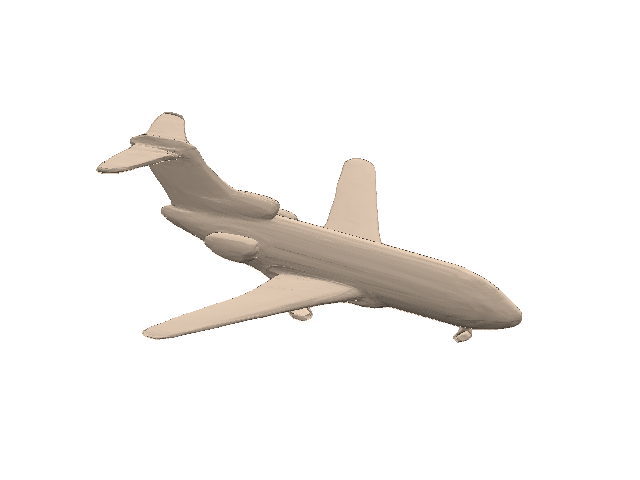}
\includegraphics[trim={2.9cm 3.6cm 3.3cm 3.5cm},clip,width=0.135\textwidth]{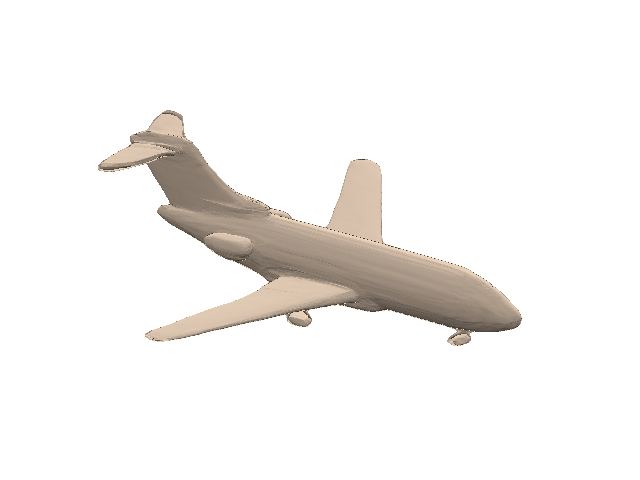}
\includegraphics[trim={2.9cm 3.6cm 3.3cm 3.5cm},clip,width=0.135\textwidth]{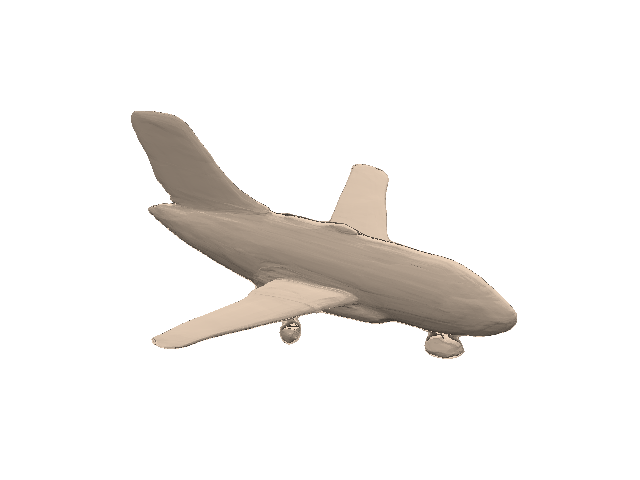} \\
\includegraphics[trim={2.9cm 3.6cm 3.3cm 3.5cm},clip,width=0.135\textwidth]{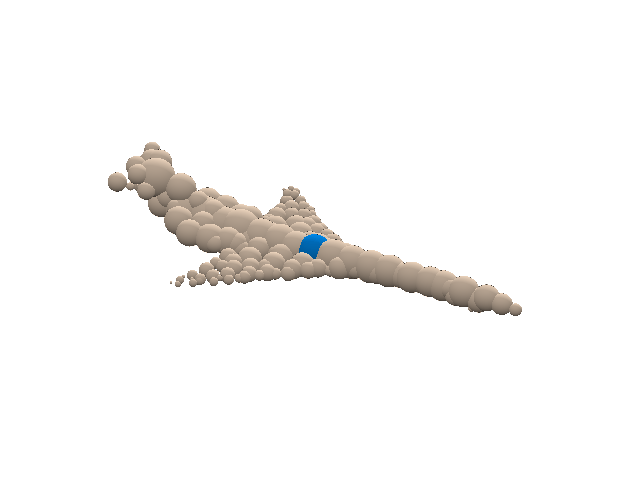}
\includegraphics[trim={2.9cm 3.6cm 3.3cm 3.5cm},clip,width=0.135\textwidth]{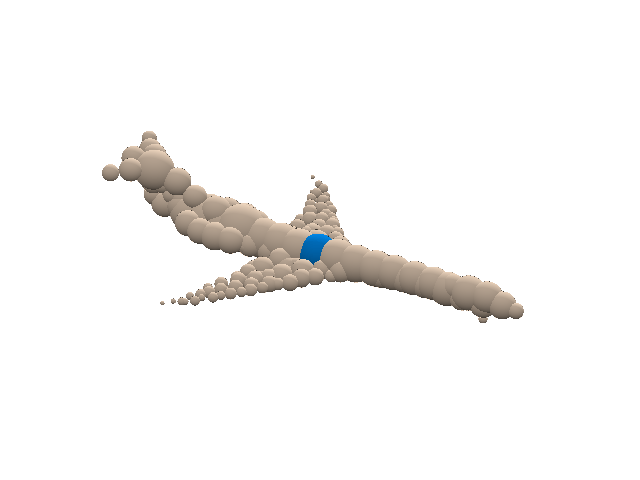}
\includegraphics[trim={2.9cm 3.6cm 3.3cm 3.5cm},clip,width=0.135\textwidth]{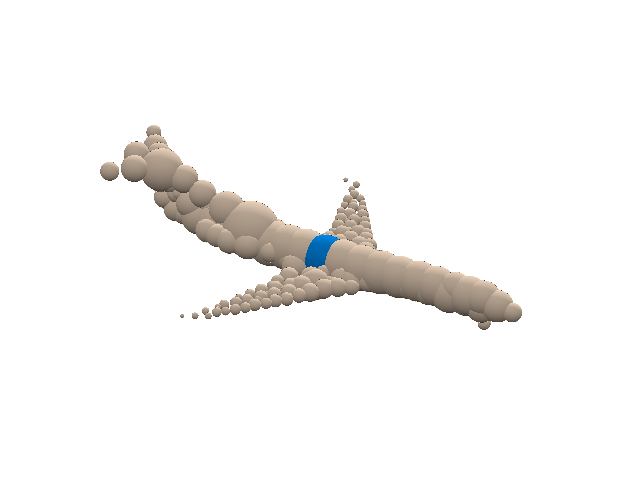}
\fbox{{\includegraphics[trim={2.9cm 3.6cm 3.3cm 3.5cm},clip,width=0.135\textwidth]{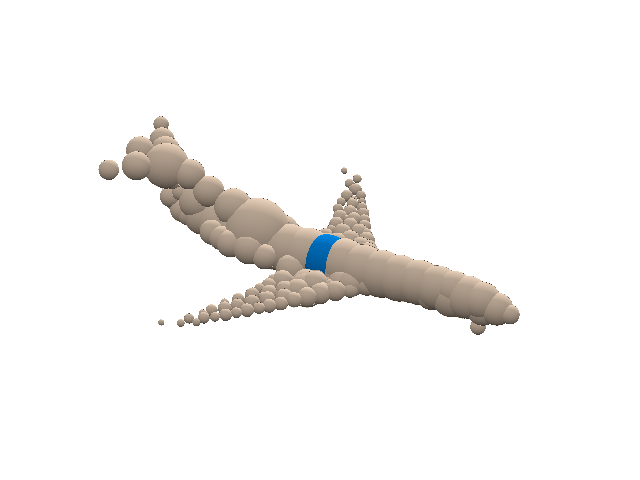}}}
\includegraphics[trim={2.9cm 3.6cm 3.3cm 3.5cm},clip,width=0.14\textwidth]{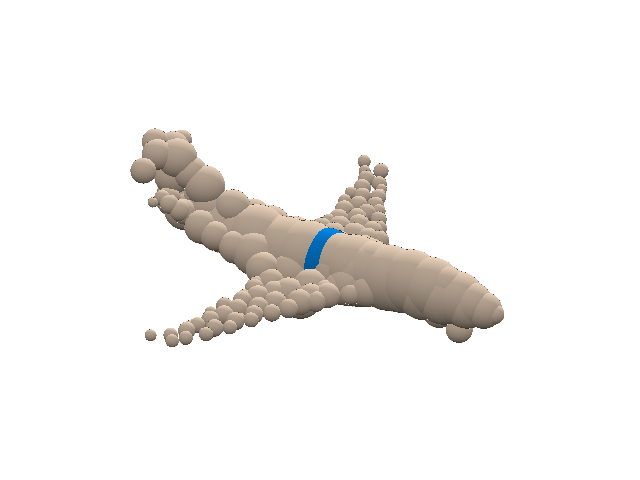}
\includegraphics[trim={2.9cm 3.6cm 3.3cm 3.5cm},clip,width=0.135\textwidth]{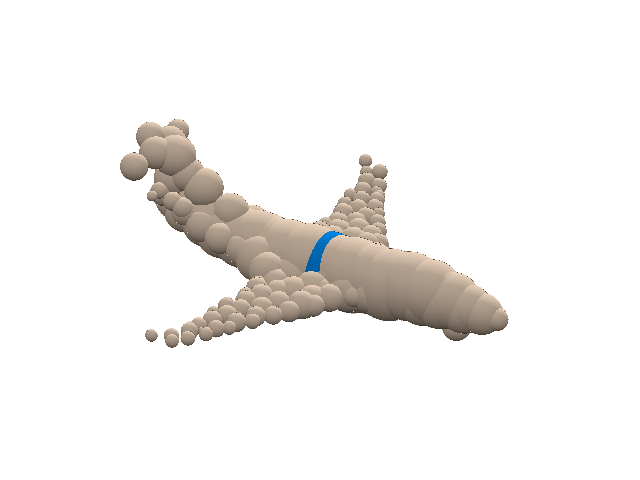}
\includegraphics[trim={2.9cm 3.6cm 3.3cm 3.5cm},clip,width=0.135\textwidth]{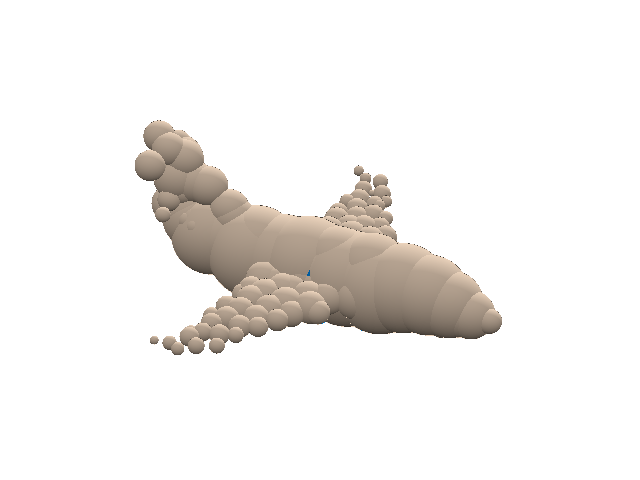}  \\ %
\includegraphics[trim={2.3cm 3.6cm 2.7cm 3.5cm},clip,width=0.135\textwidth]{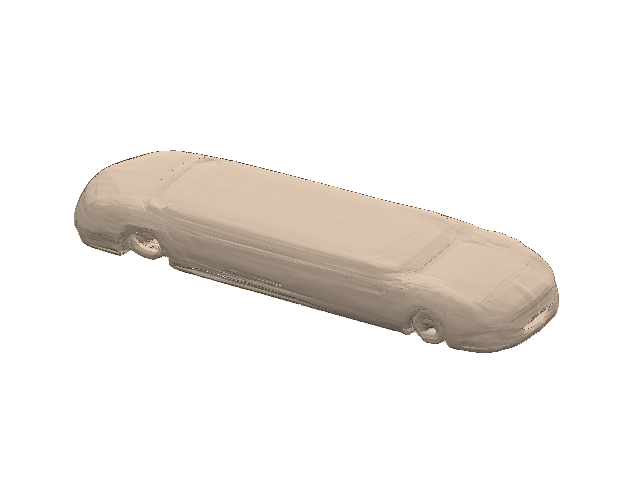}
\includegraphics[trim={2.3cm 3.6cm 2.7cm 3.5cm},clip,width=0.135\textwidth]{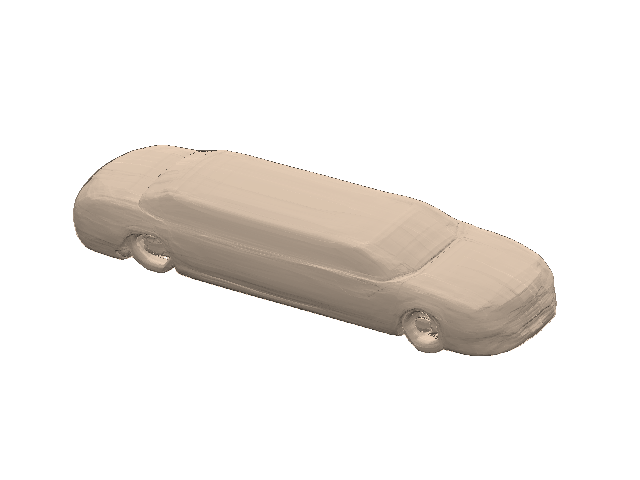}
\includegraphics[trim={2.3cm 3.6cm 2.7cm 3.5cm},clip,width=0.135\textwidth]{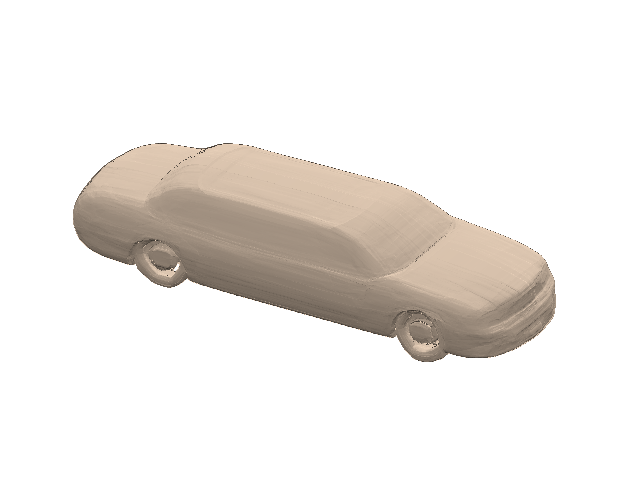}
\fbox{{\includegraphics[trim={2.3cm 3.6cm 2.7cm 3.5cm},clip,width=0.135\textwidth]{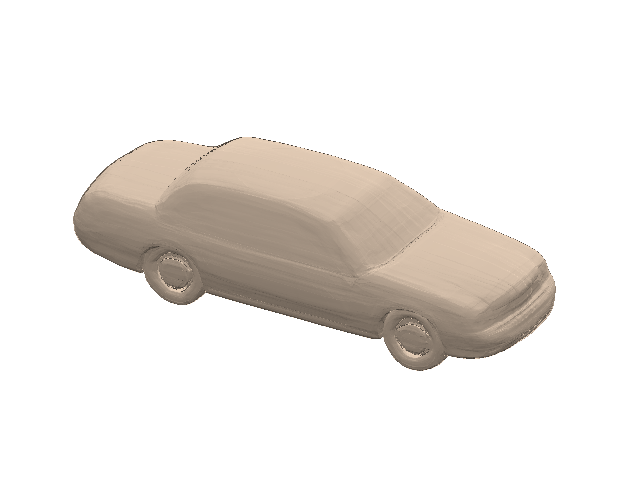}}}
\includegraphics[trim={2.3cm 3.6cm 2.7cm 3.5cm},clip,width=0.14\textwidth]{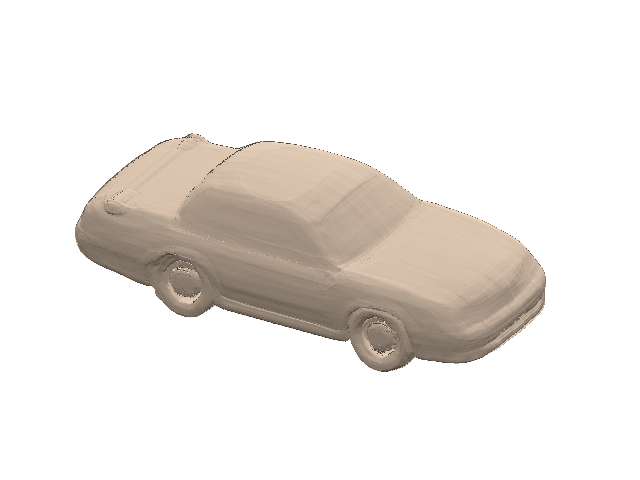}
\includegraphics[trim={2.3cm 3.6cm 2.7cm 3.5cm},clip,width=0.135\textwidth]{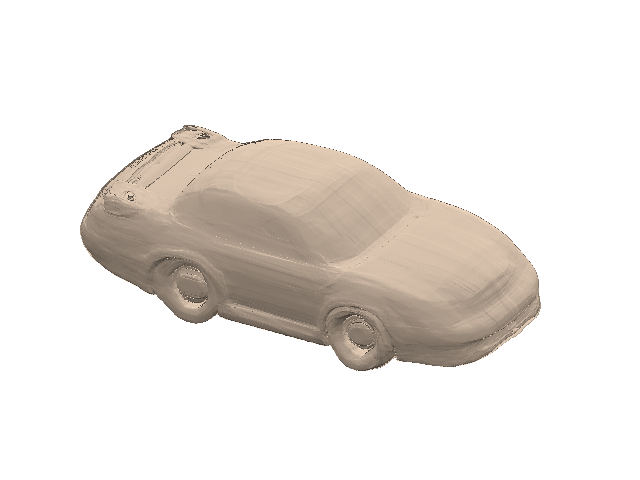}
\includegraphics[trim={2.3cm 3.6cm 2.7cm 3.5cm},clip,width=0.135\textwidth]{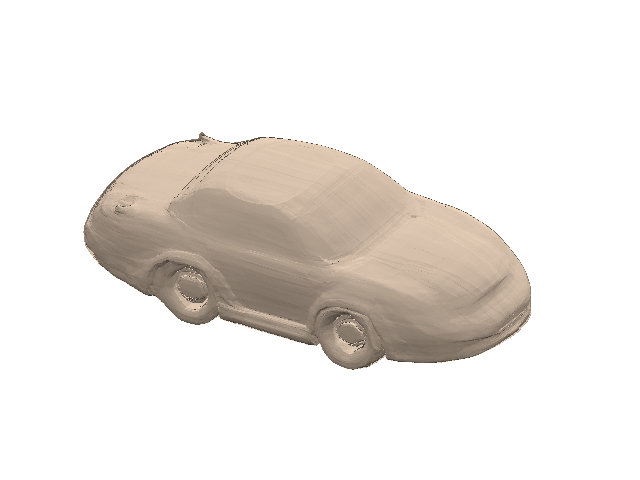} \\
\includegraphics[trim={2.3cm 3.6cm 2.7cm 3.5cm},clip,width=0.135\textwidth]{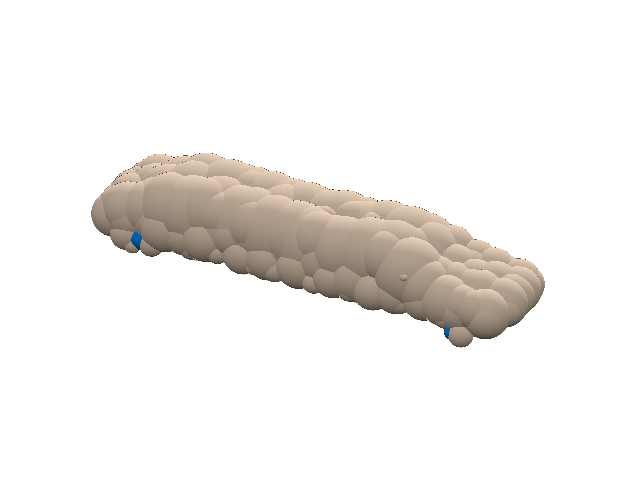}
\includegraphics[trim={2.3cm 3.6cm 2.7cm 3.5cm},clip,width=0.135\textwidth]{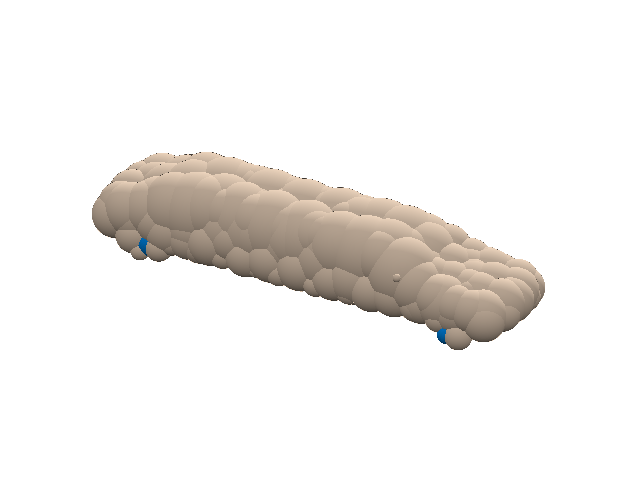}
\includegraphics[trim={2.3cm 3.6cm 2.7cm 3.5cm},clip,width=0.135\textwidth]{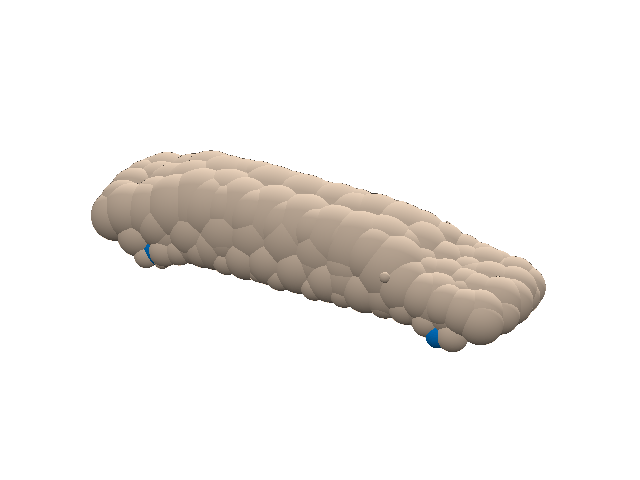}
\fbox{{\includegraphics[trim={2.3cm 3.6cm 2.7cm 3.5cm},clip,width=0.135\textwidth]{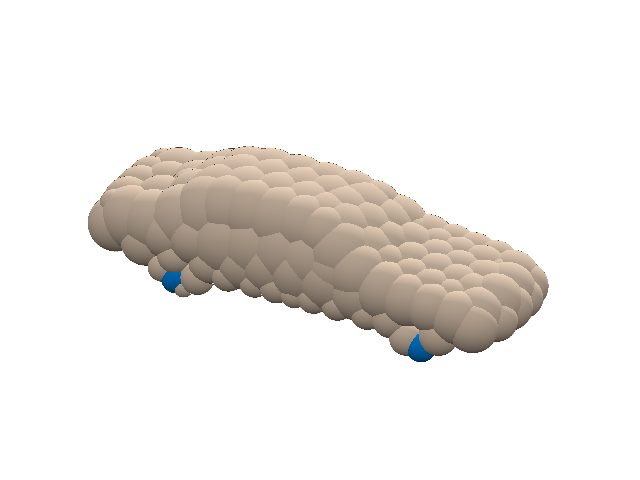}}}
\includegraphics[trim={2.3cm 3.6cm 2.7cm 3.5cm},clip,width=0.14\textwidth]{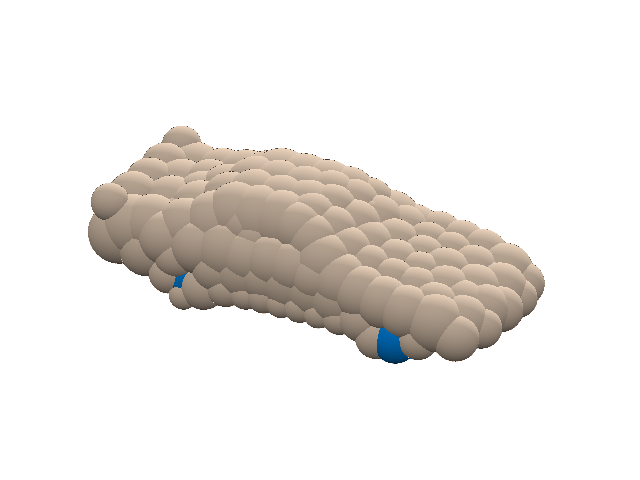}
\includegraphics[trim={2.3cm 3.6cm 2.7cm 3.5cm},clip,width=0.135\textwidth]{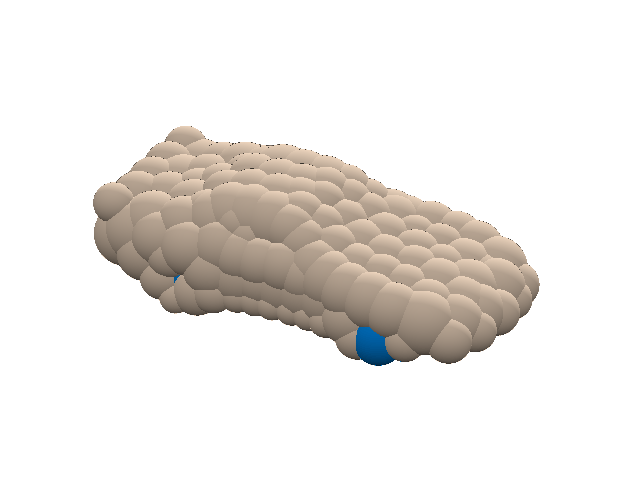}
\includegraphics[trim={2.3cm 3.6cm 2.7cm 3.5cm},clip,width=0.135\textwidth]{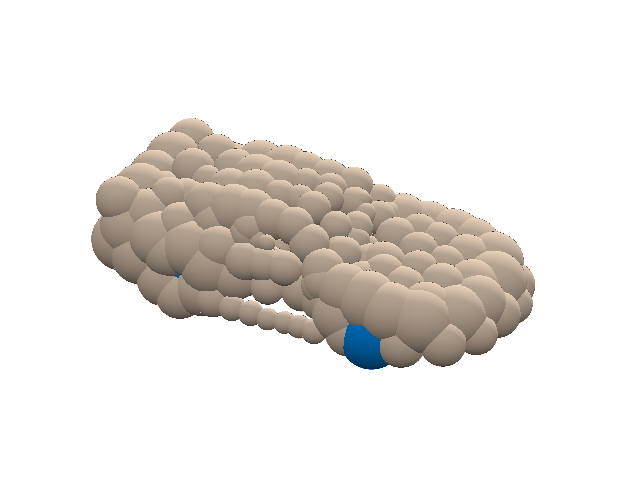} \\
\vspace{-4pt}
  \captionof{figure}{DualSDF represents shapes using two levels of granularity, allowing users to manipulate high resolution shapes (odd rows) with high-level concepts through manipulating a proxy primitive-based shape (even rows). Simple editing operations on individual primitives (colored in blue) are propagated to the other primitives and the fine-grained model in a semantically meaningful manner. Above, we illustrate how an existing shape (inside the red box) can be modified semantically by adjusting the radius of a primitive (fuselage diameter on the airplane) or the distance between two primitives (wheelbase of a car).
  }
  \vspace{10pt}
  \label{fig:teaser}
\end{center}
}]
  
\begin{abstract}
We are seeing a Cambrian explosion of 3D shape representations for use in machine learning. 
Some representations seek high expressive power in capturing high-resolution detail. Other approaches seek to represent shapes as compositions of simple parts, which are intuitive for people to understand and easy to edit and manipulate. However, it is difficult to achieve both fidelity and interpretability in the same representation. We propose DualSDF, a representation expressing shapes at two levels of granularity, one capturing fine details and the other representing an abstracted proxy shape using simple and semantically consistent shape primitives. 
To achieve a tight coupling between the two representations, we use a variational objective over a shared latent space.
Our two-level model gives rise to a new shape manipulation technique in which a user can interactively manipulate
the coarse proxy shape and see the changes instantly
mirrored in the high-resolution shape. Moreover, our model actively augments and 
guides the manipulation towards producing semantically meaningful shapes, making complex manipulations possible with minimal user input.

\end{abstract}

\section{Introduction}

There has been increasing interest in leveraging the power of neural networks to learn expressive shape representations for high-fidelity generative 3D modeling \cite{ben2018multi,groueix2018atlasnet,yang2019pointflow,park2019deepsdf,mescheder2019occupancy}. 
At the same time, other research has explored parsimonious representations of shape as compositions of primitives \cite{tulsiani2017learning,deprelle2019learning} or other 
simple, abstracted elements \cite{genova2019learning,deng2019cvxnets}. Such shape decompositions are more intuitive than a global, high-dimensional representation, and 
more suitable for tasks such as shape editing.
Unfortunately, it is difficult to achieve both fidelity and interpretability in a single representation.

\begin{figure}
    \centering
    \includegraphics[width = \columnwidth, trim=0 0.6cm 2.9cm 6.7cm,clip]{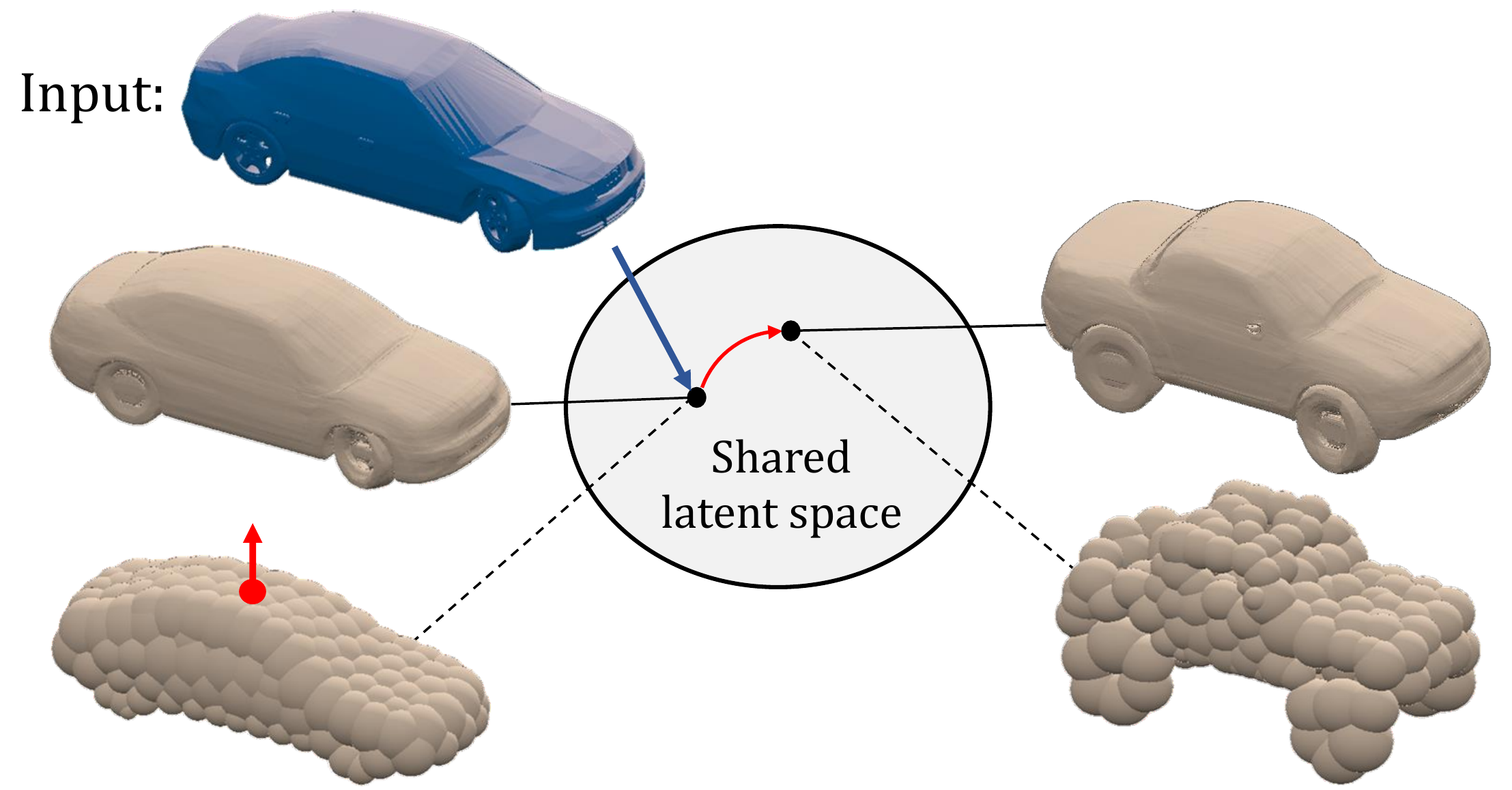}
    \caption{Our technique learns a shared latent space for an input collection of shapes, represented as meshes. From this joint space, shapes can be expressed using two levels of granularity. Shapes can be manipulated via the coarse 3D proxy shape (marked with a dotted line). %
    The figure illustrates how moving a primitive (red arrow on car) will propagate to changes to the latent code (red arrow in the latent space) -- in this case, leading to a taller car where the other parts of the car adapt accordingly.
    }
    \label{fig:JointLatent}
\end{figure}

In this work, we propose a generative \emph{two-level} model that simultaneously represents 3D shapes using two levels of granularity, one for capturing 
fine-grained detail and the other encoding a coarse structural decomposition. The two levels are tightly coupled via a shared latent space, wherein a single latent code vector decodes to two representations of the same underlying shape. An appealing consequence is that modifications to one representation can be readily propagated to the other via the shared code (as shown in Figure \ref{fig:teaser} and Figure \ref{fig:JointLatent}). 

The shared latent space is learned with a variational auto-decoder (VAD) \cite{zadeh2019variational}. This approach not only imposes a Gaussian prior on the latent space, which enables sampling,
but also encourages a compact latent space suitable for interpolation and optimization-based manipulation. Furthermore, as we empirically demonstrate, compared to an auto-encoder or auto-decoder, our model enforces a tighter coupling between different representations, even for novel shapes. %

Another key insight is that implicit surface representations, particularly signed distance fields (SDFs) \cite{park2019deepsdf,mescheder2019occupancy,chen2019learning}, are an effective substrate for both levels of granularity. Our coarse-level representation is based on the union of simple primitives, which yield efficient SDF formulations. Our fine-scale model represents SDFs with deep networks %
and is capable of capturing  high-resolution detail~\cite{park2019deepsdf}.
In addition to other desirable properties of implicit shape formulations, expressing both representations under a unified framework allows for simpler implementation and evaluation.

We show that our two-level approach offers the benefits of simplicity and interpretability without compromising fidelity.
We demonstrate our approach through a novel shape manipulation application, where a shape can be manipulated in the proxy primitive-based representation by editing individual primitives. These editions are simultaneously reflected to the high-resolution shape in a semantically meaningful way via the shared latent code. 
Moreover, minimal user input is needed to achieve complex shape manipulation. Under our optimization-based manipulation scheme, 
sparse edits on a subset of primitives can be propagated to the rest of the primitives while maintaining the shape on the manifold of likely shapes. %
Such an approach to manipulation is much more intuitive than a direct editing of the high-resolution mesh using deformation tools. A user can simply drag individual primitives in 3D to edit the shape (e.g. Figure \ref{fig:JointLatent}) while observing the rest of the primitives and the high resolution shape change accordingly at an interactive rate.

Last, we introduce two novel metrics for evaluating the manipulation performance of our model: \textit{cross-representation consistency} and \textit{primitive-based semantic consistency}. These metrics provide insights on how well the two representations agree with each other as well as how consistent the primitives are across different shapes. %
Code is available at \url{https://github.com/zekunhao1995/DualSDF}.

\section{Related Work}

\noindent \textbf{Generative 3D modeling.}
Prior to the Deep Learning era, 3D modeling of a shape collection was typically performed on a mesh representation. %
Many methods focus specifically on human models \cite{anguelov2005scape,pons2015dyna, gao2019sparse}, and aim at modeling deformations of a template model. The main limitation of most mesh-based representations, modern ones included, is that they are limited to meshes sharing the same connectivity \cite{litany2018deformable,tan2018variational}. Recently, Gao et al.~\cite{gao2019sdm} proposed a technique to generate structured deformable meshes of a shape collection, which overcomes the same-connectivity constraint. However, part annotations are needed for training their model.

Parametric surface representations are another popular modeling approach. In AtlasNet~\cite{groueix2018atlasnet}, shapes are represented using multiple surfaces parameterized by neural networks. Williams et al.~\cite{williams2019deep} use multiple charts to generate high-fidelity point cloud reconstructions in the absence of training data. Ben-Hamu et al.~\cite{ben2018multi} integrate a multi-chart representation into a GAN framework to generate sphere-like surfaces. 

Point clouds are also widely used in representing 3D shapes due to their simplicity.
Following the pioneering work of Fan et al.~\cite{fan2017point}, many common generative models have been applied to point clouds, including generative adversarial networks \cite{achlioptas2017learning,li2018point}, adversarial autoencoders \cite{zamorski2018adversarial}, 
flow-based models \cite{yang2019pointflow} and autoregressive models \cite{sun2020pointgrow}. 
However, as point clouds do not describe the shape topology, such techniques can produce only relatively coarse geometry. Furthermore, compared to primitive based representations, they are less expressive and require considerably more points to represent shapes at a similar level of detail, making them less suitable for user interaction.

Implicit representations have recently shown great promise for generative 3D modeling \cite{park2019deepsdf,mescheder2019occupancy,chen2019learning}. These methods model shapes as isosurfaces of functions. Generally, models within this category predict the condition of sampled 3D  locations with respect to the watertight shape surface (e.g., inside/outside). Unlike explicit surface representations and point cloud representations, shapes are modeled as volumes instead of thin shells. Such models have been successfully applied to a variety of applications including shape generation, completion, and single-view reconstruction. As demonstrated in prior work, they are capable of representing shapes with high level of detail.

\medskip \noindent \textbf{3D modeling with primitive shapes.}
Reconstructing surfaces using simple primitives has long found application in reverse engineering \cite{benkHo2001algorithms}, and more generally in the computer vision and graphics communities \cite{roberts1963machine,biederman1987recognition,schnabel2007efficient}. Among other use cases, prior work has demonstrated their usefulness for reconstructing scanned \cite{gal2007surface}  or incomplete \cite{schnabel2009completion} point clouds.

Several primitive types have been proposed for modeling 3D shapes using neural networks, including cuboids \cite{tulsiani2017learning,smirnov2019deep}, superquadrics \cite{paschalidou2019superquadrics}, anisotropic 3D Gaussian balls  \cite{genova2019learning}, and convex polytopes \cite{deng2019cvxnets}.
Deprelle et al.~\cite{deprelle2019learning} learn which primitives best approximate a shape collection.

\medskip \noindent \textbf{Hybrid and hierarchical  representations.}
Hybird representations benefit from the complementary nature of different representations. There are prior works that assume a shared latent space across different representations and combine voxel-based, image-based, and point-based representations for various discriminative tasks, include 3D classification and segmentation \cite{hegde2016fusionnet,su2018splatnet,muralikrishnan2019shape}. 
However, none of them has addressed the problem of shape generation and manipulation.

Some previous works learn representations in several different resolutions, 
which has become the standard in computer vision~\cite{farabet2012learning,he2015spatial,chen2017deeplab,hao2017scale}.
Many recent image-generation methods also operate hierarchically, where fine-grained results are conditioned on coarser level outputs~\cite{gulrajani2016pixelvae,dorta2017laplacian,zhang2017stackgan,huang2017stacked,karras2019style,karras2019analyzing}. 
While these works primarily utilize multi-level approaches to improve performance, our work focuses on another important yet under-explored problem: semantic shape manipulation.

\medskip \noindent \textbf{Shape manipulation.}
Shape manipulation was traditionally utilized for character animation \cite{magnenat1988joint,lewis2000pose}, where the model is first rigged to a skeleton and then a transformation is assigned to each skeleton bone in order to deform the shape. One could consider our coarse proxy as a skeleton of the shape, allowing for a simple manipulation of the high resolution model. Tulsiani et al.~\cite{tulsiani2017learning} present a learning-based technique for abstract shape modeling, fitting 3D primitives to a given shape. They demonstrate a shape manipulation application that is similar in spirit to the one we propose. However, unlike our method, the coupling between their primitive representation and the input shape is hand-designed with simple transformations, thus their method 
cannot guide the manipulation towards producing semantically meaningful shapes. Similar problems have also been studied in the image domain, where a image is manipulated semantically given masks \cite{bau2019semantic}, scribbles \cite{zhu2016generative}, or motion trajectories \cite{hao2018controllable}.

\section{Method}
We first describe our shape representation in Sections \ref{sec:prim} and \ref{sec:deepsdf}. In Section \ref{sec:joint}, we describe how to learn a shared latent space over an entire collection of shapes and over multiple representations, while maintaining a tight coupling between representations. In Section \ref{sec:manip}, we describe our approach for shape manipulation using the proposed framework.

\begin{figure}
    \centering
        \includegraphics[width=\columnwidth, trim=0 0.2cm 3.05cm 7.85cm,clip]{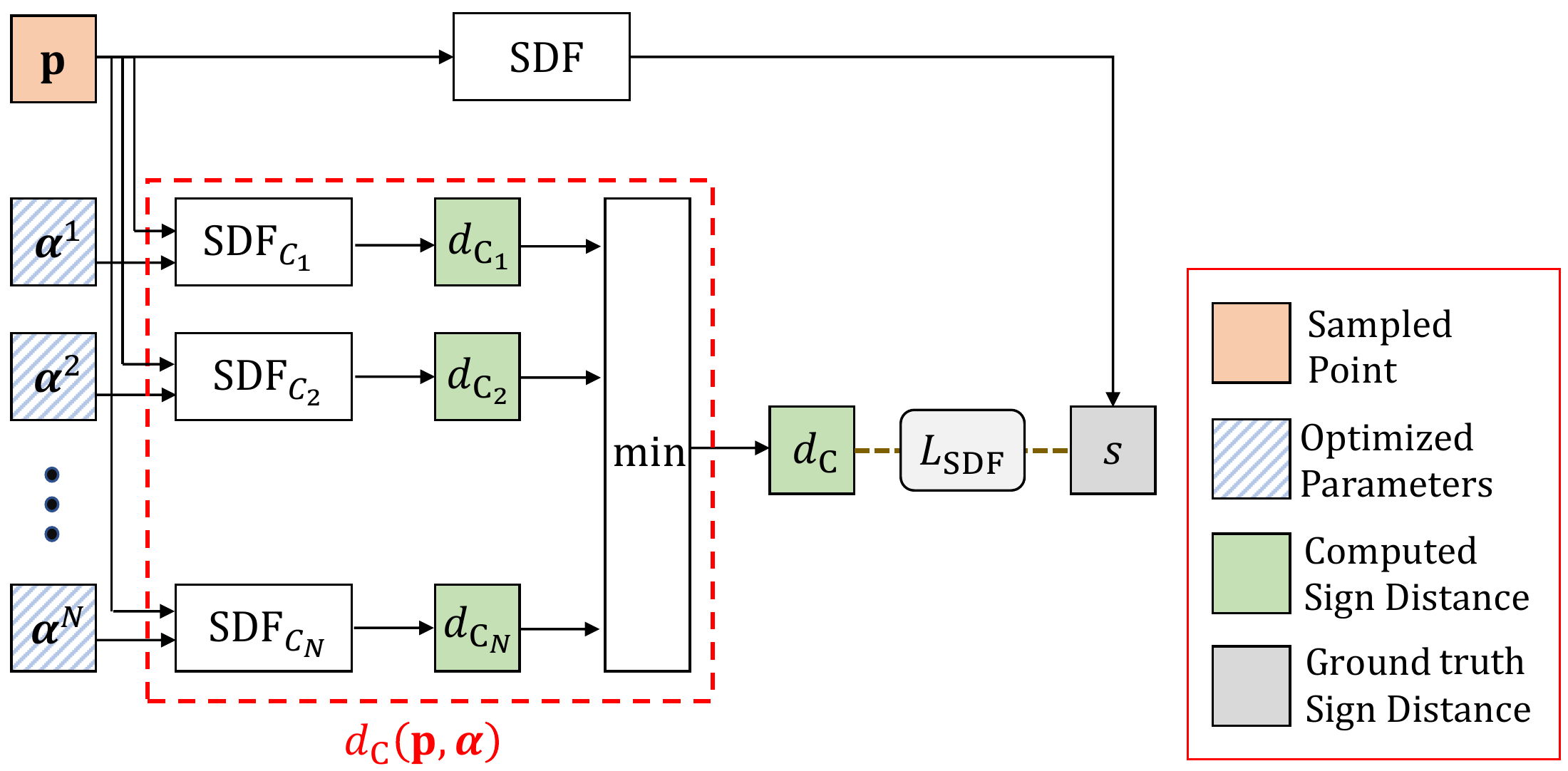}
    \caption{Learning a primitive-based representation of a single target shape. We optimize the parameters of the set of geometric elements (boxes colored with blue stripes) by minimizing the loss between the predicted and ground truth signed distance values on each sampled points.}
    \label{fig:PrimNet}
\end{figure}

\subsection{Coarse Primitive-based Shape Representation}
\label{sec:prim}
In this section, we describe our approach for approximating a 3D shape with a finite number of simple shape primitives such as spheres, rods, boxes, etc. First, we need to define a metric that measures how well the primitive-based representation approximates the ground truth. Following Tulsiani et al.~\cite{tulsiani2017learning}, we measure the difference of the signed distance fields between the target shape and the primitive-based representation.

A signed distance field specifies, for every point $\mbf{p}=\left(p_x,p_y,p_z \right)$, the distance from that point to the nearest surface, where the sign encodes whether the point is inside (negative) or outside (positive) the shape. Representing basic geometric shapes with distance fields is particularly appealing, as many of them have simple SDF formulations. Furthermore, Boolean operation across multiple shapes can be achieved using simple operators over the SDFs of individual shapes. Therefore, complex shapes can be represented in a straightforward manner as a union of simple primitives.

More precisely, we denote a set of $N$ basic shape primitives by tuples:
\begin{equation}
    \{ (\mathtt{C}^i, \pmb{\alpha}^i) | i=1,...,N \}
\end{equation}
where $\mathtt{C}^i$ describes the primitive type and $\pmb{\alpha}^i \in \mathbb{R}^{k^i}$ describes the attributes of the primitives. The dimensionality $k^i$ denotes the degree of freedom for primitive $i$, which vary across different choices of primitives. 
The signed distance function of a single element $i$ can thus be written as follows:
\begin{equation}
    d_{\mathtt{C}^i}\left(\mbf{p}, \pmb{\alpha}^i\right) =\mathtt{SDF}_{\mathtt{C}^i}\left(\mbf{p}, \pmb{\alpha}^i\right).
\end{equation}

An example of a simple geometric primitive is a sphere, which can be represented with $k^\mathtt{sphere}=4$ degrees of freedoms, i.e., $\pmb{\alpha}^{\mathtt{sphere}} = [\mbf{c}, r]$, where $\mbf{c} = \left(c_x,c_y,c_z \right)$ describe its center and $r$ is the radius. The signed distance function of the sphere takes the following form:
\begin{equation}
    d_{\mathtt{sphere}}\left(\mbf{p}, \pmb{\alpha}^{\mathtt{sphere}} \right) =\lVert\mbf{p}-\mbf{c} \rVert_2
    - r.
\end{equation}
For simplicity, we adopt spheres as our basic primitive type. However, as we later illustrate in Section \ref{sec:experiment}, our framework is directly applicable to other primitive types.

To approximate the signed distance function of an arbitrarily complex shape, we construct the signed distance function of the union of the geometric elements (spheres in our case):
\begin{equation}
    \pmb{\alpha} = \left[\pmb{\alpha}^1,...,\pmb{\alpha}^N \right],
\end{equation}
\begin{equation}
    d_{\mathtt{C}}\left(\mbf{p}, \pmb{\alpha} \right) = \min_{1 \leq i \leq N}  d_{\mathtt{C}^i}\left(\mbf{p}, \pmb{\alpha}^i \right).
\end{equation}
Alternatively, smooth minimum functions like \textit{LogSumExp} can be used in place of the (hard) minimum function to get a smooth transition over the interface between geometric elements. We refer the readers to Frisken et al.~\cite{frisken2006designing} For an in-depth explanation of signed distance fields and their Boolean operations.

To train the primitive-based model, given a target shape $x$ (usually in the form of a mesh), we sample pairs of 3D points $\mbf{p}_t$ and their corresponding ground truth signed distance values $s_t = \mathtt{SDF}_x(\mbf{p}_t)$. $\pmb{\alpha}$ can be learned by minimizing the difference between predicted and real signed distance values:
\begin{equation}
    \hat{\pmb{\alpha}} = \argmin_{\pmb{\alpha}} \sum_t{L_\mathtt{SDF}\left( d_{\mathtt{C}}\left(\mbf{p}_t, \pmb{\alpha} \right),s_t \right)}.
\end{equation}
Figure \ref{fig:PrimNet} shows the full structure of our primitive-based model.

\begin{figure*}
	\jsubfig{\includegraphics[height=5.5cm,trim=0.1cm 0.2cm 5.5cm 6cm,clip]{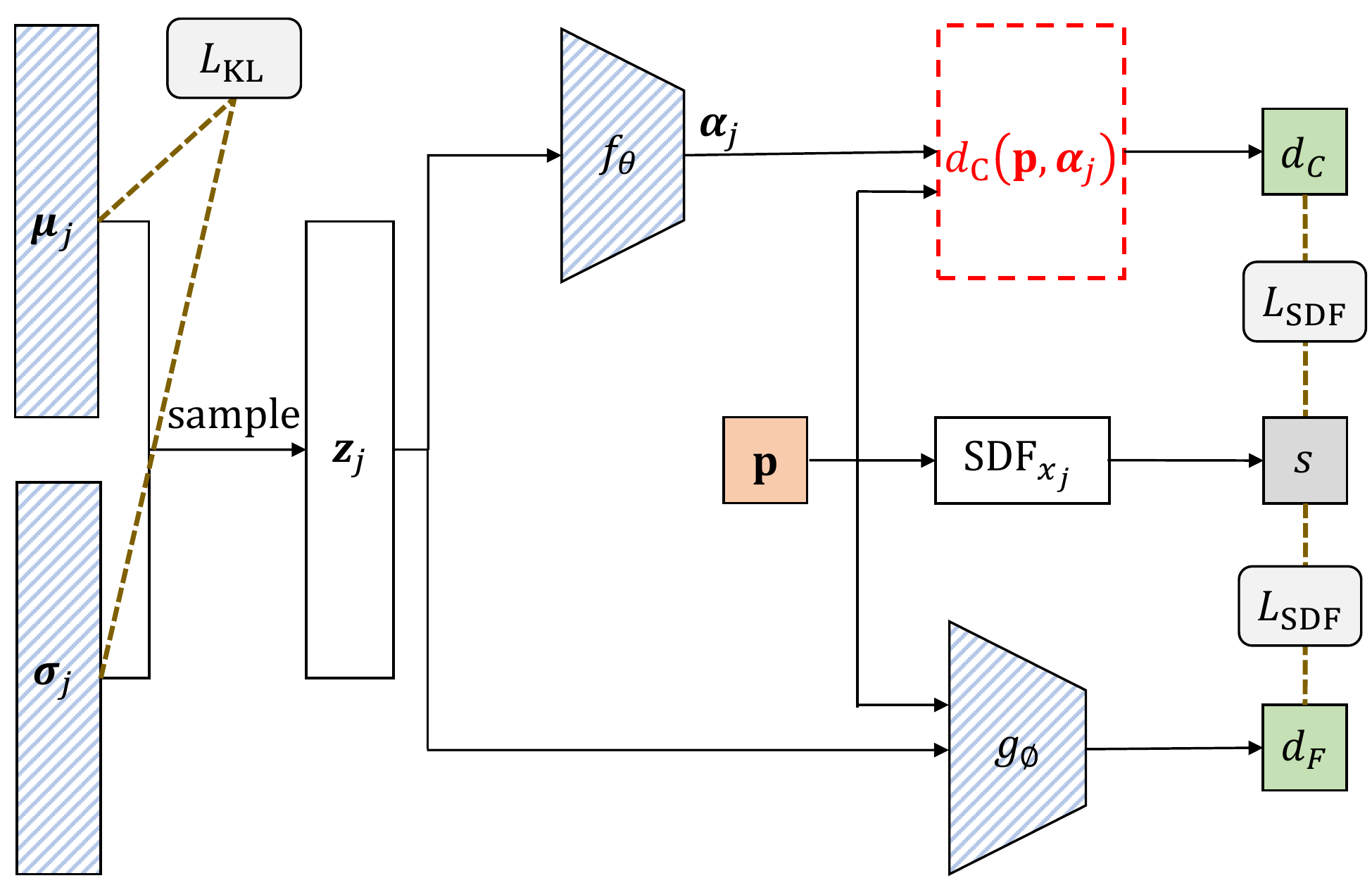}}
	{Training Optimization}%
	\hfill
	\jsubfig{\includegraphics[height=5.5cm,trim=0.1cm -2.3cm 12.7cm 8.5cm,clip]{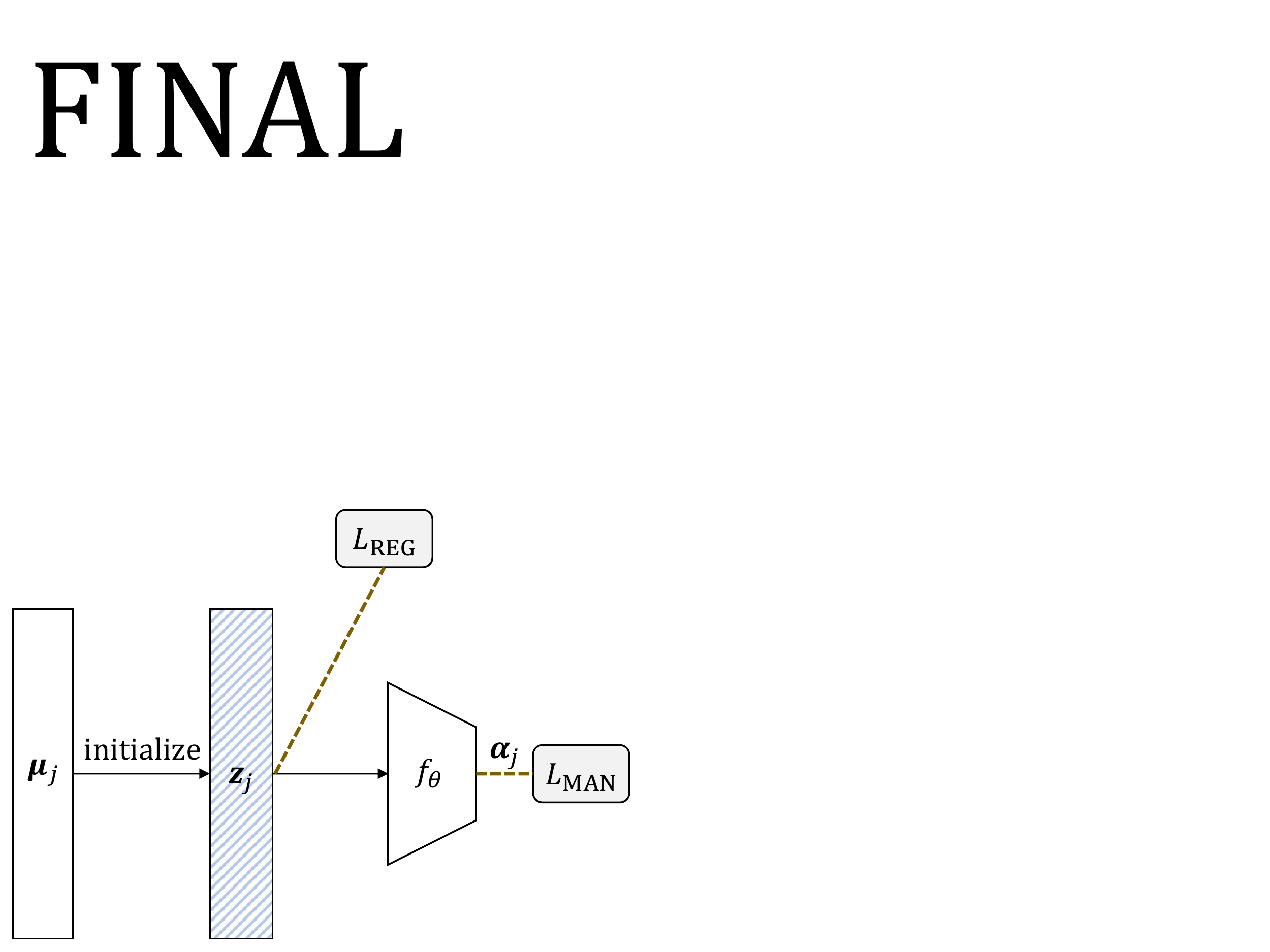}}
	{Manipulation Optimization}%
	\hfill
	\jsubfig{\includegraphics[height=5.5cm,trim=0.1cm 0.2cm 19.8cm 6cm,clip]{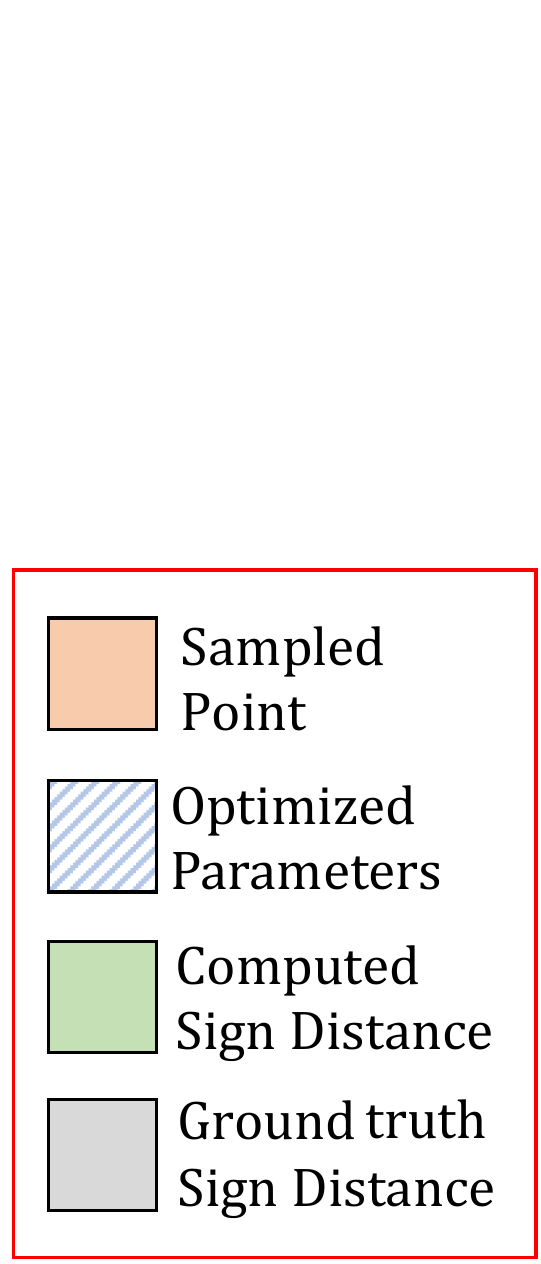}}
	{ }%
    \caption{The training and manipulation stages of our two-level model. During training (left), we jointly learn the posterior distributions (for each shape $j$) and the shared networks $f_\theta$ and $g_\phi$. The dotted red rectangle is detailed in Figure \ref{fig:PrimNet}. During manipulation (right), the networks remain fixed and only the latent code of the $j$-th shape is updated.}
    \label{fig:training}
\end{figure*}

\subsection{High Resolution Shape Representation}
\label{sec:deepsdf}
We adopt DeepSDF \cite{park2019deepsdf} for our fine-scale shape representation. Similar to the coarse-scale representation, the shapes are modeled with SDFs.
However, instead of constraining the shape to be within the family of shapes that can be constructed by simple primitives, we directly learn the signed distance function with a neural network $g_\phi$:
\begin{equation}
    g_\phi(\mbf{p}) \approx \mathtt{SDF}_x (\mbf{p}).
\end{equation}

Just like the coarse representation, its zero iso-surface w.r.t.\ $\mbf{p}$ implicitly defines the surface of the shape, and can be retrieved efficiently with ray-marching algorithms.
The training of the fine-scale SDF model follows the same procedure as the coarse-scale model, described in Section \ref{sec:prim}.

\subsection{Learning a Tightly Coupled Latent Space}
\label{sec:joint}
We learn a two-level shape representation over an entire class of shapes $\{x_j | j = 1,...,M\}$ by using two representation models that share the same latent code $\mbf{z}_j$ (Figure \ref{fig:training} left).

For representing multiple shapes with the primitive based coarse-scale representation, we reparameterize $\pmb{\alpha}$ with a neural network $f_\theta$:
\begin{equation}
    \pmb{\alpha}_j = f_\theta(\mbf{z}_j),
\end{equation}
where $f_\theta$ is shared across all shapes. Likewise, for the fine-scale representation, we condition the neural network $g_\phi$ on the latent code $\mbf{z}_j$:
\begin{equation}
    g_\phi(\mbf{z}_j, \mbf{p}) \approx \SDF_{x_j}(\mbf{p}).
\end{equation}

To ensure that the manipulation made on one representation has the same effect on other representations, we would like to learn a shared latent space where every feasible latent vector is mapped to the same shape in both representations (see Figure \ref{fig:JointLatent} for an illustrative example). Furthermore, we also expect the latent space to be compact, so that latent code interpolation and optimization become less likely to ``fall off the manifold.'' Thus we utilize the variational auto-decoder (VAD) framework~\cite{zadeh2019variational} which enforces a strong regularization on the latent space by representing the latent vector of each individual shape ($\mbf{z}_j$) with the parameters of its approximate posterior distributions ($\muj$, $\sigmaj$), similar to a VAE~\cite{kingma2013auto}.

In the language of probability, we select the family of Gaussian distributions with diagonal covariance matrix as the approximate posterior of $\mbf{z}$ given shape $x_j$:
\begin{equation}
    q(\mbf{z}|x = x_j) := \mathcal{N}(\mbf{z}; \pmb{\mu}_j, \pmb{\sigma}_j^2 \cdot\mbf{I}).
\end{equation}
We apply the reparameterization trick~\cite{kingma2013auto}, sampling $\pmb{\epsilon} \sim \mathcal{N}(\mbf{0},\mbf{I})$ and setting $\mbf{z}_j = \pmb{\mu}_j + \pmb{\sigma}_j \odot \pmb{\epsilon}$ to allow direct optimization of the distribution parameters $\muj$ and $\sigmaj$ via gradient descent.

During training, we maximize the lower bound of the marginal likelihood (ELBO) over the whole dataset, which is the sum over the lower bound of each individual shape $x$ presented below:
\begin{align}
    \label{eq:elbo}
    \log p_{\theta, \phi}(x) &\ge \mathbb{E}_{\mbf{z} \sim q(\mbf{z}|x)}[\log p_{\theta, \phi}(x|\mbf{z})] \nonumber \\ 
    &\qquad - D_{KL}(q(\mbf{z}|x) || p(\mbf{z})).
\end{align}
Here the learnable parameters are $\theta$, $\phi$, as well as the variational parameters $\{(\muj,\sigmaj) | j = 1,...,M\}$ that parameterize $q(\mbf{z}|x)$. 
Since we would like the two representations to be tightly coupled, i.e., to both assign high probability density to a shape $x_j$ given its latent code $\mbf{z}_j \sim q(\mbf{z}|x=x_j)$, we model the first term of Eq.~\ref{eq:elbo} using a a mixture model:
\begin{align}
    \label{eq:twobranchlikelihood}
    p_{\theta, \phi}(x|\mbf{z}) := \frac{p_{\theta}(x|\mbf{z}) + p_{\phi}(x|\mbf{z})}{2}.
\end{align}
Here $p_{\theta}(x|\mbf{z})$ and $p_{\phi}(x|\mbf{z})$ are the posterior distributions of coarse and fine representations, implied by the signed distance function loss $L_{\mathtt{SDF}}$ and its sampling strategies. Following Park et al.~\cite{park2019deepsdf}, we assume they take the form of:
\begin{align}
    \label{eq:primlikelihood}
    \log p_\theta(x | \mbf{z})\! = \! - \lambda_1 \! \int \! p(\mbf{p}) L_{\mathtt{SDF}}\Big(d_c(\mbf{p}, f_\theta(\mbf{z})), \mathtt{SDF}_x(\mbf{p})\Big) d\mbf{p},
\end{align}
\begin{align}
    \label{eq:deepsdflikelihood}
    \log p_\phi(x | \mbf{z}) = - \lambda_2 \int p(\mbf{p}) L_{\mathtt{SDF}}\Big(g_\phi(\mbf{z}, \mbf{p}), \mathtt{SDF}_x(\mbf{p})\Big)d\mbf{p}.
\end{align}
Eq.~\ref{eq:primlikelihood} and \ref{eq:deepsdflikelihood} can be approximated via Monte Carlo method, where $\mbf{p}$ is sampled randomly from the 3D space following a specific rule $p(\mbf{p})$.

The benefits of using a VAD objective are two-fold: First, it encourages the model to learn a smooth and densely packed latent space. A similar effect has been leveraged in a related technique called \textit{conditioning augmentation}~\cite{zhang2017stackgan}. This not only benefits optimization-based manipulation, but also improves coupling on novel shapes (shapes not seen during training). Secondly, being able to model the lower bound of the likelihood of every shape provides us with a way of regularizing the manipulation process by actively guiding the user away from unlikely results (Section \ref{sec:manip}). 
Detailed experiment and analysis on the effect of VAD are presented in Section \ref{sec:experiment}.

\subsection{Interactive Shape Manipulation}
\label{sec:manip}
Our two-level model enables users to perform modifications on the primitive-based representation in an interactive manner while simultaneously mirror the effect of the modifications onto the high-resolution representation. Additionally, our model is able to augment and regularize the user input in order to avoid generating unrealistic shapes. This form of manipulation is extremely useful, as it is generally hard for users to directly edit the mesh of a 3D shape. Even for a minor change, many accompanying (and time-consuming) changes are required to obtain a reasonable result.

In contrast, shape manipulation is much more intuitive for users with our model. To start with, we encode a user-provided shape into the latent space by optimizing the variational parameters w.r.t. the same VAD objective used during training. Alternatively, we can also start with a randomly sampled shape.
Users can then efficiently modify the high-resolution shape by manipulating the shape primitives that represents parts of the shapes. 

Our model support any manipulation operation that can be expressed as minimizing an objective function over primitive attributes $\pmb{\alpha}$, such as increasing the radius of a sphere, moving a primitive one unit further towards the $z$ axis, or increasing the distance between two primitives, as well as a combination of them. The manipulation operation can be either dense, which involves all the attributes, or sparse, which only involves a subset of attributes or primitives. In the case of sparse manipulations, our model can automatically adapt the value of the unconstrained attributes in order to produce a more convincing result. For example, when a user makes one of the legs of a chair longer, the model automatically adjusts the rest of the legs, resulting a valid chair.

To reiterate, $\pmb{\alpha}$ contains the location as well as the primitive-specific attributes for each primitive. We use gradient descent to minimize the given objective function by optimizing the $\mbf{z}$:
\begin{equation}
    \label{eq:manip_optim}
    \hat{\mbf{z}} = \argmin_{\mbf{z}}\Big(L_{\mathtt{MAN}}(f_\theta(\mbf{z})) + L_{\mathtt{REG}}(\mbf{z})\Big),
\end{equation}
\begin{equation}
    L_{\mathtt{REG}}(\mbf{z}) = \gamma \max (\lVert \mbf{z} \rVert_2^2, \beta).
\end{equation}
Note that $L_{\mathtt{MAN}}$ is the optimization objective of the specific manipulation operation. For example, the objective of moving a single sphere $i$ (parameterized by $\pmb{\alpha}^i= [c_i, r_i]$) to a new position $\hat{c}$ is as follows: 
\begin{equation}
    L_\textit{MAN}^\mathrm{Move}(\pmb{\alpha}) = \lVert c_i - \hat{c}\rVert_2
\end{equation}
The attributes that are not constrained by the objective, including the position of other spheres, as well as the radii of all the spheres, are allowed to adjust freely during the optimization.

The latent code $\mbf{z}$ is initialized as the expectation of $q(\mbf{z}|x)$, where $x$ is the shape to be modified. An appropriate choice of $\gamma$ and $\beta$ in the regularization term ensures a likely $\mbf{z}$ under the Gaussian prior, which empirically leads to a more plausible shape. Multiple different manipulation steps can be executed consecutively to achieve complex or interactive manipulations. The optimization process is illustrated in Figure \ref{fig:training} (right).

Another important prerequisite for a successful shape manipulation framework is that every individual primitive should stay approximately at the same region of the shape throughout the entire class of shapes. As we later show in Section \ref{sec:experiment}, primitives retain their semantic meanings well across all the shapes. 

Our model is also advantageous in terms of speed. The coarse model can run at an interactive rate, which is crucial in providing users with immediate feedback. The high-resolution model is capable of dynamically adjusting the trade-off between quality and speed by using different rendering resolution and different number of ray-marching iterations. High quality result can be rendered only as needed, once the user is satisfied with the manipulated result.

\section{Experiments}
\label{sec:experiment}
\begin{figure*}
\centering%
\includegraphics[trim={6cm 0.2cm 6cm 0.2cm},clip,width=0.155\textwidth]{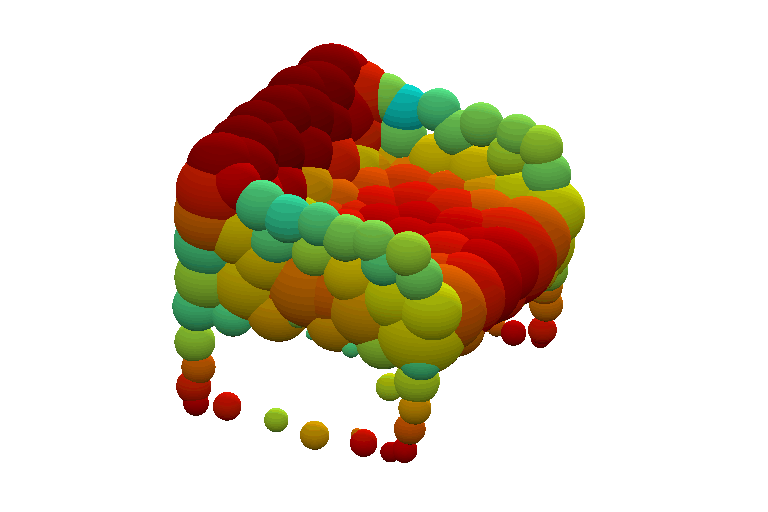}
\includegraphics[trim={6cm 0.2cm 6cm 0.2cm},clip,width=0.155\textwidth]{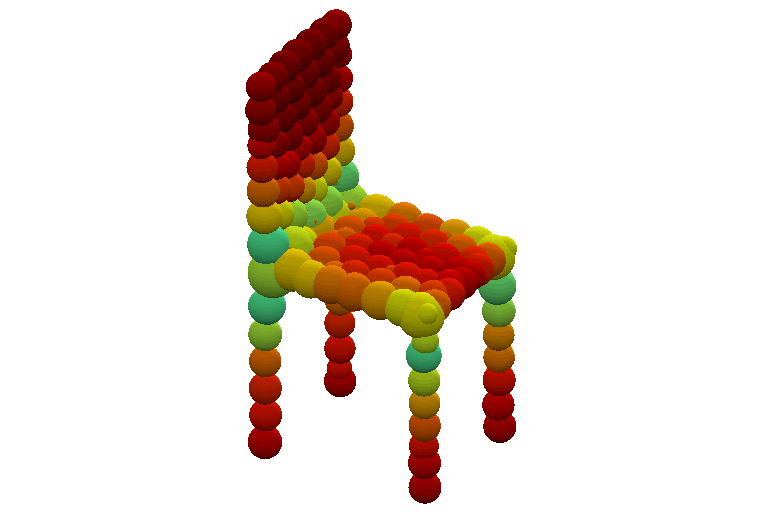}
\includegraphics[trim={6cm 0.2cm 6cm 0.2cm},clip,width=0.155\textwidth]{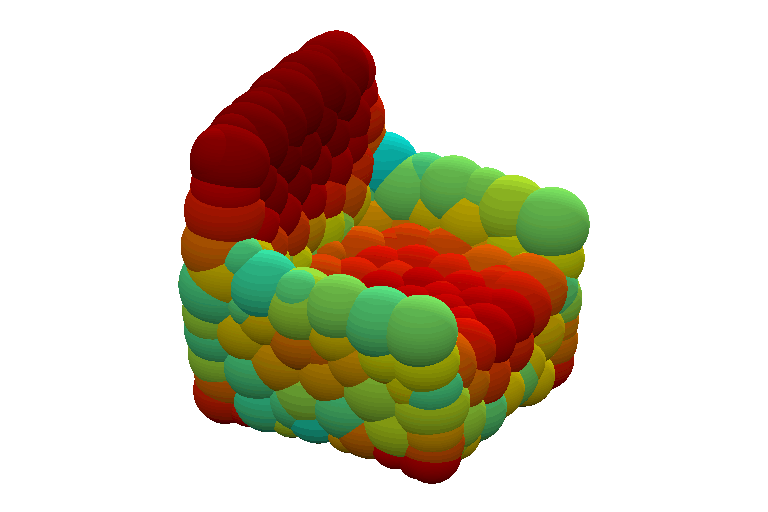}
\includegraphics[trim={6cm 0.2cm 6cm 0.2cm},clip,width=0.155\textwidth]{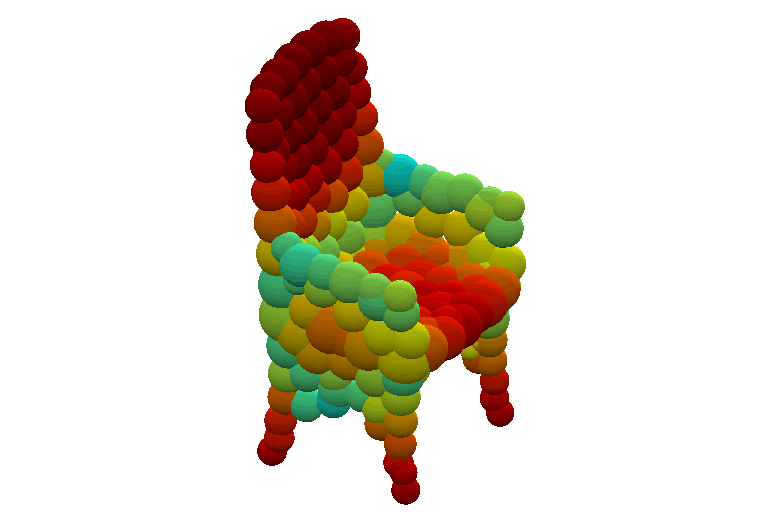}
\includegraphics[trim={6cm 0.2cm 6cm 0.2cm},clip,width=0.155\textwidth]{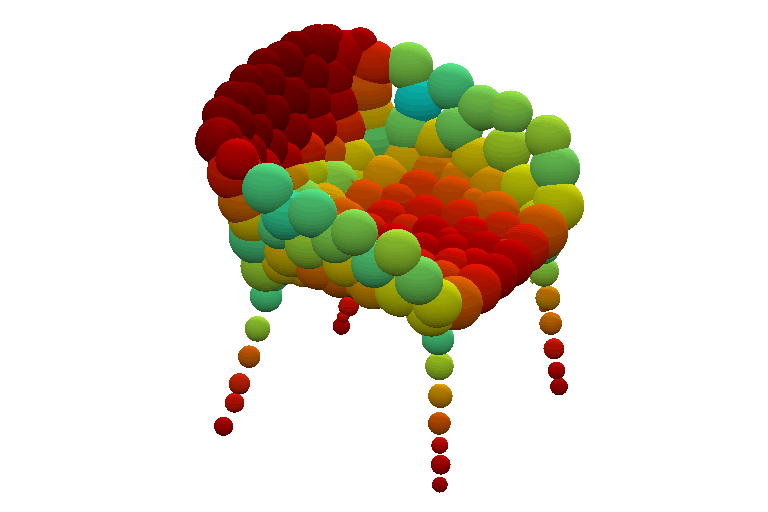}
\includegraphics[trim={6cm 0.2cm 6cm 0.2cm},clip,width=0.155\textwidth]{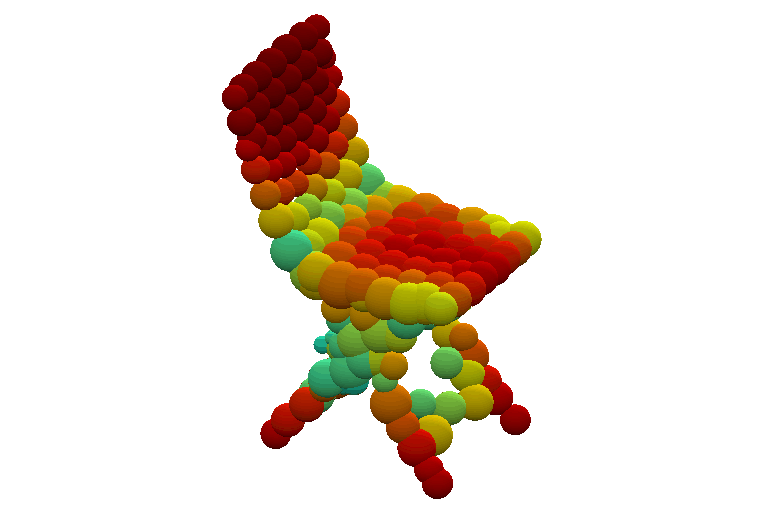}
\includegraphics[width=0.017\textwidth]{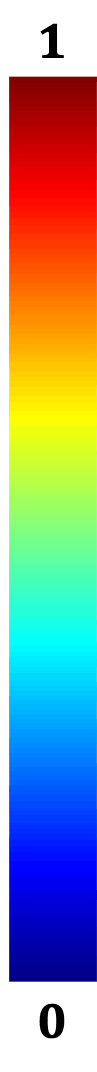} \\
\vspace{-4pt}
\caption{Measuring semantic consistency across the entire Chair collection. Above we illustrate the scores obtained on a few chair samples, where each primitive is colored according to the consistency score computed over the entire collection. Warmer colors correspond to higher scores (more consistent). }
\label{fig:SemanticRand}
\end{figure*}
\begin{figure*}
\newcommand{\tu}{4.3cm}
\newcommand{\tb}{3.5cm}
\centering%
\includegraphics[trim={3cm {\tb} 3cm {\tu}},clip,width=0.16\textwidth]{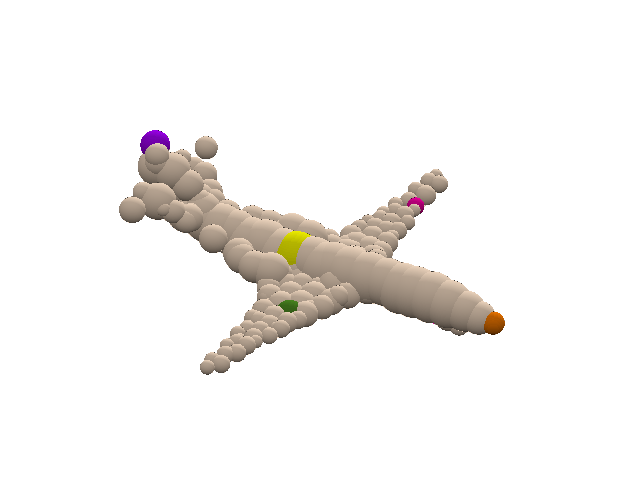}
\includegraphics[trim={3cm {\tb} 3cm {\tu}},clip,width=0.16\textwidth]{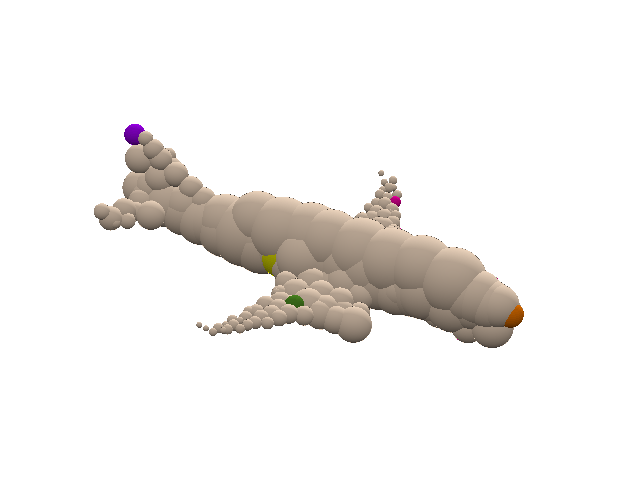}
\includegraphics[trim={3cm {\tb} 3cm {\tu}},clip,width=0.16\textwidth]{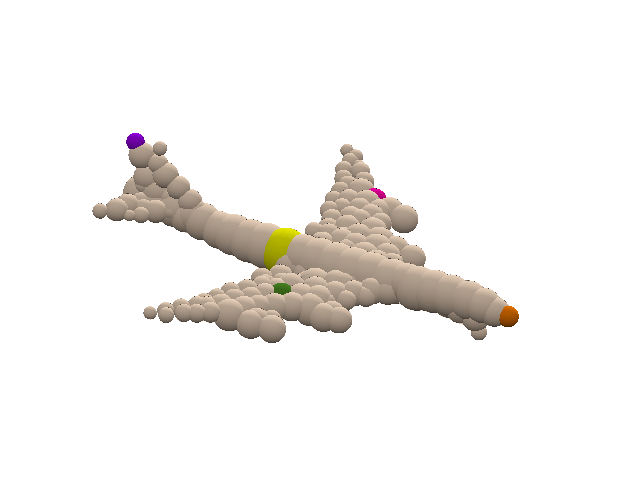}
\includegraphics[trim={3cm {\tb} 3cm {\tu}},clip,width=0.16\textwidth]{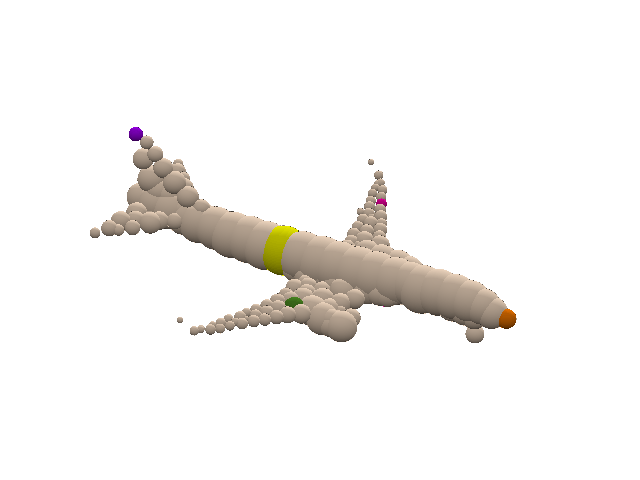}
\includegraphics[trim={3cm {\tb} 3cm {\tu}},clip,width=0.16\textwidth]{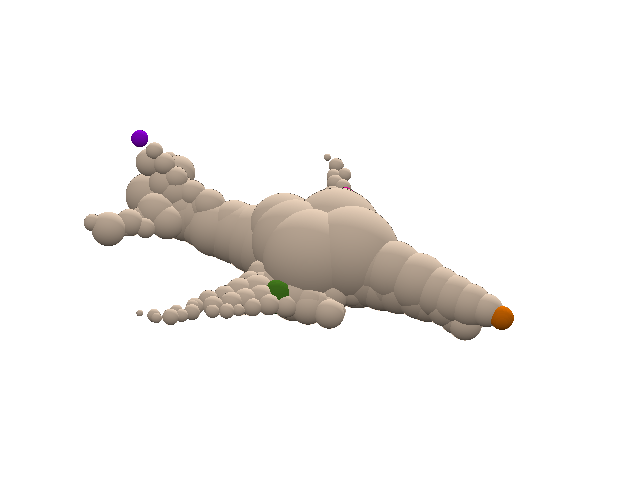}
\includegraphics[trim={3cm {\tb} 3cm {\tu}},clip,width=0.16\textwidth]{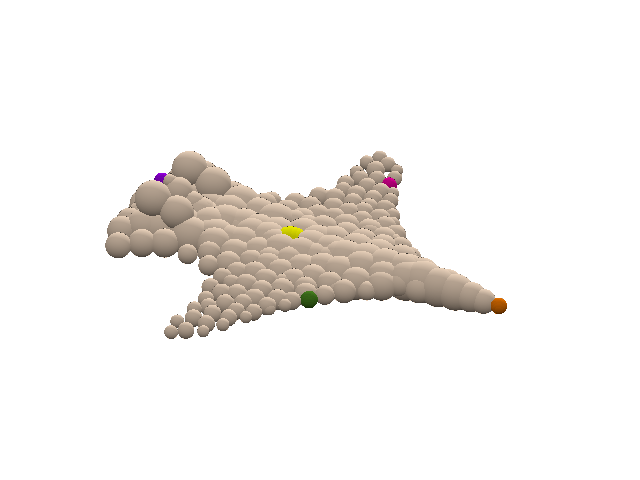}\\
\includegraphics[trim={3cm {\tb} 3cm {\tu}},clip,width=0.16\textwidth]{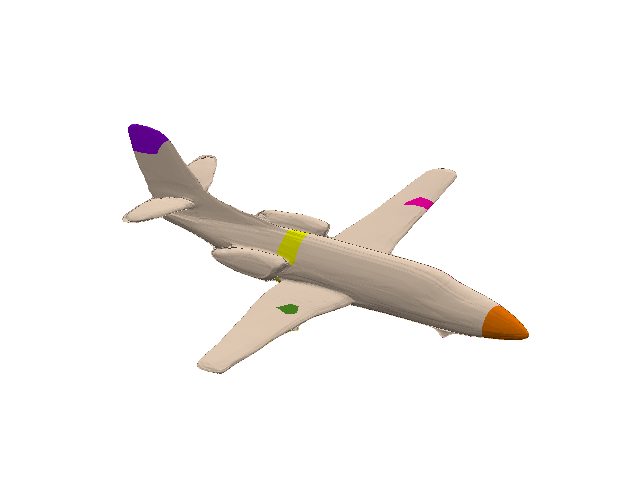}
\includegraphics[trim={3cm {\tb} 3cm {\tu}},clip,width=0.16\textwidth]{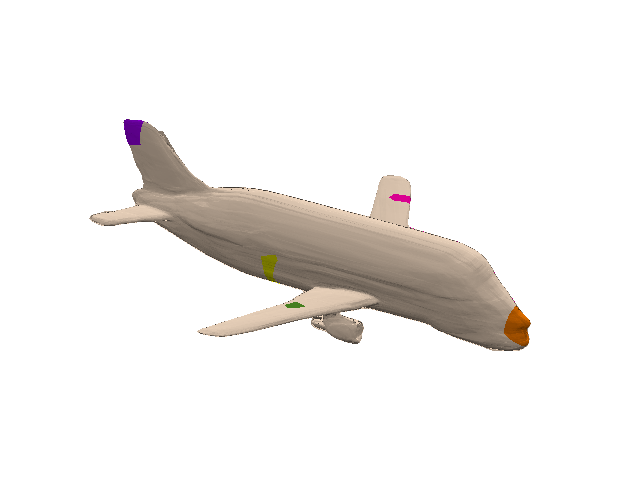}
\includegraphics[trim={3cm {\tb} 3cm {\tu}},clip,width=0.16\textwidth]{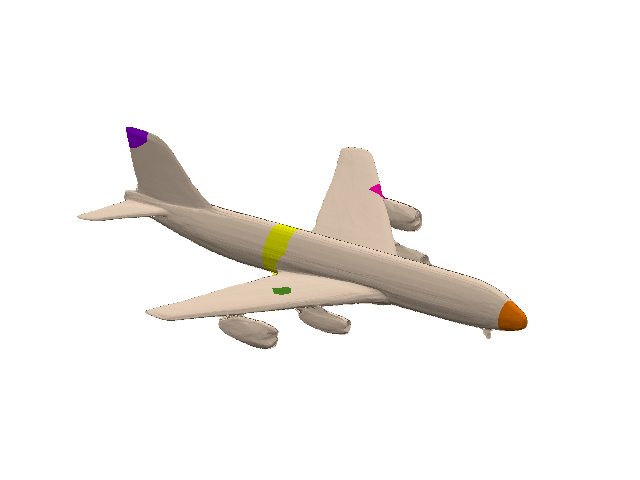}
\includegraphics[trim={3cm {\tb} 3cm {\tu}},clip,width=0.16\textwidth]{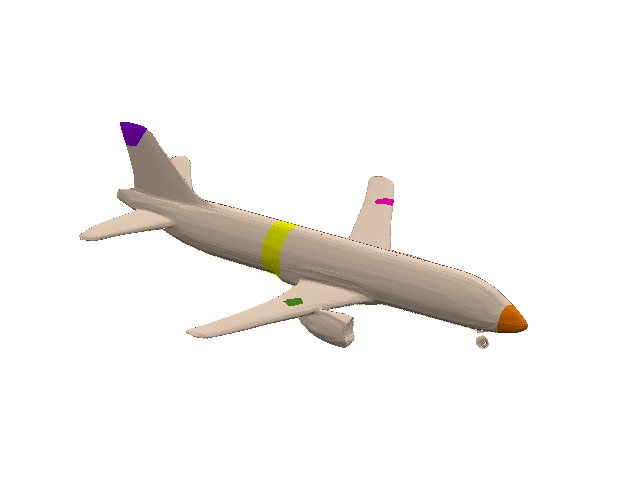}
\includegraphics[trim={3cm {\tb} 3cm {\tu}},clip,width=0.16\textwidth]{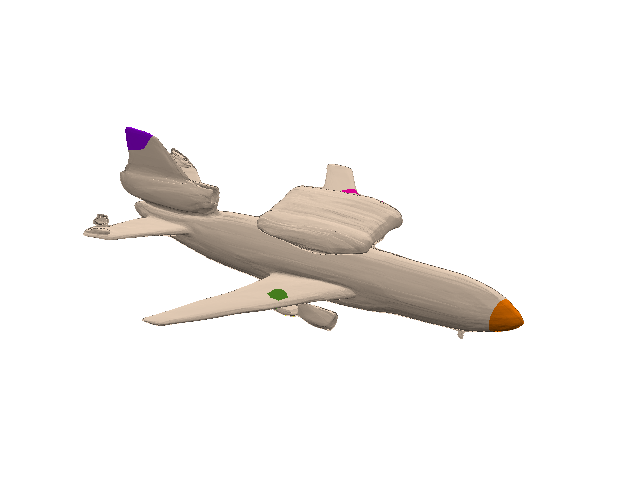}
\includegraphics[trim={3cm {\tb} 3cm {\tu}},clip,width=0.16\textwidth]{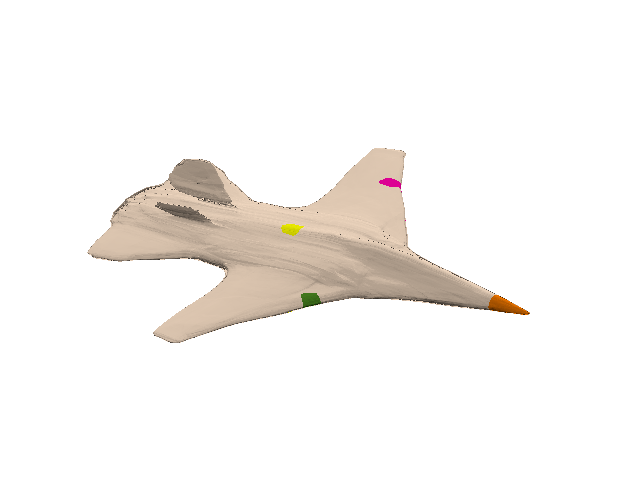} \\
\caption{Shape correspondence via the coarse shape proxy. Above we demonstrate shape reconstructions from the Airplane dataset, with several primitives highlighted in unique colors. As the figure illustrates, the shape primitives are consistently mapped to the same regions. These correspondences can then be propagated to the fine-scale reconstructions.   }
\label{fig:correspondence}
\end{figure*}

We demonstrate the shape representation power of our model as well as its potential for shape manipulation with various experiments.

We first show that our model is capable of representing shapes in high quality, comparing it with various state-of-the-art methods on the ShapeNet dataset~\cite{chang2015shapenet}, using a set of standard quality metrics.

To demonstrate the suitability of our model in the context of shape manipulation, we separately evaluate two aspects: First, we evaluate how tightly the two levels of representations are coupled by sampling novel shapes from the latent space and evaluating the volumetric intersect-over-union (IoU) between the two representations. As all of the manipulations are first performed on the primitive-based representation and then propagated to high-resolution representation through the latent code, a tight coupling is a crucial indicator for \textit{faithful} shape manipulation. Second, we evaluate how well each primitive retains its semantic meaning across different shapes with a semantic consistency score. A semantically consistent primitive stays associated to the same part of the object across all the objects, which enables \textit{intuitive} shape manipulation. We complement the quantitative evaluation by presenting a diversified collection of shapes manipulated with our method, demonstrating the \textit{flexibility} of our manipulation framework and the \textit{fidelity} of the result.

\medskip \noindent \textbf{Data preparation.}
We normalize each individual shape to be inside a unit sphere. To sample signed distance values from mesh, we implemented a custom CUDA kernel for calculating the minimum distance from a point to the mesh surface. To determine the inside/outside of each point (and thus its sign), we use a ray stabbing method~\cite{nooruddin2003simplification}, which is robust to non-watertight meshes and meshes with internal structures and it does not require any pre-processing. For training the high-resolution representation, we use the same sampling strategy used in Park et al. \cite{park2019deepsdf}. For training the primitive-based representation, we sample points uniformly inside a unit sphere centered at the origin.

\begin{table}
\centering%
\vspace{-5pt}
\ra{1.0}
\setlength{\tabcolsep}{1.9pt}
\small
\begin{tabular}{@{}llcccccccccccccccccc@{}}
\toprule
  && \multicolumn{4}{c} {\textbf{Airplane}} && \multicolumn{4}{c} {\textbf{Chair}}  \\
 && CD$^\star$ & CD$^\dag$ & EMD & ACC && CD$^\star$ & CD$^\dag$ & EMD & ACC \\ \midrule
 AtlasNet-Sph. && 0.19 & 0.08 & 0.04 & 0.013  &&  0.75 & 0.51 & 0.07 & 0.033 \\
 AtlasNet-25 &&   0.22 & 0.07 & 0.04 & 0.013 &&  0.37 & 0.28 & 0.06 & 0.018 \\
  DeepSDF &&   0.14 & 0.04 & 0.03 & 0.004 &&  0.20 & 0.07 & 0.05 & 0.009  \\
 DualSDF  &&   0.22 & 0.14 & 0.04 & 0.010 &&  0.45 & 0.21 & 0.05 & 0.014 \\
 \midrule
DualSDF (K)  &&   0.19 & 0.13 & 0.04 & 0.009 &&  0.65 & 0.19 & 0.05 & 0.012 \\ 
\bottomrule
\end{tabular}
\caption{Reconstruction results on unknown shapes (top rows) and known (K) shapes (bottom row) for the Airplane and Chair collections. We report the mean and median of Chamfer distance (denoted by CD$^\star$ and  CD$^\dag$, respectively, multiplied by $10^3$), EMD and mesh accuracy (ACC).
\label{tab:reconstruction}
}
\end{table}

\medskip \noindent \textbf{Shape reconstruction.} We report reconstruction results for known and unknown shapes (i.e., shapes belonging to the train and test sets) in Table \ref{tab:reconstruction}. Following prior work, we report several metrics: Chamfer distance (mean and median), EMD and mesh accuracy~\cite{seitz2006comparison}.

For unknown shapes, we compare our reconstruction performance against two variants of AtlasNet \cite{groueix2018atlasnet} (one generating surfaces from a sphere parameterization and one from 25 square patches) and DeepSDF \cite{park2019deepsdf}, which we adopt for our fine-scale representation. As the table illustrates, our reconstruction performance is comparable to state-of-the-art techniques. As suggested in Park et al. \cite{park2019deepsdf}, the use of a VAD objective trades reconstruction performance for a smoother latent space.

\medskip \noindent \textbf{Effect of VAD objective on cross-representation consistency.}
\begin{table}
\centering%
\vspace{-5pt}
\ra{1.0}
\setlength{\tabcolsep}{2.8pt}
\begin{tabular}{@{}lccccc@{}}
\toprule
            & \multicolumn{5}{c} {Intersection-over-union (IoU)} \\
            & \textbf{Airplane}  & \textbf{Car}   & \textbf{Chair} & \textbf{Bottle} & \textbf{Vase} \\
\midrule
DualSDF (S) & 0.52& 0.76 & 0.50& 0.68& 0.44 \\
w/o VAD (S\textsuperscript{\dag}) & 0.41       & 0.65   & 0.30   & 0.58   & 0.29  \\
\midrule
DualSDF (K) & 0.56 & 0.70 & 0.53 & 0.69 & 0.54  \\
w/o VAD (K) & 0.53       & 0.70   & 0.53  & 0.69   &  0.55 \\
\bottomrule
\end{tabular}
\caption{Cross-representation consistency evaluation. In the top rows, we measure the consistency of primitive based model and the high resolution model by randomly sampling (S) shapes from the latent space and calculating the intersection-over-union (IoU) of the two representations. We also report scores over known (K) shapes in the bottom rows. Note that due to the approximate nature of primitive based model, the numbers are only comparable with models trained under similar settings. \textsuperscript{\dag}We train an additional VAE on top of the latent code to enable sampling.
\label{tab:coupling}
}
\end{table}
We evaluate the consistency between fine and coarse shapes generated with our model by randomly sampling shapes from the latent space and evaluating the average volumetric IoU. We also evaluate the mean IoU on training data as a reference. We compare our method against a baseline method which uses the same backbone network and training procedure, with the only difference being that it uses an auto-decoder~\cite{park2019deepsdf} objective instead of a VAD objective. Results are shown in Table \ref{tab:coupling}. While both models perform similarly on shapes in the training set, VAD significantly boosts the cross-representation consistency on novel generated shapes. We conjecture that the improved consistency comes from the fact that, unlike the auto-decoder objective which only focuses on individual data points, the VAD objective actively explores the latent space during training.

\begin{table}
\centering%
\vspace{-5pt}
\ra{1.0}
\setlength{\tabcolsep}{2.8pt}
\begin{tabular}{@{}lccccccccccccccccccc@{}}
\toprule
Dataset & \#lbls &&& Top-1 & Top-2 & Top-3  \\ \midrule
 Chair & 5 &&&  0.71 & 0.91 & 0.98  \\
 Bottle & 5 &&&  0.90 & 0.96 & 0.99  \\
 Vase & 3 &&&  0.80 & 0.98 & 1.00  \\
\bottomrule
\end{tabular}
\caption{Semantic consistency evaluation. For each primitive index, we measure the fraction of shapes in each collection that agree with that primitive's most commonly associated labels (i.e., the top-1, top-2 and top-3 most frequent labels). We report averages over all the primitives.
\label{tab:semantic}
}
\end{table}

\medskip \noindent \textbf{Semantic part-based abstraction.}
We perform a quantitative evaluation on the PartNet dataset \cite{mo2019partnet} to demonstrate that the semantic interpretation of the primitives in our model is consistent across different shapes.
PartNet dataset contains part-labels of several levels of granularities. We train our model individually on Chair, Bottle and Vase collections, and evaluate the semantic consistency of the learned shape primitives using the $1000$ labeled 3D points (per shape) provided by the dataset. We measure performance on the first level of the hierarchies, which contains 3-5 semantic labels per category. 
We would like to show that primitives are consistently mapped to the same semantic part of the shapes across the entire shape collection. Thus, for each shape, we assign primitives with part labels according to their closest labeled 3D point. We calculate the semantic consistency score by measuring the fraction of shapes in the collection that agree with the most frequent labels. 

In Figure \ref{fig:SemanticRand} we illustrate the per-primitive semantic consistency scores on several samples from the Chair category. As the figure illustrates, some primitives have a clear semantic meaning (e.g., the legs of the chairs are consistently labelled as chair legs). Also unavoidably, some primitives have to ``adjust'' semantically to accommodate for the large variability within the shape collection, for instance, to generate chairs with and without arms. In Table \ref{tab:semantic} we report the average scores obtained on all the primitives (for each collection). 
We also report the fraction of shapes that agree with the top-2 and the top-3 labels. As the table illustrates, the semantic meanings of the primitives learned by our model are highly consistent among different shapes. This property allows the user to intuitively regard primitives as the proxies for shape parts.

\begin{figure}
\centering%
\includegraphics[trim={5cm 0 5cm 0},clip,width=0.19\columnwidth]{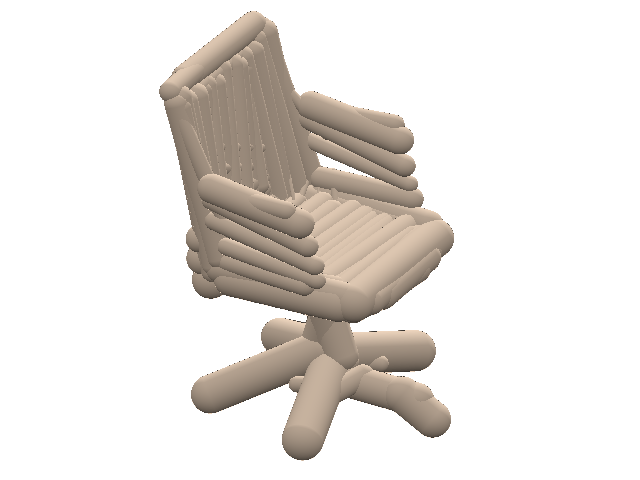}
\includegraphics[trim={5cm 0 5cm 0},clip,width=0.19\columnwidth]{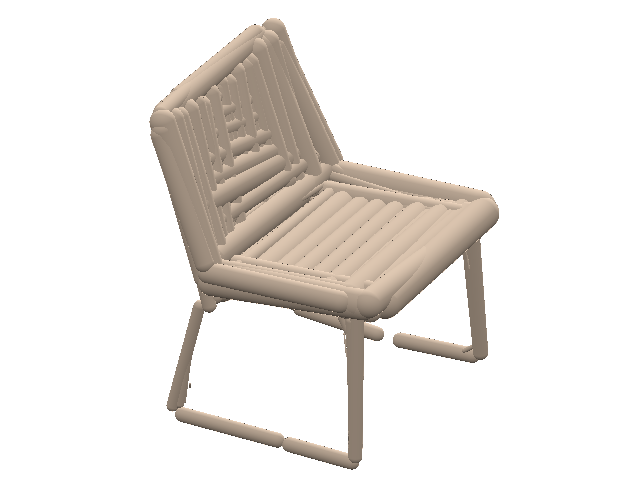}
\includegraphics[trim={5cm 0 5cm 0},clip,width=0.19\columnwidth]{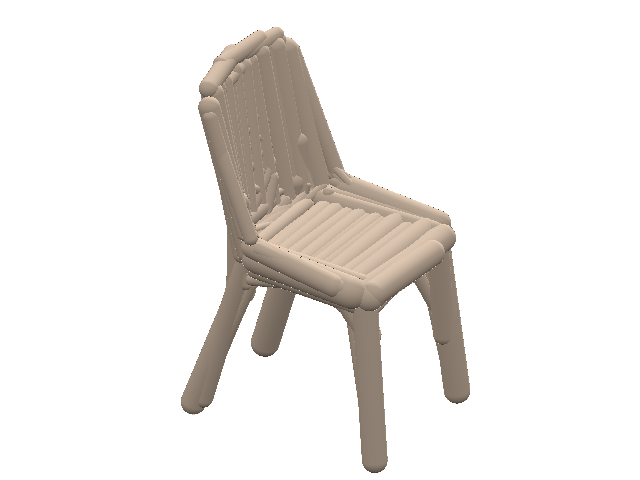}
\includegraphics[trim={5cm 0 5cm 0},clip,width=0.19\columnwidth]{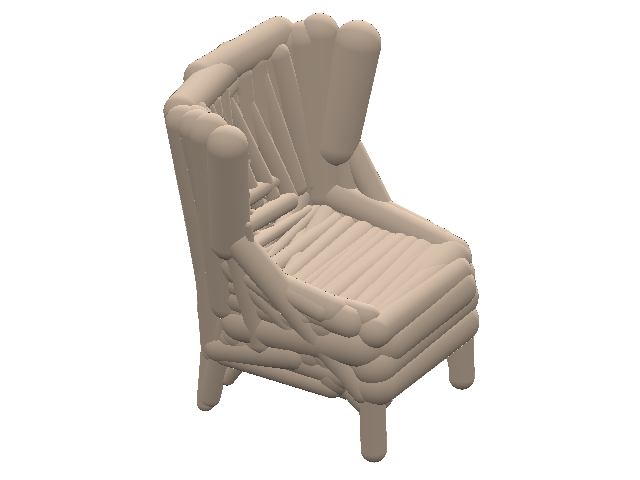}
\includegraphics[trim={5cm 0 5cm 0},clip,width=0.19\columnwidth]{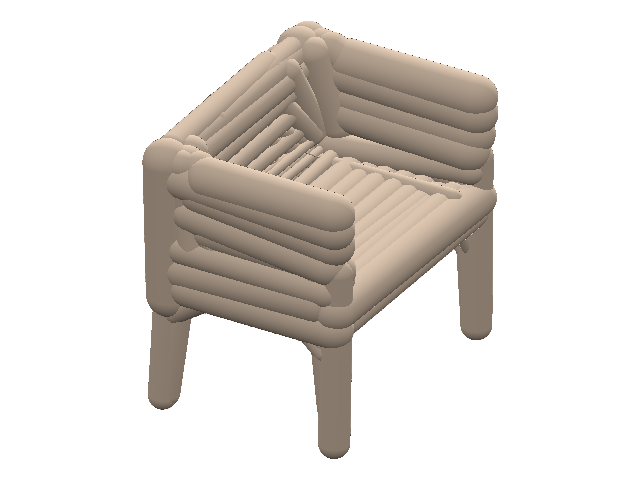}\\
\includegraphics[trim={5cm 0 5cm 0},clip,width=0.19\columnwidth]{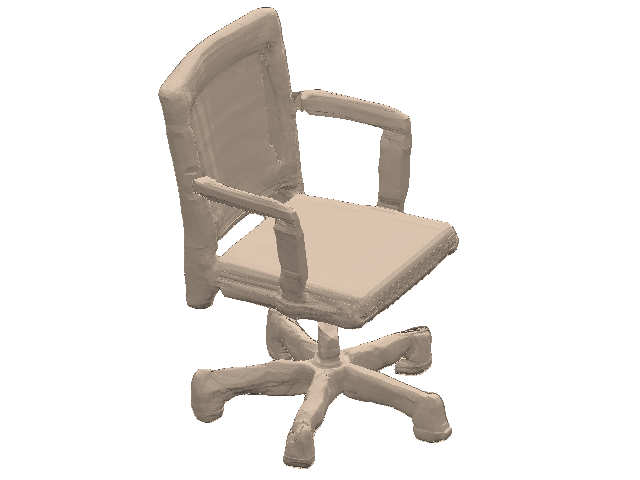}
\includegraphics[trim={5cm 0 5cm 0},clip,width=0.19\columnwidth]{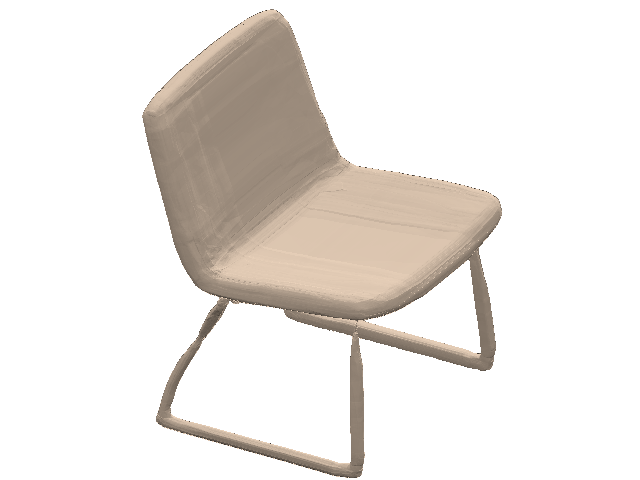}
\includegraphics[trim={5cm 0 5cm 0},clip,width=0.19\columnwidth]{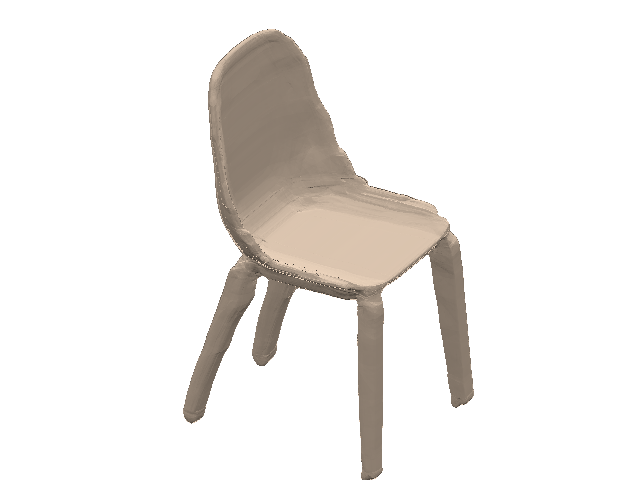}
\includegraphics[trim={5cm 0 5cm 0},clip,width=0.19\columnwidth]{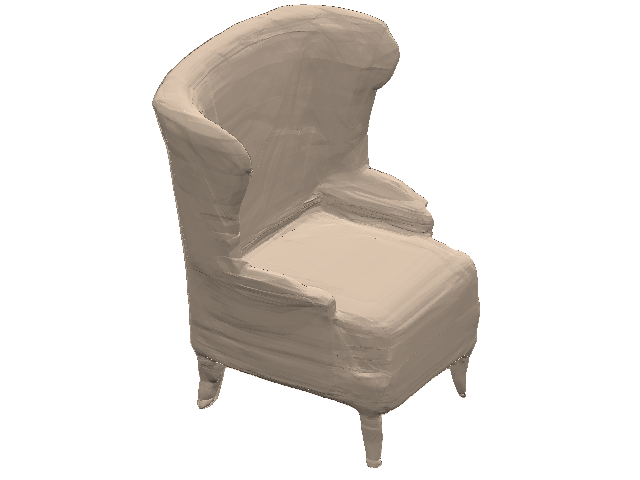}
\includegraphics[trim={5cm 0 5cm 0},clip,width=0.19\columnwidth]{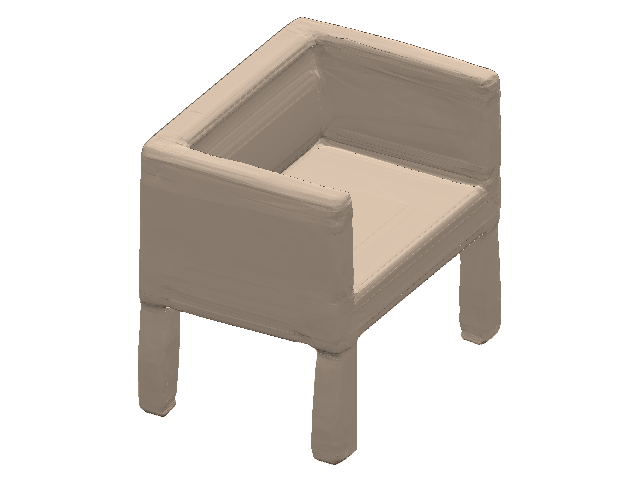}
\caption{Learning with other primitive types. Our technique is directly applicable for other shapes which can be represented with SDFs. Above we demonstrate shapes represented with capsule primitives (cylinders with rounded ends), and their corresponding high-resolution representation. }
\label{fig:DiffPrim}
\end{figure}

\medskip \noindent \textbf{Exploring other primitive types.}
While all of our results are illustrated on spherical shape primitives, our technique can directly incorporate other elementary shapes that can be represented with signed distance functions into the primitive-based representation. Figure \ref{fig:DiffPrim}, demonstrates a variant of our model that uses capsule primitives. We present the results with more primitive types in the supplementary material.

\subsection{Applications}
Our main application is shape manipulation using our coarse primitive-based representation as a proxy (see Section \ref{sec:manip}, Figures \ref{fig:teaser}-\ref{fig:JointLatent}, and many more examples in the supplementary material). In the following we speculate on several other applications enabled by our two-level representation. %

\medskip \noindent \textbf{Shape interpolation.} Similar to other generative models, our technique allows for a smooth interpolation between two real or generated shapes via interpolating the latent code. %
Furthermore, as an extension to our manipulation-through-optimization framework, our technique allows for \emph{controllable interpolation}, where instead of interpolating the black box latent code, we interpolate the primitive attributes in the coarse representation via optimization. This enables selective interpolation. For example, the user can specify to only interpolate the height of one chair to the height of the other chair.
Although this application is somewhat related to shape manipulation, there is one important distinction between the two: this application deals with two (or more) given shapes while shape manipulation deals with one shape only. In the supplementary material we demonstrate many interpolation results in both regular (latent space) and controllable (primitive attribute space) settings. %

\medskip \noindent \textbf{Shape correspondence.} As our primitives are semantically meaningful, we can also perform shape correspondence between the high resolution shapes via the coarse shape proxy. To do so, we map every point on the surface of the high-resolution shape to its closest primitive shape. 
Figure \ref{fig:correspondence} illustrates several corresponding regions over airplanes which are structurally different. %

\medskip \noindent \textbf{Real-time traversal and rendering.} Previous work has shown that perception can be improved by arranging results by similarity \cite{rodden2001does}. As the shape primitives can be rendered in real-time, our two-level representation allows for a real-time smooth exploration of the generative shape space. %
Once the user would like to ``zoom-in'' to a shape of interest, the system can render the slower high resolution model.

\section{Conclusions}

In this work, we have presented DualSDF, a novel 3D shape representation which represents shapes in two levels of granularities. We have shown that our fine-scale representation is highly expressive %
and that our coarse-scale primitive based representation learns a semantic decomposition, which is effective for shape manipulation. We have demonstrated that the two representations are tightly coupled, and thus modifications on the coarse-scale representation can be faithfully propagated to the fine-scale representation. Technically, we have formulated our shared latent space model with the variational autodecoder framework, which regularizes the latent space for better generation, manipulation and coupling. %

\section{Acknowledgements}

We would like to thank Abe Davis for his insightful feedback. This work was supported in part by grants from Facebook and by the generosity of Eric and Wendy Schmidt by recommendation of the Schmidt Futures program.
{\small

\begin{thebibliography}{10}\itemsep=-1pt

\bibitem{achlioptas2017learning}
Panos Achlioptas, Olga Diamanti, Ioannis Mitliagkas, and Leonidas Guibas.
\newblock Learning representations and generative models for 3d point clouds.
\newblock {\em arXiv preprint arXiv:1707.02392}, 2017.

\bibitem{anguelov2005scape}
Dragomir Anguelov, Praveen Srinivasan, Daphne Koller, Sebastian Thrun, Jim
  Rodgers, and James Davis.
\newblock Scape: shape completion and animation of people.
\newblock In {\em ACM transactions on graphics (TOG)}, volume~24, pages
  408--416. ACM, 2005.

\bibitem{bau2019semantic}
David Bau, Hendrik Strobelt, William Peebles, Jonas Wulff, Bolei Zhou, Jun-Yan
  Zhu, and Antonio Torralba.
\newblock Semantic photo manipulation with a generative image prior.
\newblock {\em ACM Transactions on Graphics (TOG)}, 38(4):1--11, 2019.

\bibitem{ben2018multi}
Heli Ben-Hamu, Haggai Maron, Itay Kezurer, Gal Avineri, and Yaron Lipman.
\newblock Multi-chart generative surface modeling.
\newblock In {\em SIGGRAPH Asia 2018 Technical Papers}, page 215. ACM, 2018.

\bibitem{benkHo2001algorithms}
P{\'a}l Benk{\H{o}}, Ralph~R Martin, and Tam{\'a}s V{\'a}rady.
\newblock Algorithms for reverse engineering boundary representation models.
\newblock {\em Computer-Aided Design}, 33(11):839--851, 2001.

\bibitem{biederman1987recognition}
Irving Biederman.
\newblock Recognition-by-components: a theory of human image understanding.
\newblock {\em Psychological review}, 94(2):115, 1987.

\bibitem{chang2015shapenet}
Angel~X Chang, Thomas Funkhouser, Leonidas Guibas, Pat Hanrahan, Qixing Huang,
  Zimo Li, Silvio Savarese, Manolis Savva, Shuran Song, Hao Su, et~al.
\newblock Shapenet: An information-rich 3d model repository.
\newblock {\em arXiv preprint arXiv:1512.03012}, 2015.

\bibitem{chen2017deeplab}
Liang-Chieh Chen, George Papandreou, Iasonas Kokkinos, Kevin Murphy, and Alan~L
  Yuille.
\newblock Deeplab: Semantic image segmentation with deep convolutional nets,
  atrous convolution, and fully connected crfs.
\newblock {\em IEEE transactions on pattern analysis and machine intelligence},
  40(4):834--848, 2017.

\bibitem{chen2019learning}
Zhiqin Chen and Hao Zhang.
\newblock Learning implicit fields for generative shape modeling.
\newblock In {\em Proceedings of the IEEE Conference on Computer Vision and
  Pattern Recognition}, pages 5939--5948, 2019.

\bibitem{deng2019cvxnets}
Boyang Deng, Kyle Genova, Soroosh Yazdani, Sofien Bouaziz, Geoffrey Hinton, and
  Andrea Tagliasacchi.
\newblock Cvxnets: Learnable convex decomposition.
\newblock {\em arXiv preprint arXiv:1909.05736}, 2019.

\bibitem{deprelle2019learning}
Theo Deprelle, Thibault Groueix, Matthew Fisher, Vladimir~G Kim, Bryan~C
  Russell, and Mathieu Aubry.
\newblock Learning elementary structures for 3d shape generation and matching.
\newblock {\em arXiv preprint arXiv:1908.04725}, 2019.

\bibitem{dorta2017laplacian}
Garoe Dorta, Sara Vicente, Lourdes Agapito, Neill~DF Campbell, Simon Prince,
  and Ivor Simpson.
\newblock Laplacian pyramid of conditional variational autoencoders.
\newblock In {\em Proceedings of the 14th European Conference on Visual Media
  Production (CVMP 2017)}, page~7. ACM, 2017.

\bibitem{fan2017point}
Haoqiang Fan, Hao Su, and Leonidas~J Guibas.
\newblock A point set generation network for 3d object reconstruction from a
  single image.
\newblock In {\em Proceedings of the IEEE conference on computer vision and
  pattern recognition}, pages 605--613, 2017.

\bibitem{farabet2012learning}
Clement Farabet, Camille Couprie, Laurent Najman, and Yann LeCun.
\newblock Learning hierarchical features for scene labeling.
\newblock {\em IEEE transactions on pattern analysis and machine intelligence},
  35(8):1915--1929, 2012.

\bibitem{frisken2006designing}
Sarah~F Frisken and Ronald~N Perry.
\newblock Designing with distance fields.
\newblock In {\em ACM SIGGRAPH 2006 Courses}, pages 60--66. ACM, 2006.

\bibitem{gal2007surface}
Ran Gal, Ariel Shamir, Tal Hassner, Mark Pauly, and Daniel Cohen-Or.
\newblock Surface reconstruction using local shape priors.
\newblock In {\em Symposium on Geometry Processing}, number CONF, pages
  253--262, 2007.

\bibitem{gao2019sparse}
Lin Gao, Yu-Kun Lai, Jie Yang, Zhang Ling-Xiao, Shihong Xia, and Leif Kobbelt.
\newblock Sparse data driven mesh deformation.
\newblock {\em IEEE transactions on visualization and computer graphics}, 2019.

\bibitem{gao2019sdm}
Lin Gao, Jie Yang, Tong Wu, Yu-Jie Yuan, Hongbo Fu, Yu-Kun Lai, and Hao Zhang.
\newblock Sdm-net: Deep generative network for structured deformable mesh.
\newblock {\em arXiv preprint arXiv:1908.04520}, 2019.

\bibitem{genova2019learning}
Kyle Genova, Forrester Cole, Daniel Vlasic, Aaron Sarna, William~T Freeman, and
  Thomas Funkhouser.
\newblock Learning shape templates with structured implicit functions.
\newblock {\em arXiv preprint arXiv:1904.06447}, 2019.

\bibitem{groueix2018atlasnet}
Thibault Groueix, Matthew Fisher, Vladimir Kim, Bryan Russell, and Mathieu
  Aubry.
\newblock Atlasnet: A papier-m{\^a}ch{\'e} approach to learning 3d surface
  generation.
\newblock In {\em CVPR 2018}, 2018.

\bibitem{gulrajani2016pixelvae}
Ishaan Gulrajani, Kundan Kumar, Faruk Ahmed, Adrien~Ali Taiga, Francesco Visin,
  David Vazquez, and Aaron Courville.
\newblock Pixelvae: A latent variable model for natural images.
\newblock {\em arXiv preprint arXiv:1611.05013}, 2016.

\bibitem{hao2018controllable}
Zekun Hao, Xun Huang, and Serge Belongie.
\newblock Controllable video generation with sparse trajectories.
\newblock In {\em Proceedings of the IEEE Conference on Computer Vision and
  Pattern Recognition}, pages 7854--7863, 2018.

\bibitem{hao2017scale}
Zekun Hao, Yu Liu, Hongwei Qin, Junjie Yan, Xiu Li, and Xiaolin Hu.
\newblock Scale-aware face detection.
\newblock In {\em Proceedings of the IEEE Conference on Computer Vision and
  Pattern Recognition}, pages 6186--6195, 2017.

\bibitem{he2015spatial}
Kaiming He, Xiangyu Zhang, Shaoqing Ren, and Jian Sun.
\newblock Spatial pyramid pooling in deep convolutional networks for visual
  recognition.
\newblock {\em IEEE transactions on pattern analysis and machine intelligence},
  37(9):1904--1916, 2015.

\bibitem{hegde2016fusionnet}
Vishakh Hegde and Reza Zadeh.
\newblock Fusionnet: 3d object classification using multiple data
  representations.
\newblock {\em arXiv preprint arXiv:1607.05695}, 2016.

\bibitem{huang2017stacked}
Xun Huang, Yixuan Li, Omid Poursaeed, John Hopcroft, and Serge Belongie.
\newblock Stacked generative adversarial networks.
\newblock In {\em Proceedings of the IEEE Conference on Computer Vision and
  Pattern Recognition}, pages 5077--5086, 2017.

\bibitem{karras2019style}
Tero Karras, Samuli Laine, and Timo Aila.
\newblock A style-based generator architecture for generative adversarial
  networks.
\newblock In {\em Proceedings of the IEEE Conference on Computer Vision and
  Pattern Recognition}, pages 4401--4410, 2019.

\bibitem{karras2019analyzing}
Tero Karras, Samuli Laine, Miika Aittala, Janne Hellsten, Jaakko Lehtinen, and
  Timo Aila.
\newblock Analyzing and improving the image quality of stylegan.
\newblock {\em arXiv preprint arXiv:1912.04958}, 2019.

\bibitem{kingma2013auto}
Diederik~P Kingma and Max Welling.
\newblock Auto-encoding variational bayes.
\newblock {\em arXiv preprint arXiv:1312.6114}, 2013.

\bibitem{lewis2000pose}
John~P Lewis, Matt Cordner, and Nickson Fong.
\newblock Pose space deformation: a unified approach to shape interpolation and
  skeleton-driven deformation.
\newblock In {\em Proceedings of the 27th annual conference on Computer
  graphics and interactive techniques}, pages 165--172. ACM
  Press/Addison-Wesley Publishing Co., 2000.

\bibitem{li2018point}
Chun-Liang Li, Manzil Zaheer, Yang Zhang, Barnabas Poczos, and Ruslan
  Salakhutdinov.
\newblock Point cloud gan.
\newblock {\em arXiv preprint arXiv:1810.05795}, 2018.

\bibitem{litany2018deformable}
Or Litany, Alex Bronstein, Michael Bronstein, and Ameesh Makadia.
\newblock Deformable shape completion with graph convolutional autoencoders.
\newblock In {\em Proceedings of the IEEE Conference on Computer Vision and
  Pattern Recognition}, pages 1886--1895, 2018.

\bibitem{magnenat1988joint}
Nadia Magnenat-Thalmann, Richard Laperrire, and Daniel Thalmann.
\newblock Joint-dependent local deformations for hand animation and object
  grasping.
\newblock In {\em In Proceedings on Graphics interface’88}. Citeseer, 1988.

\bibitem{mescheder2019occupancy}
Lars Mescheder, Michael Oechsle, Michael Niemeyer, Sebastian Nowozin, and
  Andreas Geiger.
\newblock Occupancy networks: Learning 3d reconstruction in function space.
\newblock In {\em Proceedings of the IEEE Conference on Computer Vision and
  Pattern Recognition}, pages 4460--4470, 2019.

\bibitem{mo2019partnet}
Kaichun Mo, Shilin Zhu, Angel~X Chang, Li Yi, Subarna Tripathi, Leonidas~J
  Guibas, and Hao Su.
\newblock Partnet: A large-scale benchmark for fine-grained and hierarchical
  part-level 3d object understanding.
\newblock In {\em Proceedings of the IEEE Conference on Computer Vision and
  Pattern Recognition}, pages 909--918, 2019.

\bibitem{muralikrishnan2019shape}
Sanjeev Muralikrishnan, Vladimir~G Kim, Matthew Fisher, and Siddhartha
  Chaudhuri.
\newblock Shape unicode: A unified shape representation.
\newblock In {\em Proceedings of the IEEE Conference on Computer Vision and
  Pattern Recognition}, pages 3790--3799, 2019.

\bibitem{nooruddin2003simplification}
Fakir~S. Nooruddin and Greg Turk.
\newblock Simplification and repair of polygonal models using volumetric
  techniques.
\newblock {\em IEEE Transactions on Visualization and Computer Graphics},
  9(2):191--205, 2003.

\bibitem{park2019deepsdf}
Jeong~Joon Park, Peter Florence, Julian Straub, Richard Newcombe, and Steven
  Lovegrove.
\newblock Deepsdf: Learning continuous signed distance functions for shape
  representation.
\newblock In {\em Proceedings of the IEEE Conference on Computer Vision and
  Pattern Recognition}, pages 165--174, 2019.

\bibitem{paschalidou2019superquadrics}
Despoina Paschalidou, Ali~Osman Ulusoy, and Andreas Geiger.
\newblock Superquadrics revisited: Learning 3d shape parsing beyond cuboids.
\newblock In {\em Proceedings of the IEEE Conference on Computer Vision and
  Pattern Recognition}, pages 10344--10353, 2019.

\bibitem{pons2015dyna}
Gerard Pons-Moll, Javier Romero, Naureen Mahmood, and Michael~J Black.
\newblock Dyna: A model of dynamic human shape in motion.
\newblock {\em ACM Transactions on Graphics (TOG)}, 34(4):120, 2015.

\bibitem{roberts1963machine}
Lawrence~G Roberts.
\newblock {\em Machine perception of three-dimensional solids}.
\newblock PhD thesis, Massachusetts Institute of Technology, 1963.

\bibitem{rodden2001does}
Kerry Rodden, Wojciech Basalaj, David Sinclair, and Kenneth Wood.
\newblock Does organisation by similarity assist image browsing?
\newblock In {\em Proceedings of the SIGCHI conference on Human factors in
  computing systems}, pages 190--197. ACM, 2001.

\bibitem{schnabel2009completion}
Ruwen Schnabel, Patrick Degener, and Reinhard Klein.
\newblock Completion and reconstruction with primitive shapes.
\newblock In {\em Computer Graphics Forum}, volume~28, pages 503--512. Wiley
  Online Library, 2009.

\bibitem{schnabel2007efficient}
Ruwen Schnabel, Roland Wahl, and Reinhard Klein.
\newblock Efficient ransac for point-cloud shape detection.
\newblock In {\em Computer graphics forum}, volume~26, pages 214--226. Wiley
  Online Library, 2007.

\bibitem{seitz2006comparison}
Steven~M Seitz, Brian Curless, James Diebel, Daniel Scharstein, and Richard
  Szeliski.
\newblock A comparison and evaluation of multi-view stereo reconstruction
  algorithms.
\newblock In {\em 2006 IEEE computer society conference on computer vision and
  pattern recognition (CVPR'06)}, volume~1, pages 519--528. IEEE, 2006.

\bibitem{smirnov2019deep}
Dmitriy Smirnov, Matthew Fisher, Vladimir~G Kim, Richard Zhang, and Justin
  Solomon.
\newblock Deep parametric shape predictions using distance fields.
\newblock {\em arXiv preprint arXiv:1904.08921}, 2019.

\bibitem{su2018splatnet}
Hang Su, Varun Jampani, Deqing Sun, Subhransu Maji, Evangelos Kalogerakis,
  Ming-Hsuan Yang, and Jan Kautz.
\newblock Splatnet: Sparse lattice networks for point cloud processing.
\newblock In {\em Proceedings of the IEEE Conference on Computer Vision and
  Pattern Recognition}, pages 2530--2539, 2018.

\bibitem{sun2020pointgrow}
Yongbin Sun, Yue Wang, Ziwei Liu, Joshua Siegel, and Sanjay Sarma.
\newblock Pointgrow: Autoregressively learned point cloud generation with
  self-attention.
\newblock In {\em The IEEE Winter Conference on Applications of Computer
  Vision}, pages 61--70, 2020.

\bibitem{tan2018variational}
Qingyang Tan, Lin Gao, Yu-Kun Lai, and Shihong Xia.
\newblock Variational autoencoders for deforming 3d mesh models.
\newblock In {\em Proceedings of the IEEE Conference on Computer Vision and
  Pattern Recognition}, pages 5841--5850, 2018.

\bibitem{tulsiani2017learning}
Shubham Tulsiani, Hao Su, Leonidas~J Guibas, Alexei~A Efros, and Jitendra
  Malik.
\newblock Learning shape abstractions by assembling volumetric primitives.
\newblock In {\em Proceedings of the IEEE Conference on Computer Vision and
  Pattern Recognition}, pages 2635--2643, 2017.

\bibitem{williams2019deep}
Francis Williams, Teseo Schneider, Claudio Silva, Denis Zorin, Joan Bruna, and
  Daniele Panozzo.
\newblock Deep geometric prior for surface reconstruction.
\newblock In {\em Proceedings of the IEEE Conference on Computer Vision and
  Pattern Recognition}, pages 10130--10139, 2019.

\bibitem{yang2019pointflow}
Guandao Yang, Xun Huang, Zekun Hao, Ming-Yu Liu, Serge Belongie, and Bharath
  Hariharan.
\newblock Pointflow: 3d point cloud generation with continuous normalizing
  flows.
\newblock In {\em Proceedings of the IEEE International Conference on Computer
  Vision}, pages 4541--4550, 2019.

\bibitem{zadeh2019variational}
Amir Zadeh, Yao-Chong Lim, Paul~Pu Liang, and Louis-Philippe Morency.
\newblock Variational auto-decoder: Neural generative modeling from partial
  data.
\newblock {\em arXiv preprint arXiv:1903.00840}, 2019.

\bibitem{zamorski2018adversarial}
Maciej Zamorski, Maciej Zi{\k{e}}ba, Rafa{\l} Nowak, Wojciech Stokowiec, and
  Tomasz Trzci{\'n}ski.
\newblock Adversarial autoencoders for generating 3d point clouds.
\newblock {\em arXiv preprint arXiv:1811.07605}, 2018.

\bibitem{zhang2017stackgan}
Han Zhang, Tao Xu, Hongsheng Li, Shaoting Zhang, Xiaogang Wang, Xiaolei Huang,
  and Dimitris~N Metaxas.
\newblock Stackgan: Text to photo-realistic image synthesis with stacked
  generative adversarial networks.
\newblock In {\em Proceedings of the IEEE International Conference on Computer
  Vision}, pages 5907--5915, 2017.

\bibitem{zhu2016generative}
Jun-Yan Zhu, Philipp Kr{\"a}henb{\"u}hl, Eli Shechtman, and Alexei~A Efros.
\newblock Generative visual manipulation on the natural image manifold.
\newblock In {\em European Conference on Computer Vision}, pages 597--613.
  Springer, 2016.

\end{thebibliography}

}
\newpage
\clearpage
\appendix

\twocolumn[{%
\begin{minipage}{\textwidth}
   \null
   \vspace*{0.375in}
   \begin{center}
      {\Large \bf Supplementary Material for\\DualSDF: Semantic Shape Manipulation using a Two-Level Representation \par}
   \end{center}
   \vspace*{0.375in}
\end{minipage}
}]

\begin{figure}
\newcommand{\tl}{3.5cm}
\newcommand{\tr}{3.5cm}
\centering%
\includegraphics[trim={{\tl} 0 {\tr} 0},clip,width=0.192\columnwidth]{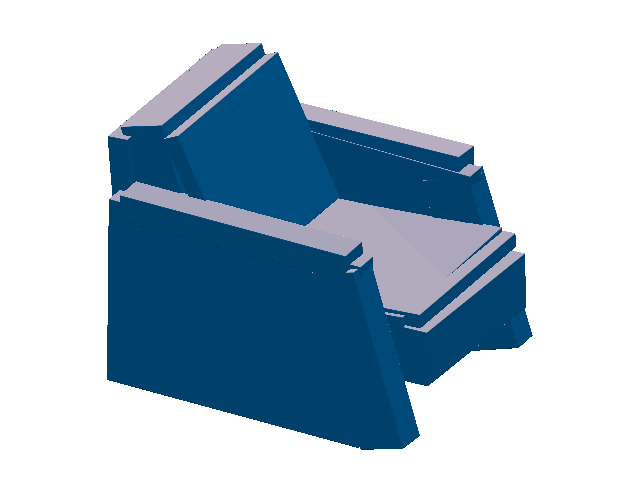}
\includegraphics[trim={{\tl} 0 {\tr} 0},clip,width=0.192\columnwidth]{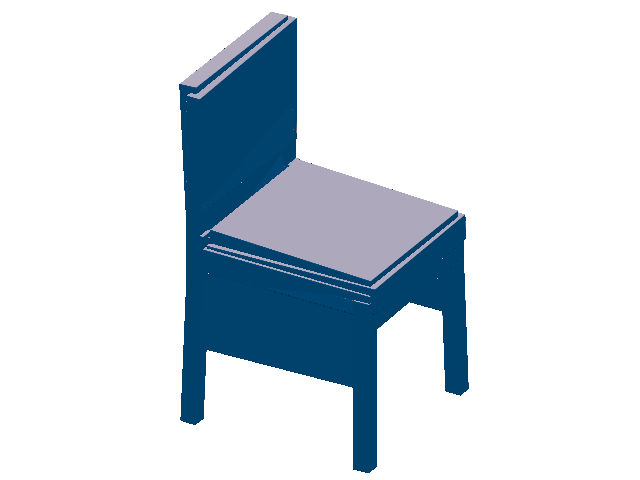}
\includegraphics[trim={{\tl} 0 {\tr} 0},clip,width=0.192\columnwidth]{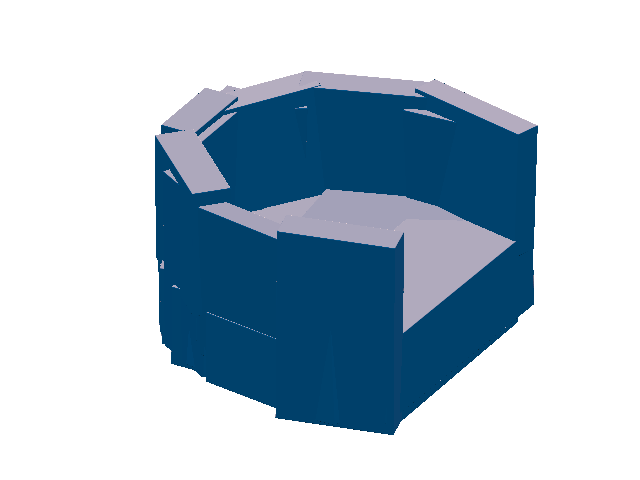}
\includegraphics[trim={{\tl} 0 {\tr} 0},clip,width=0.192\columnwidth]{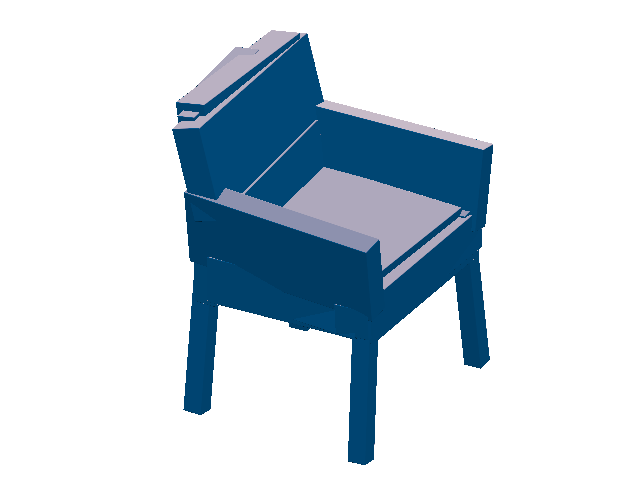}
\includegraphics[trim={{\tl} 0 {\tr} 0},clip,width=0.192\columnwidth]{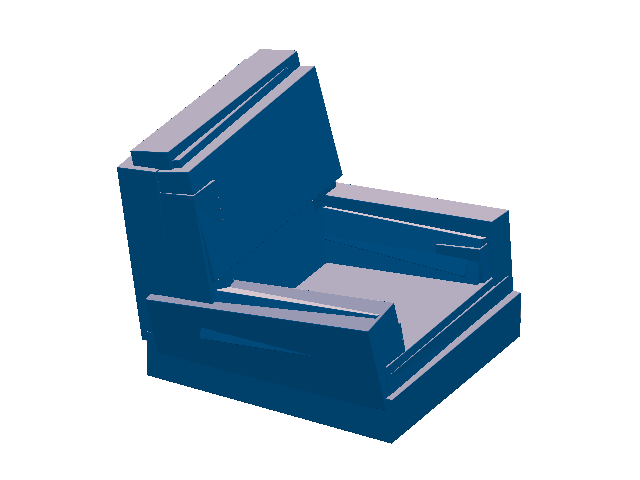} \\
\includegraphics[trim={{\tl} 0 {\tr} 0},clip,width=0.192\columnwidth]{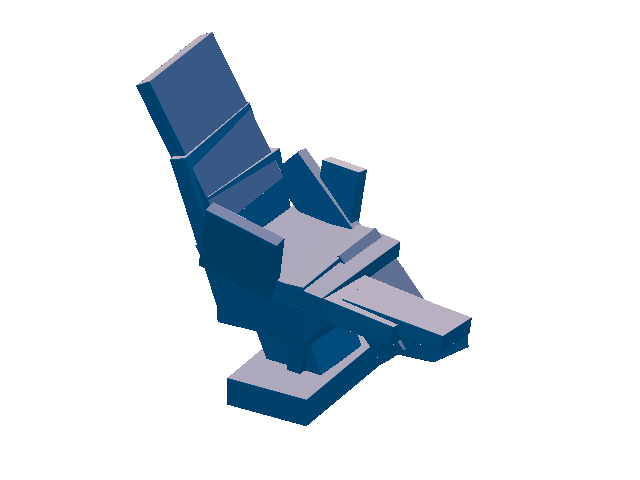}
\includegraphics[trim={{\tl} 0 {\tr} 0},clip,width=0.192\columnwidth]{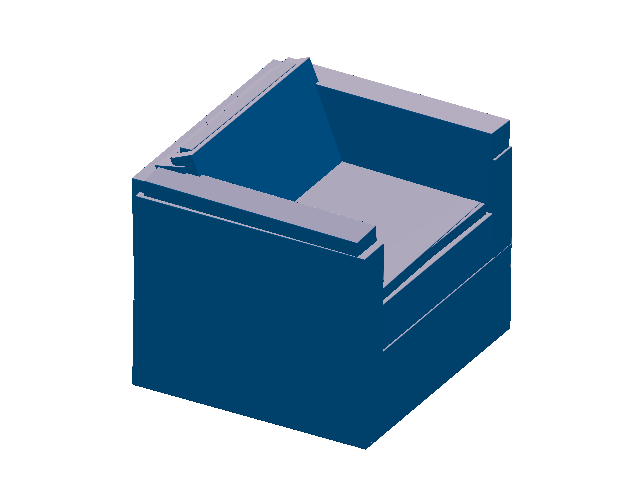}
\includegraphics[trim={{\tl} 0 {\tr} 0},clip,width=0.192\columnwidth]{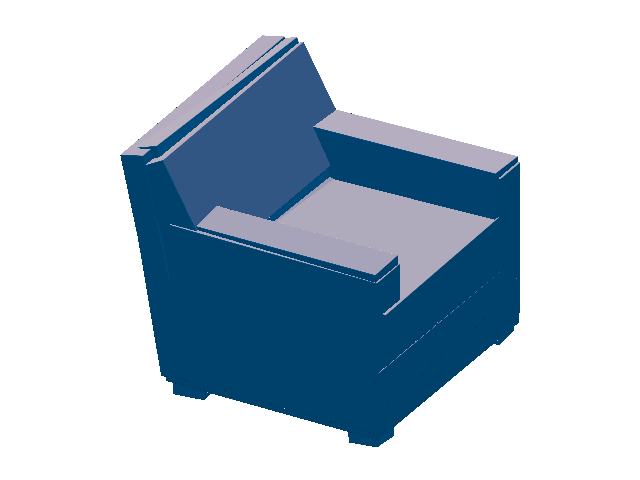}
\includegraphics[trim={{\tl} 0 {\tr} 0},clip,width=0.192\columnwidth]{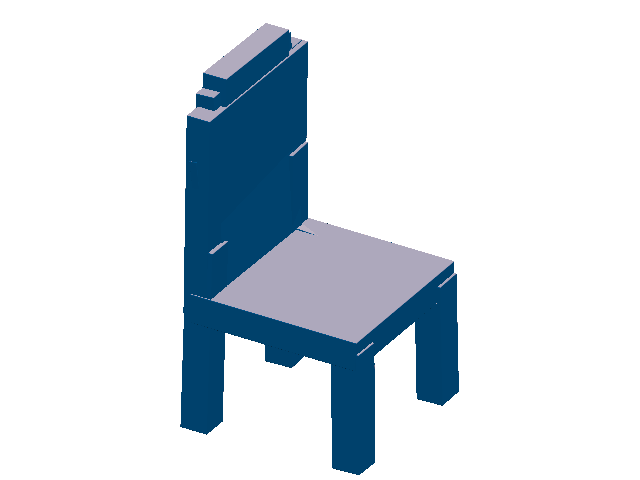}
\includegraphics[trim={{\tl} 0 {\tr} 0},clip,width=0.192\columnwidth]{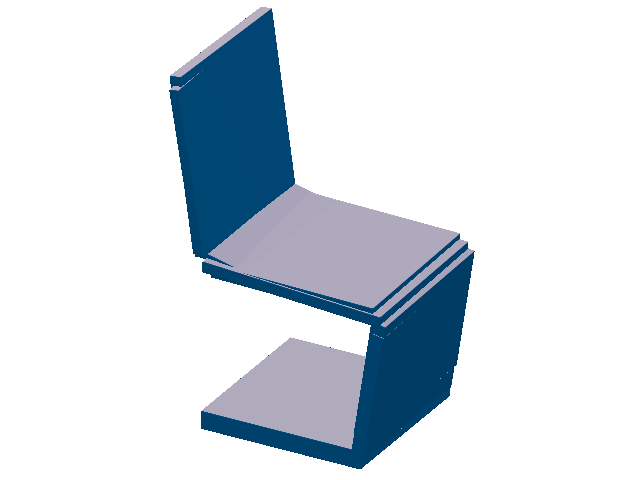} \\
\includegraphics[trim={{\tl} 0 {\tr} 0},clip,width=0.192\columnwidth]{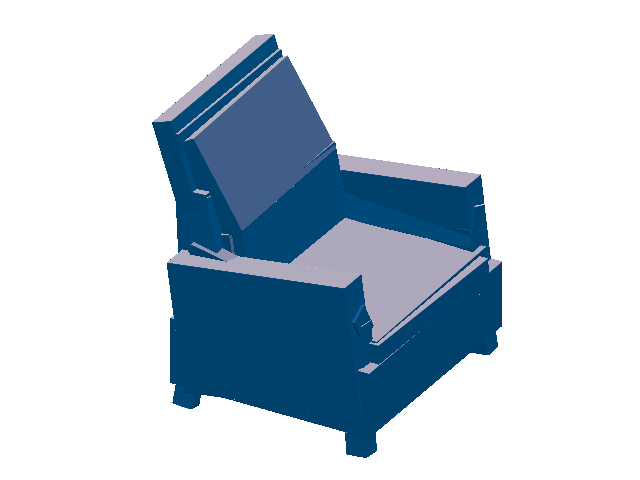}
\includegraphics[trim={{\tl} 0 {\tr} 0},clip,width=0.192\columnwidth]{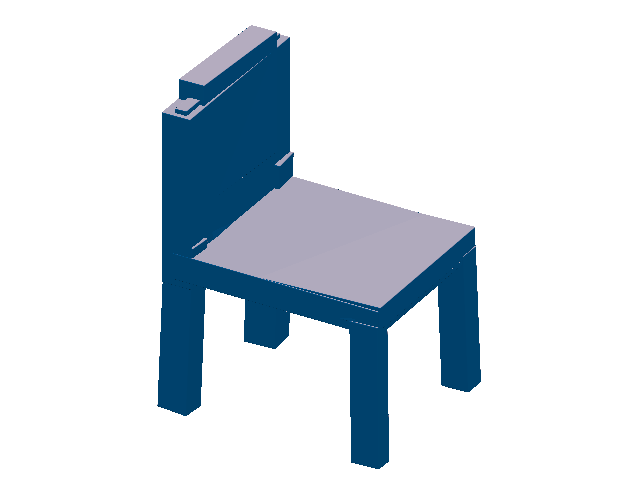}
\includegraphics[trim={{\tl} 0 {\tr} 0},clip,width=0.192\columnwidth]{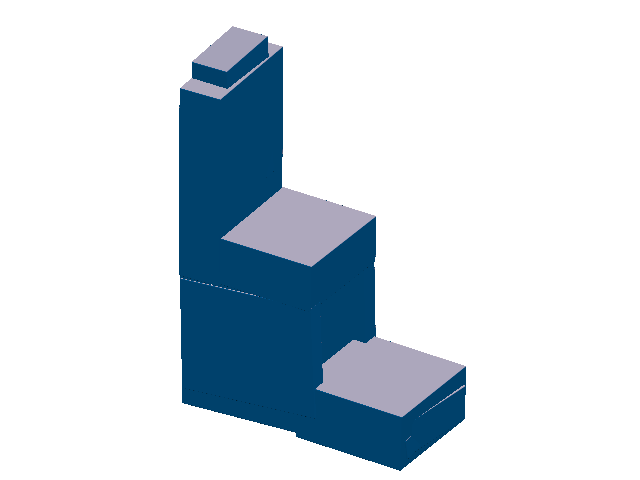}
\includegraphics[trim={{\tl} 0 {\tr} 0},clip,width=0.192\columnwidth]{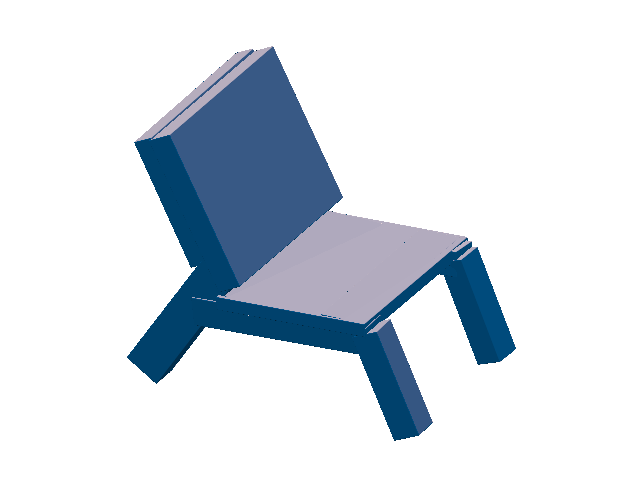}
\includegraphics[trim={{\tl} 0 {\tr} 0},clip,width=0.192\columnwidth]{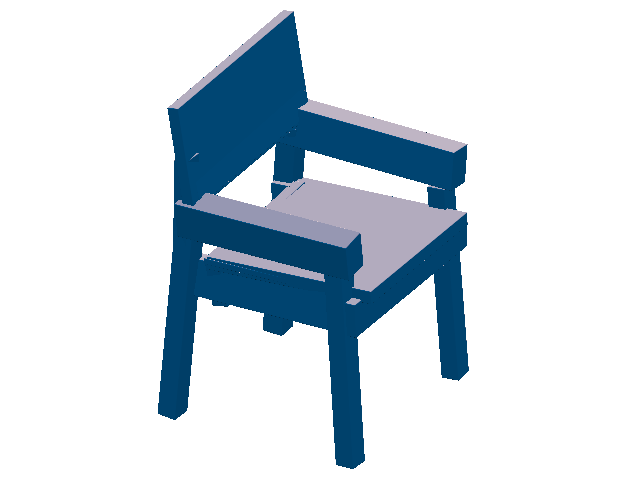} \\
\includegraphics[trim={{\tl} 0 {\tr} 0},clip,width=0.192\columnwidth]{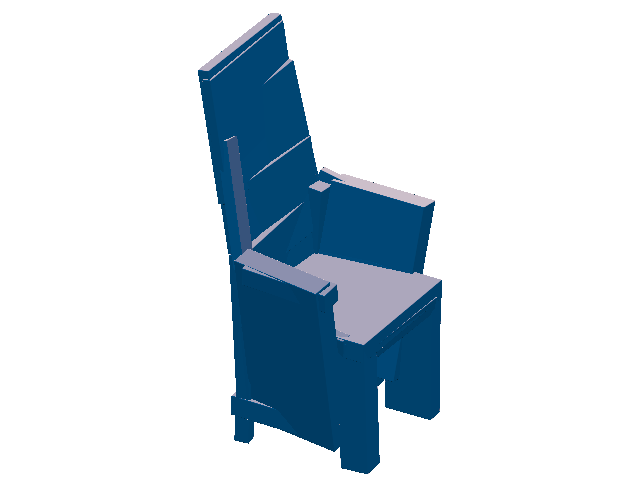}
\includegraphics[trim={{\tl} 0 {\tr} 0},clip,width=0.192\columnwidth]{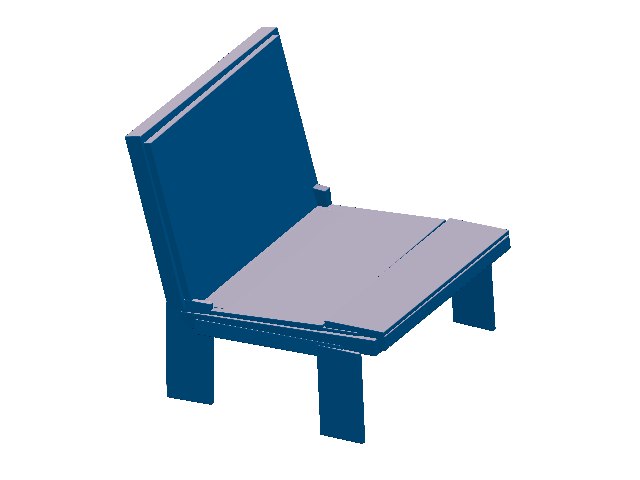}
\includegraphics[trim={{\tl} 0 {\tr} 0},clip,width=0.192\columnwidth]{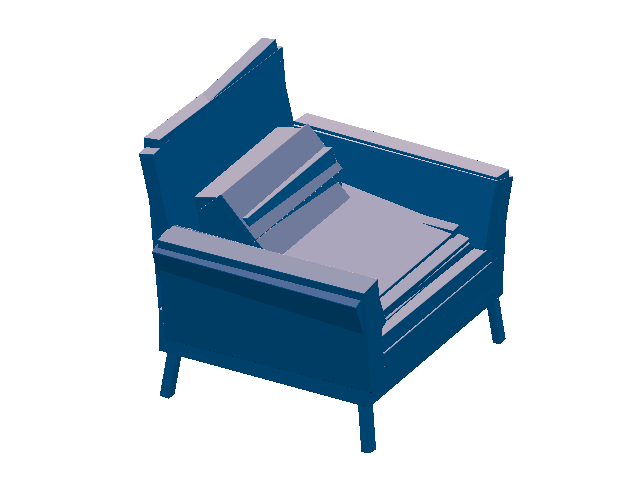}
\includegraphics[trim={{\tl} 0 {\tr} 0},clip,width=0.192\columnwidth]{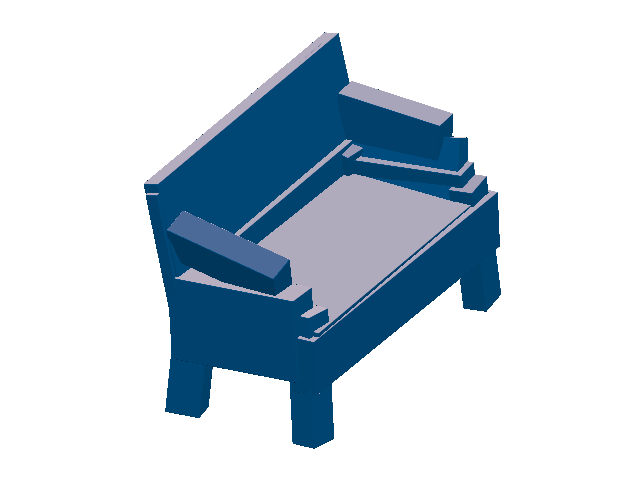}
\includegraphics[trim={{\tl} 0 {\tr} 0},clip,width=0.192\columnwidth]{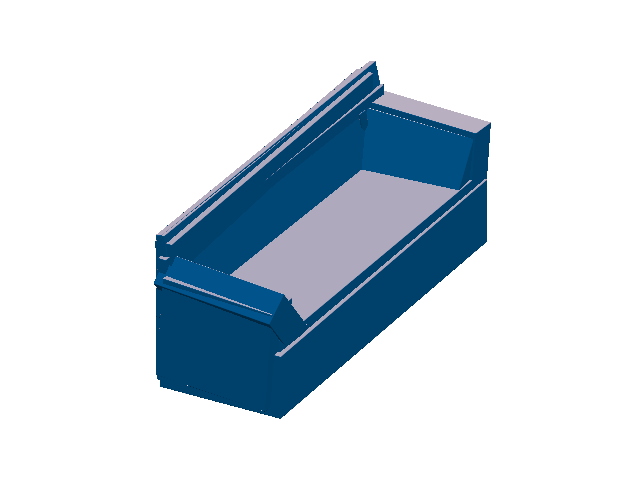}
\caption{Learning with box primitives. Our technique is directly applicable for geometric shapes which can be represented with SDFs. Above we demonstrate coarse shape reconstructions learned by our model with box primitives.  }
\label{fig:DiffPrim}
\end{figure}
\section{DualSDF with Box Primitives}
In our work, we use sphere primitives for our coarse representation. In the main paper, we also show reconstruction results obtained with capsule primitives.
In Figure \ref{fig:DiffPrim}, we demonstrate that box primitives can be utilized in our framework as well. In fact, our framework is very flexible in terms of primitive choice. Any primitive that can be represented with signed distance function can be incorporated into the coarse representation.

\begin{figure}[t]
\begin{center}
   \includegraphics[trim=0.5cm 0.5cm 0.5cm 0.5cm ,clip,width=0.6\linewidth]{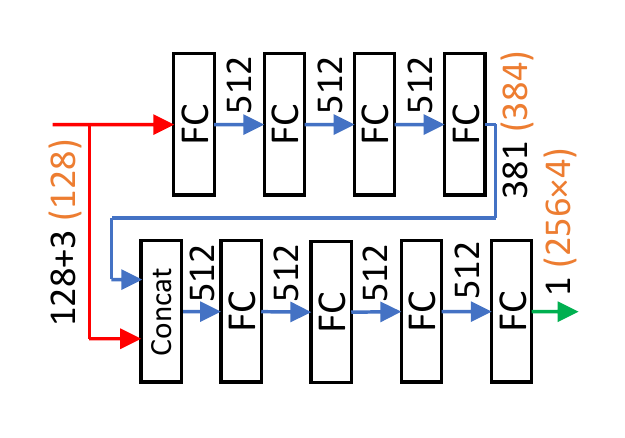}
\end{center}
   \caption{Network structure. The different settings used for the primitive-based representation are marked by parenthesis.}
\label{fig:sdfnet}
\end{figure}

\begin{figure*}[t]
    \begin{center}
    \newcommand{\sizea}{0.12\linewidth}
    \newcommand{\tal}{1cm}
    \newcommand{\tab}{0cm}
    \newcommand{\tar}{1cm}
    \newcommand{\tat}{0cm}
    \setlength{\tabcolsep}{0pt}
    \setlength{\fboxrule}{2pt}
    \renewcommand{\arraystretch}{0}
    \begin{tabular}{ccccccccc}
         \parbox[t]{4mm}{\multirow{3}{*}[11ex]{\rotatebox[origin=c]{90}{\textbf{With} Code Regularization}}}
         &\fbox{\includegraphics[width=\sizea, trim={\tal} {\tab} {\tar} {\tat},clip]{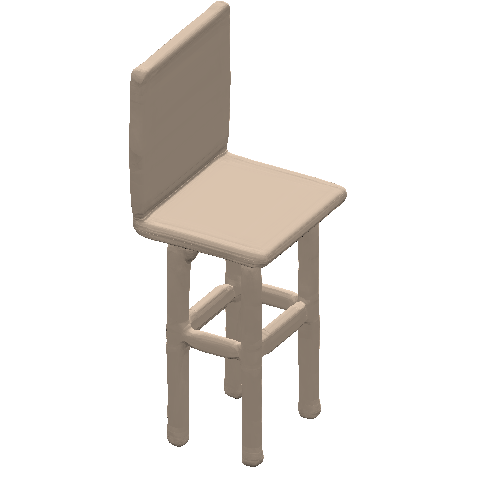}}& 
        \includegraphics[width=\sizea, trim={\tal} {\tab} {\tar} {\tat},clip]{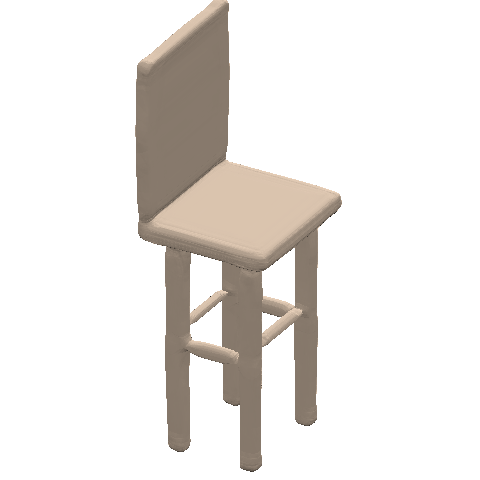}& 
        \includegraphics[width=\sizea, trim={\tal} {\tab} {\tar} {\tat},clip]{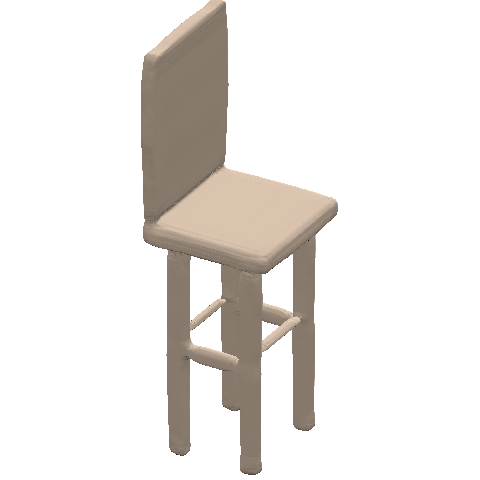}& 
        \includegraphics[width=\sizea, trim={\tal} {\tab} {\tar} {\tat},clip]{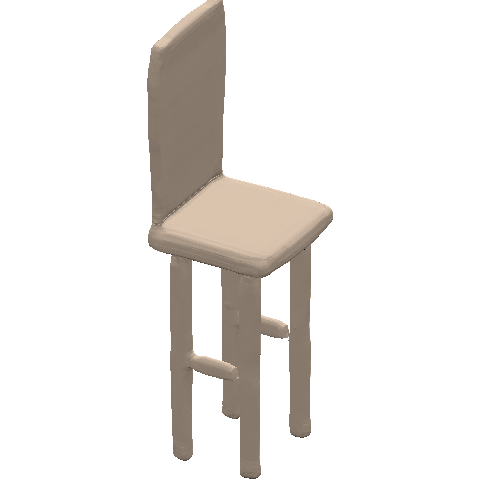}& 
        \includegraphics[width=\sizea, trim={\tal} {\tab} {\tar} {\tat},clip]{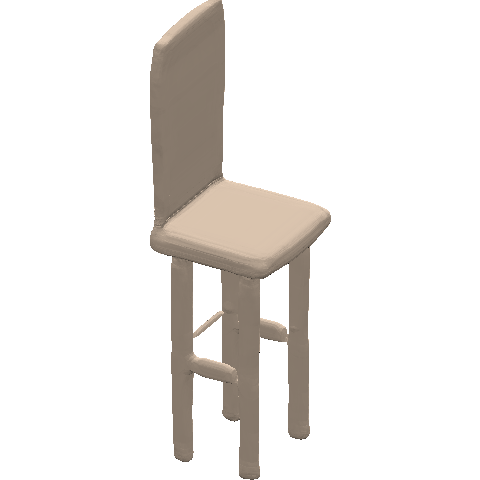}& 
        \includegraphics[width=\sizea, trim={\tal} {\tab} {\tar} {\tat},clip]{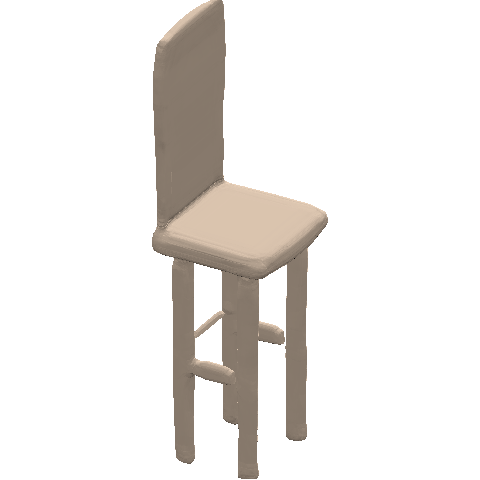}& 
        \includegraphics[width=\sizea, trim={\tal} {\tab} {\tar} {\tat},clip]{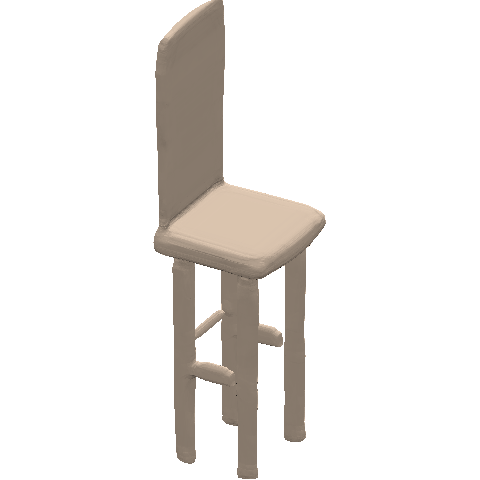}& 
        \includegraphics[width=\sizea, trim={\tal} {\tab} {\tar}
        {\tat},clip]{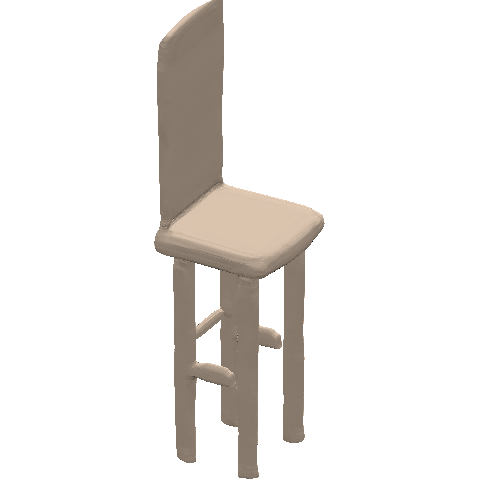}
        \\
        &\fbox{\includegraphics[width=\sizea, trim={\tal} {\tab} {\tar} {\tat},clip]{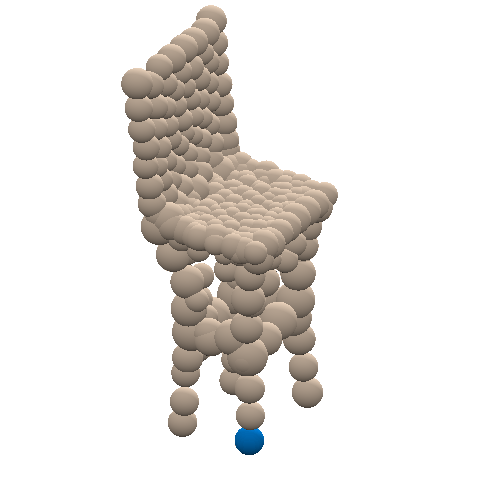}}& 
        \includegraphics[width=\sizea, trim={\tal} {\tab} {\tar} {\tat},clip]{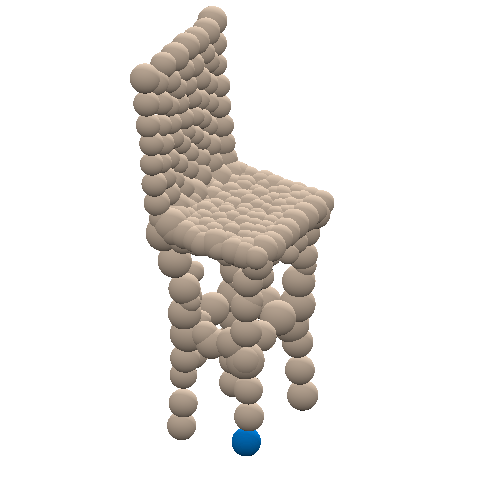}& 
        \includegraphics[width=\sizea, trim={\tal} {\tab} {\tar} {\tat},clip]{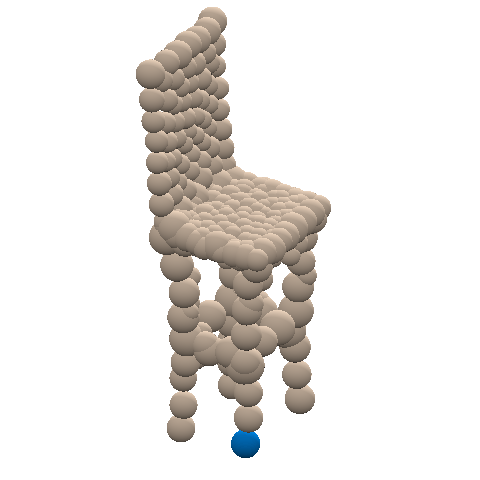}& 
        \includegraphics[width=\sizea, trim={\tal} {\tab} {\tar} {\tat},clip]{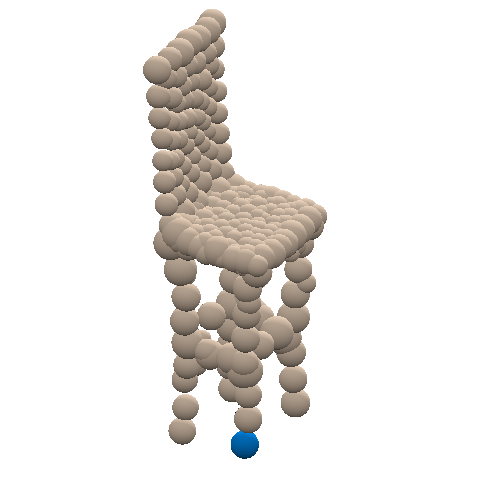}& 
        \includegraphics[width=\sizea, trim={\tal} {\tab} {\tar} {\tat},clip]{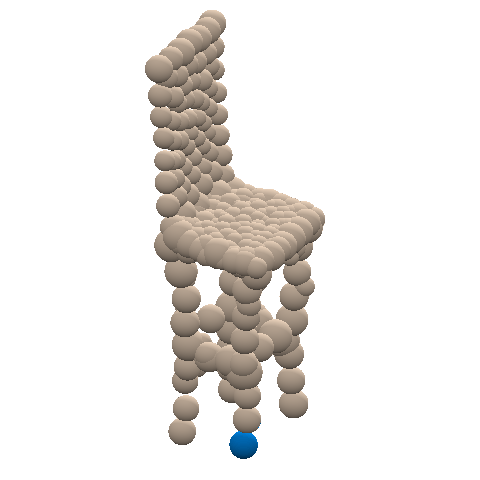}& 
        \includegraphics[width=\sizea, trim={\tal} {\tab} {\tar} {\tat},clip]{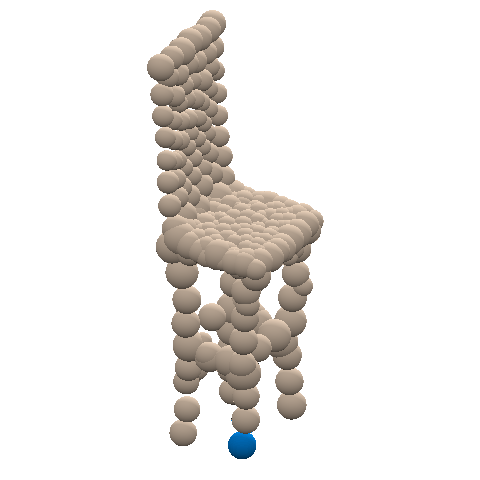}& 
        \includegraphics[width=\sizea, trim={\tal} {\tab} {\tar} {\tat},clip]{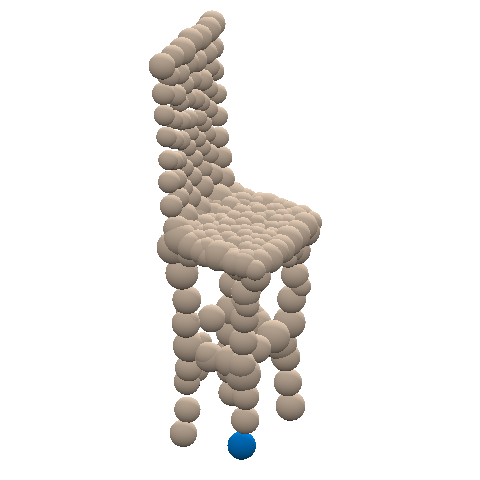}& 
        \includegraphics[width=\sizea, trim={\tal} {\tab} {\tar} {\tat},clip]{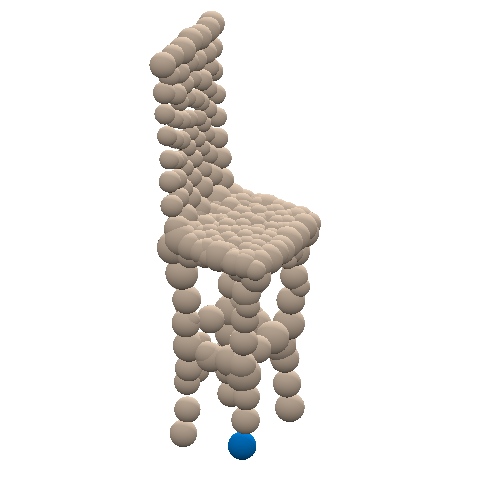}
        \\
        \hdashline
        \parbox[t]{4mm}{\multirow{2}{*}[12ex]{\rotatebox[origin=c]{90}{\textbf{Without} Code Regularization}}}
        &\fbox{\includegraphics[width=\sizea, trim={\tal} {\tab} {\tar} {\tat},clip]{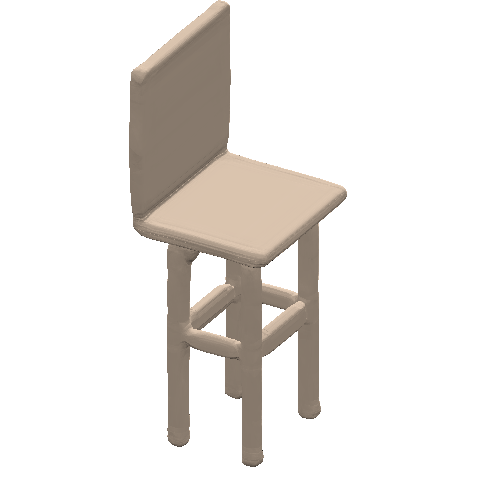}}& 
        \includegraphics[width=\sizea, trim={\tal} {\tab} {\tar} {\tat},clip]{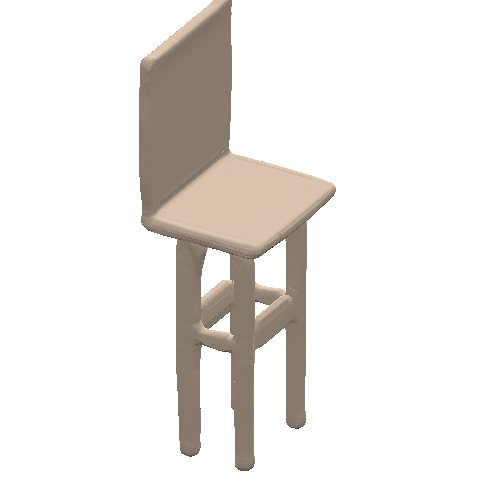}& 
        \includegraphics[width=\sizea, trim={\tal} {\tab} {\tar} {\tat},clip]{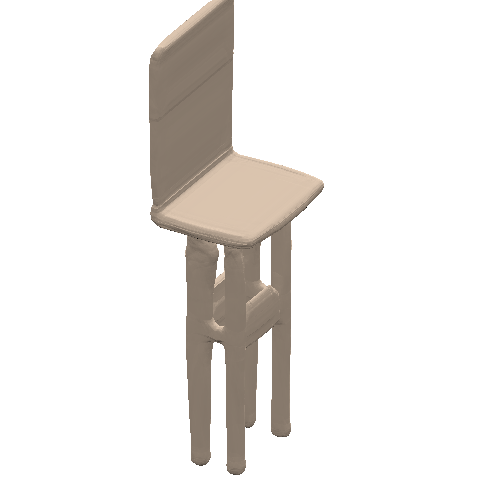}& 
        \includegraphics[width=\sizea, trim={\tal} {\tab} {\tar} {\tat},clip]{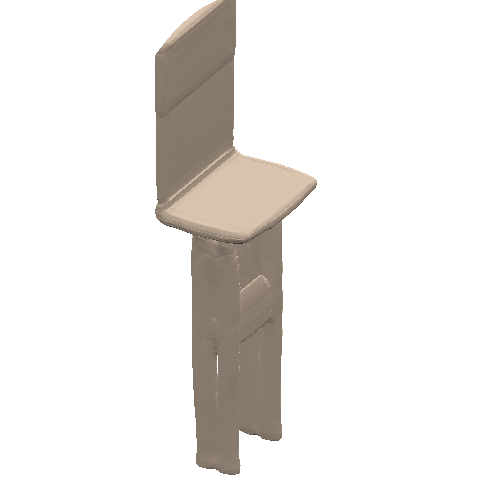}&
        \includegraphics[width=\sizea, trim={\tal} {\tab} {\tar} {\tat},clip]{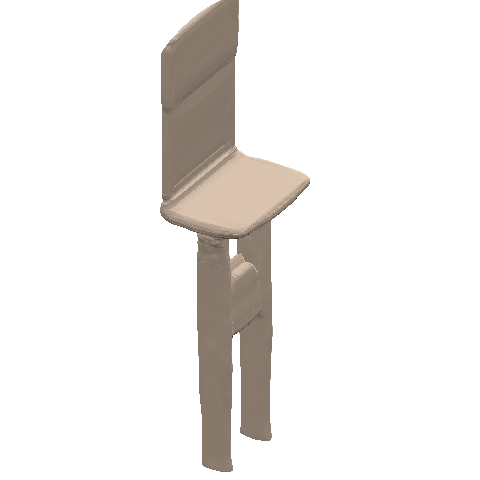}& 
        \includegraphics[width=\sizea, trim={\tal} {\tab} {\tar} {\tat},clip]{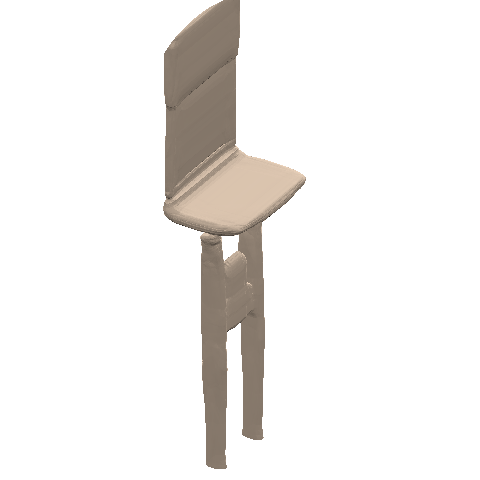}& 
        \includegraphics[width=\sizea, trim={\tal} {\tab} {\tar} {\tat},clip]{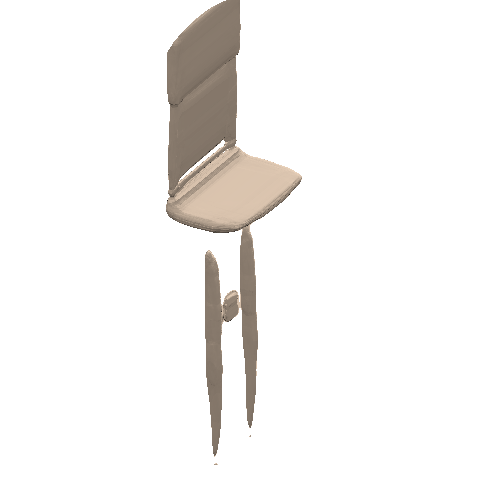}& 
        \includegraphics[width=\sizea, trim={\tal} {\tab} {\tar} {\tat},clip]{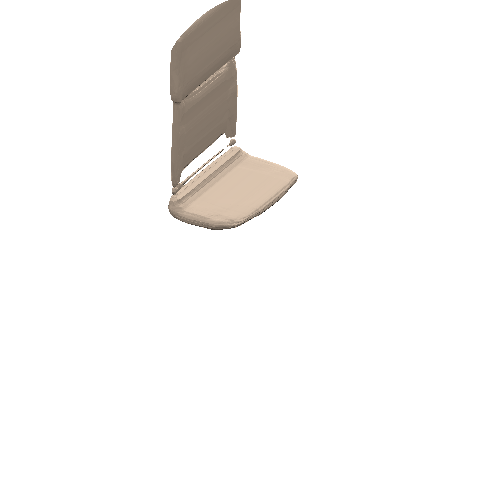}
        \\
        &\fbox{\includegraphics[width=\sizea, trim={\tal} {\tab} {\tar} {\tat},clip]{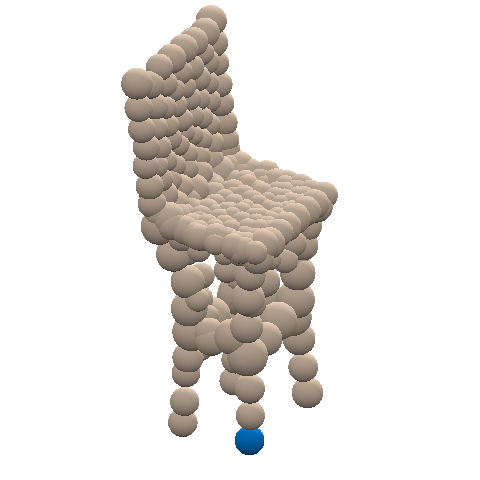}}& 
        \includegraphics[width=\sizea, trim={\tal} {\tab} {\tar} {\tat},clip]{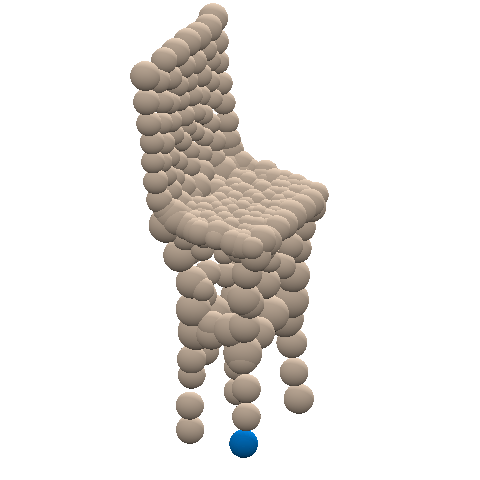}& 
        \includegraphics[width=\sizea, trim={\tal} {\tab} {\tar} {\tat},clip]{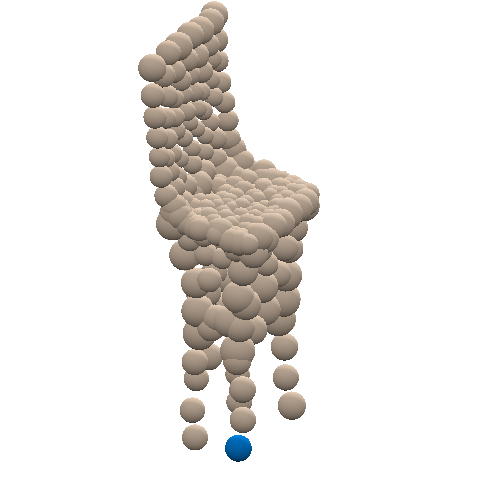}& 
        \includegraphics[width=\sizea, trim={\tal} {\tab} {\tar} {\tat},clip]{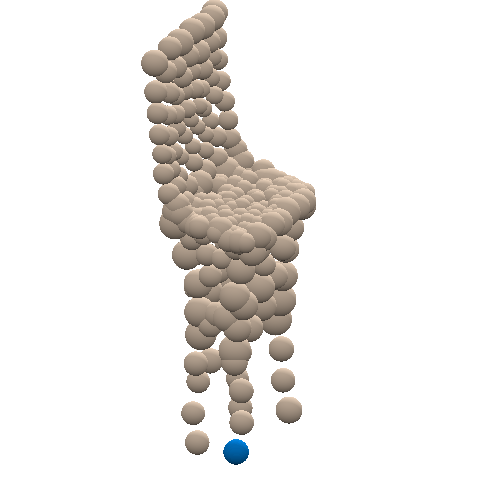}&
        \includegraphics[width=\sizea, trim={\tal} {\tab} {\tar} {\tat},clip]{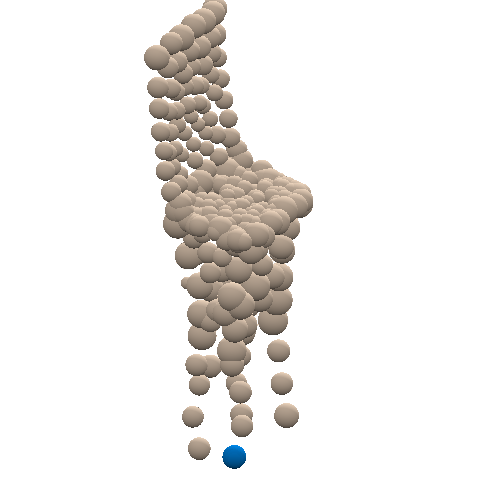}& 
        \includegraphics[width=\sizea, trim={\tal} {\tab} {\tar} {\tat},clip]{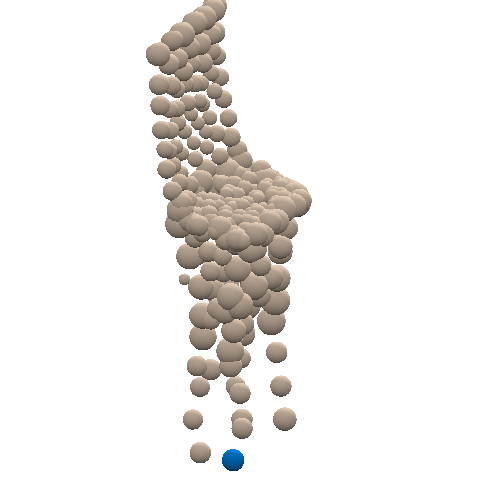}& 
        \includegraphics[width=\sizea, trim={\tal} {\tab} {\tar} {\tat},clip]{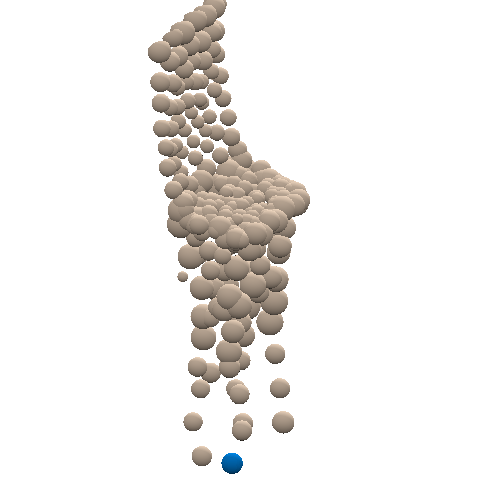}& 
        \includegraphics[width=\sizea, trim={\tal} {\tab} {\tar} {\tat},clip]{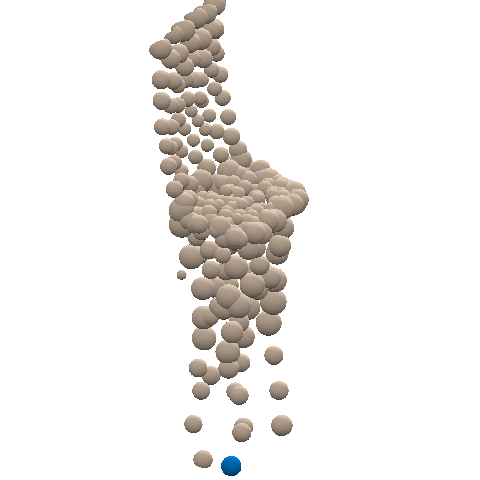}
    \end{tabular}
    \end{center}
    \caption{The effect of the latent code regularization term in the shape manipulation objective on the quality of the resulting shape. Here we move the blue sphere downwards in an attempt to make the legs of the chair longer. From left to right we show the original shape (marked in red boxes) and the intermediate shapes during the manipulation process.}
    \label{fig:latent_reg}
\end{figure*}

\section{Effect of Latent Code Regularization on Shape Manipulation}
Figure~\ref{fig:latent_reg} shows the effect of latent code regularization term $L_{\mathtt{REG}}$ on the shape manipulation process. Empirically, a latent code with high likelihood under the prior $p(\mbf{z})$ usually decodes to more plausible shapes. $L_{\mathtt{REG}}$ keeps the latent code from deviating too far from the prior during the manipulation process, improving the quality of the result shape. From a user's perspective, the regularization term guide the user input towards more semantically meaningful shapes by moving the unconstrained primitives to the correct places and guarding the user input against unreasonable configurations.

\begin{figure*}[t]
    \begin{center}
    \newcommand{\sizea}{0.13\linewidth}
    \newcommand{\tal}{1cm}
    \newcommand{\tab}{0cm}
    \newcommand{\tar}{1cm}
    \newcommand{\tat}{0cm}
    \setlength{\tabcolsep}{0pt}
    \begin{tabular}{lccccccc}
        \textbf{Output} &
        \includegraphics[width=\sizea, trim={\tal} {\tab} {\tar} {\tat},clip]{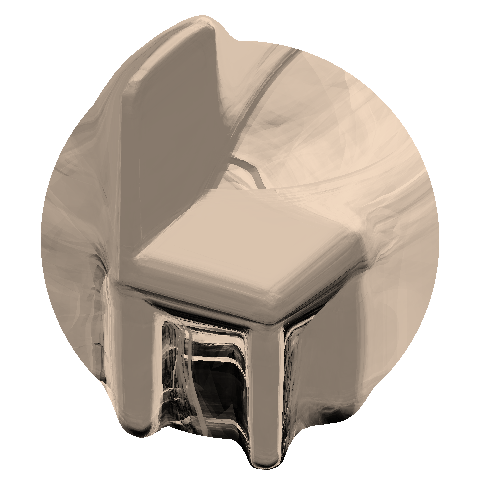}& 
        \includegraphics[width=\sizea, trim={\tal} {\tab} {\tar} {\tat},clip]{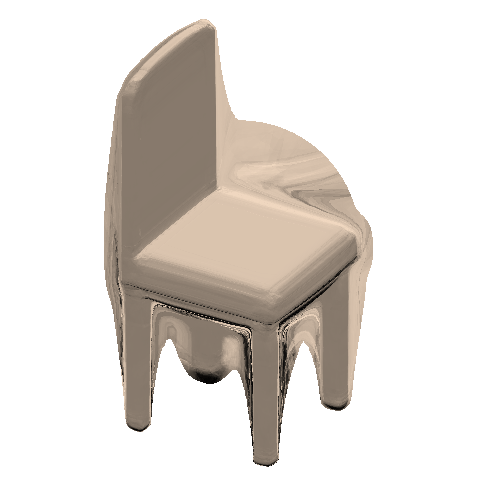}& 
        \includegraphics[width=\sizea, trim={\tal} {\tab} {\tar} {\tat},clip]{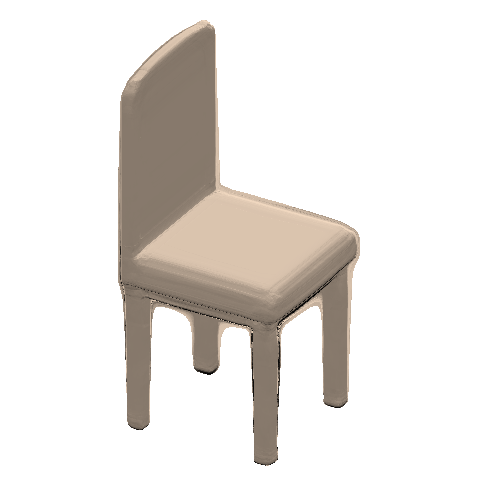}& 
        \includegraphics[width=\sizea, trim={\tal} {\tab} {\tar} {\tat},clip]{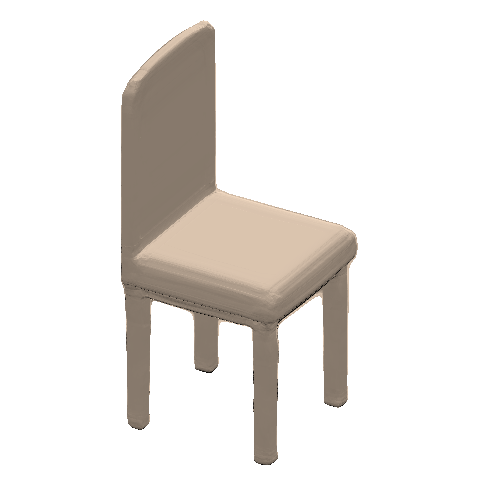}& 
        \includegraphics[width=\sizea, trim={\tal} {\tab} {\tar} {\tat},clip]{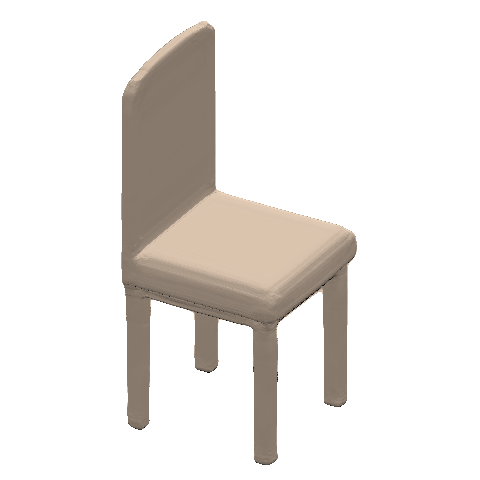}& 
        \includegraphics[width=\sizea, trim={\tal} {\tab} {\tar} {\tat},clip]{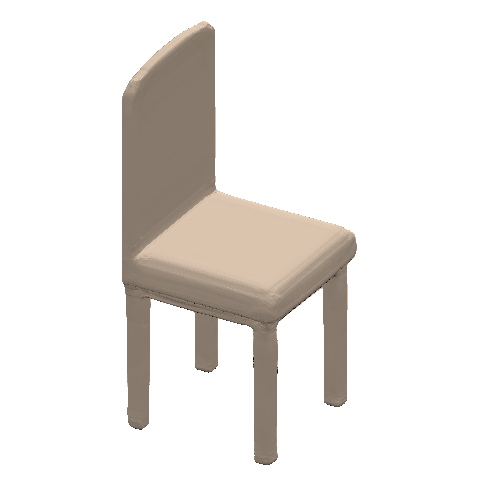}&
        \includegraphics[width=\sizea, trim={\tal} {\tab} {\tar} {\tat},clip]{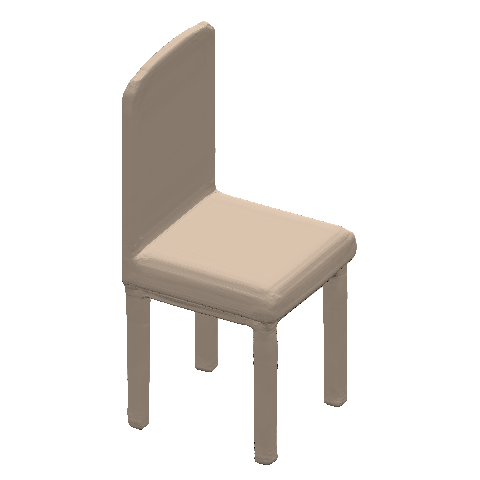}\\
        \textbf{\#Steps} & 16 & 24 & 32 & 40 & 48 & 56 & 64 \\
        \textbf{Resolution} & $480 \times 480$ & $480 \times 480$ & $480 \times 480$ & $480 \times 480$ & $480 \times 480$ & $480 \times 480$ & $480 \times 480$ \\
        \textbf{Time (s)} & 2.35 & 3.36 & 4.36 & 5.37 & 6.39 & 7.40 & 8.41 \\
        \hdashline
        \textbf{Output} &
        \includegraphics[width=\sizea, trim=0.167cm 0 0.167cm 0,clip]{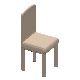}& 
        \includegraphics[width=\sizea, trim=0.333cm 0 0.333cm 0,clip]{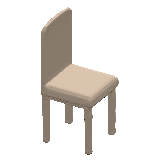}&
        \includegraphics[width=\sizea, trim=0.5cm 0 0.5cm 0,clip]{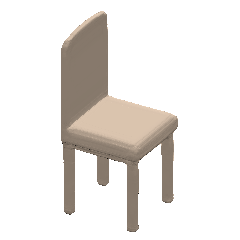}&
        \includegraphics[width=\sizea, trim=0.667cm 0 0.667cm 0,clip]{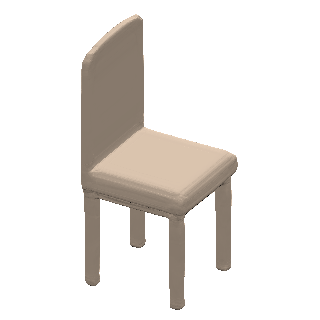}&
        \includegraphics[width=\sizea, trim=0.8cm 0 0.8cm 0,clip]{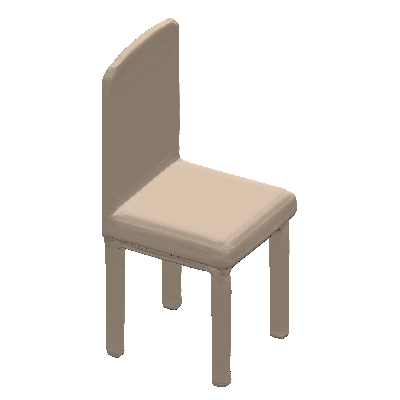}&
        \includegraphics[width=\sizea, trim=1cm 0 1cm 0, clip]{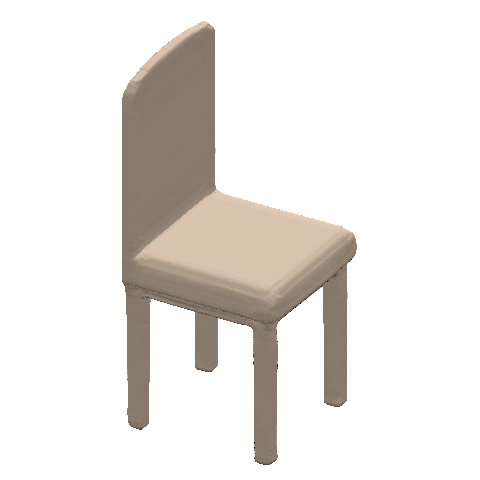}&
        \includegraphics[width=\sizea, trim=1.167cm 0 1.167cm 0,clip]{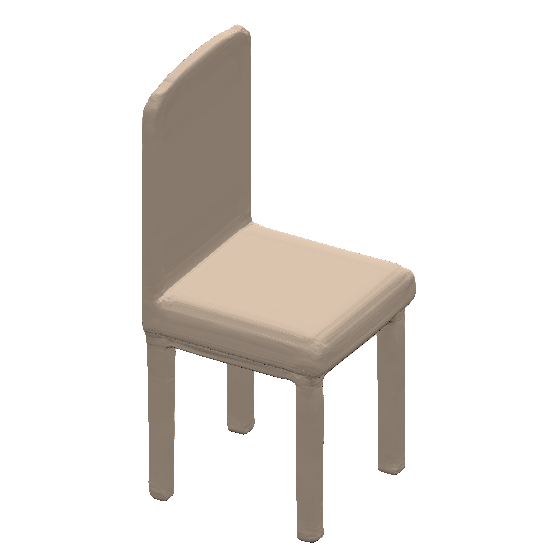}\\
        \textbf{\#Steps} & 64 & 64 & 64 & 64 & 64 & 64 & 64 \\
        \textbf{Resolution} & $80 \times 80$ & $160 \times 160$ & $240 \times 240$ & $320 \times 320$ & $400 \times 400$ & $480 \times 480$ & $560 \times 560$ \\
        \textbf{Time (s)} & 0.26 & 0.95 & 2.11 & 3.73 & 5.85 & 8.44 & 11.49 \\
    \end{tabular}
    \end{center}
    \caption{Trade-off between rendering quality and speed on the fine-scale representation. Higher resolution and larger number of ray-marching iterations lead to better image quality, at the cost of longer running time. Our model is capable of adjusting the trade-off on the fly to adjust to different scenarios. For example, we can render shapes in lower quality during interactive manipulation, and render a high quality result once the manipulation is done.}
    \label{fig:speed}
\end{figure*}

\section{Analysis of Running Time}
As mentioned in the main paper, our shape manipulation framework is able to run at real time. Here we provide a more comprehensive analysis on the running time of our model to further back up our claim. All of the benchmarks are implemented with PyTorch and run on a single GTX 1080ti GPU. We assume a single-user scenario, where the batch size is only one.

For the primitive-based representation section, it takes an average of 0.66ms to obtain the primitive attributes from latent code, and it takes an average of 2.74ms to perform one gradient descent step (including forward and backward) to update the latent code and to obtain the attributes of the updated shape, after receiving the user input. Even with our less-that-optimal implementation, this is already fast enough to provide the user with real-time feedback. Once the primitive attributes are obtained, the shape can then be rendered conveniently and rapidly with real time rendering engines and hardware acceleration.

For the high-resolution representation, assuming we are rendering the SDF with ray-marching method, there are two parameters that determine the render quality and speed: image resolution and number of ray-marching steps. The effects of the two parameters on image quality and rendering time are presented in Figure \ref{fig:speed}. All the renderings shown in the main paper are rendered with 64 iterations at a resolution of $480 \times 480$. Although at full resolution and highest quality, the rendering of high-resolution representation is not as fast as the primitive-based representation, a reasonable trade-off between time and quality can be obtained. During interactive manipulation, we can present the user with reduced resolution rendering at a reduced rate, in addition to the real-time rendering of the primitive-based representation, and render the full resolution result only when needed.

\section{Detailed Experimental Setting}
We use a 128-dimensional latent code $\mathbf{z}_j$ throughout the experiments.
For the high-resolution SDF representation, we use a 8-layer MLP with one cross-connection. The 131-dim input is the concatenation of the 128-dim latent code $\mathbf{z}_j$ and the 3D coordinate $\mathbf{p}$. The output is the predicted SDF value at this 3D coordinate. For the primitive-based representation, we use the same network architecture with $\mathbf{z}_j$ as the only input, and the attributes (center coordinates and log radii, denoted by $\pmb{\alpha}_j$ in the main paper) of 256 spheres as output. The networks are shown in Figure \ref{fig:sdfnet}. Weight normalization is used on all the fully connected layers. We use ReLU activation on all but the last fully connected layers. We use dropout with a probability of 0.2 on the output of all but the last fully connected layers, only in the high-resolution network $g_\theta$.

For all the experiments, we use Adam optimizer with $\beta_1 = 0.9$ and $\beta_2 = 0.999$. The learning rates are 5e-9
for $\theta, \phi$ and 1e-8 for $\pmb{\mu}_j, \pmb{\sigma}_j$. Each batch consists of 64 shapes; for each shape we sample 2048 SDF values for the high-res representation and 1024 for the primitive-based representation. We train the model for 2800 epochs and drop the learning rate by 50\% after every 700 epochs. We empirically set $\lambda_1 = \lambda_2 = 1\mathrm{e}5$. Their values affect the trade-off between latent space compactness and reconstruction quality due to $D_\textit{KL}$.

\section{SDF Losses}
We use truncated SDF loss on both coarse and fine shape representations during training and shape encoding. This has been shown beneficial in DeepSDF. For high resolution representation, we truncate the SDF on both inside and outside:
\begin{equation}
    L_\mathtt{SDF}^\mathrm{fine}(d, s) = 
    \begin{cases}
        \max(d, -\delta) + \delta & \quad s < -\delta, \\
        |d-s| & \quad -\delta \leq s \leq \delta, \\
        \delta - \min(d, \delta) & \quad s > \delta.
    \end{cases}
\end{equation}
For primitive-based representation, we only truncate the SDF inside the shape to zero, as the SDF outside the shape is guaranteed to be valid (metric):
\begin{equation}
    L_\mathtt{SDF}^\mathrm{coarse}(d, s) = 
    \begin{cases}
        \max(d, 0) & \quad s < 0, \\
        |d-s| & \quad s \geq 0.
    \end{cases}
\end{equation}

\begin{figure*}
\vspace{-4pt}
  \centering
  \setlength{\fboxrule}{2.0pt}
\includegraphics[width=0.07\textwidth]{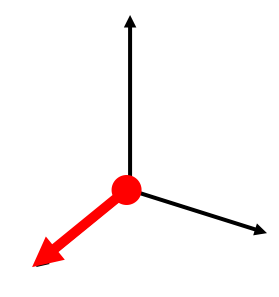}
\fbox{{\includegraphics[trim={2.9cm 0.6cm 3.3cm 0.5cm},clip,width=0.125\textwidth]{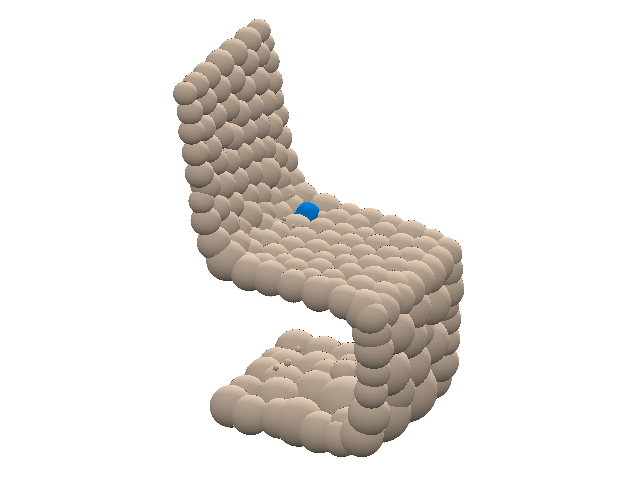}}}
\includegraphics[trim={2.9cm 0.6cm 3.3cm 0.5cm},clip,width=0.125\textwidth]{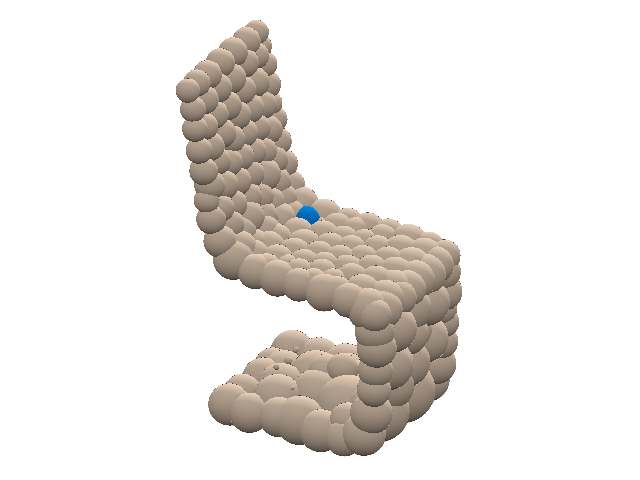}
\includegraphics[trim={2.9cm 0.6cm 3.3cm 0.5cm},clip,width=0.125\textwidth]{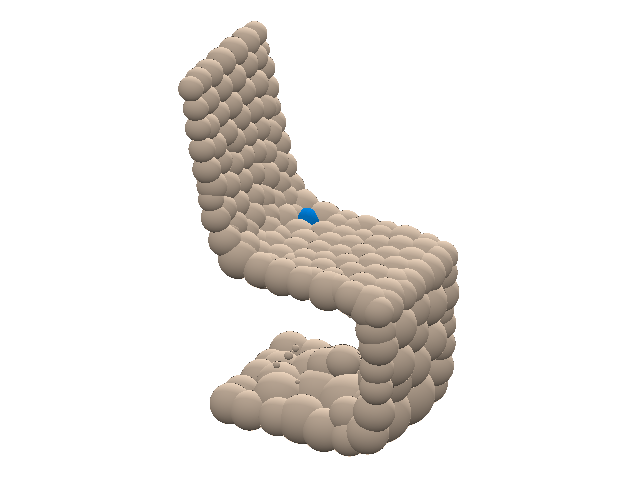}
\includegraphics[trim={2.9cm 0.6cm 3.3cm 0.5cm},clip,width=0.125\textwidth]{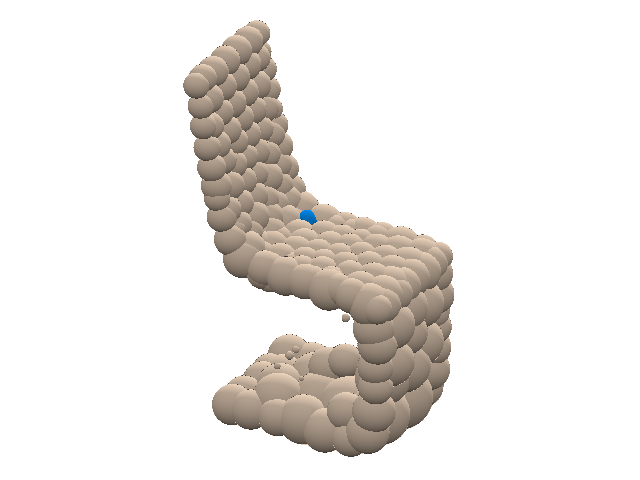} 
\includegraphics[trim={2.9cm 0.6cm 3.3cm 0.5cm},clip,width=0.125\textwidth]{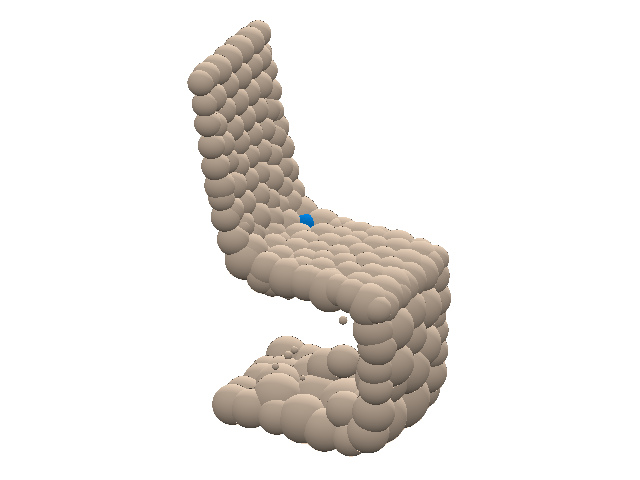}
\includegraphics[trim={2.9cm 0.6cm 3.3cm 0.5cm},clip,width=0.125\textwidth]{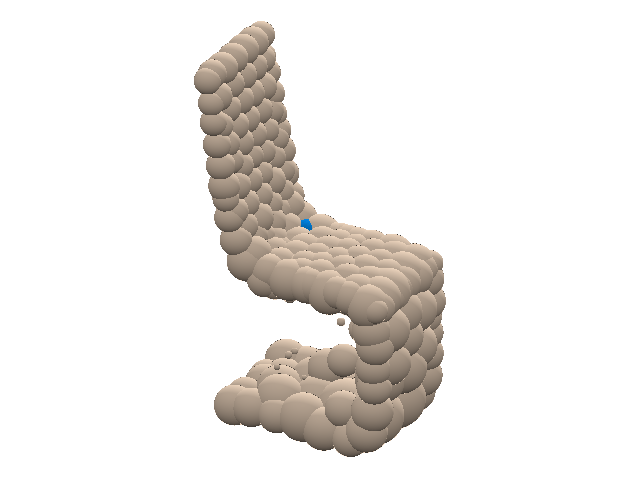}
\includegraphics[trim={2.9cm 0.6cm 3.3cm 0.5cm},clip,width=0.125\textwidth]{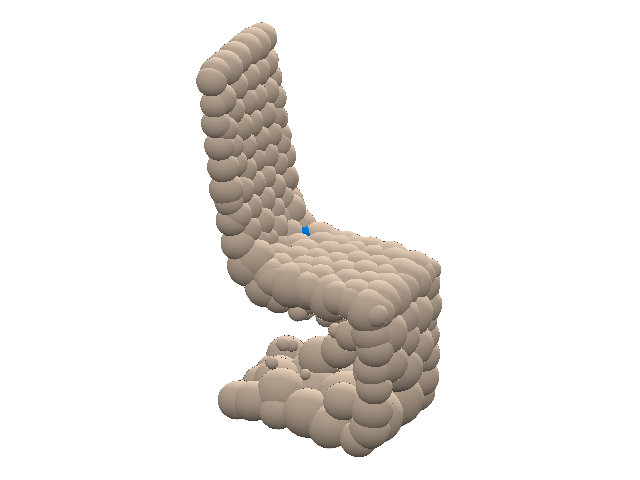} \\
\includegraphics[width=0.07\textwidth]{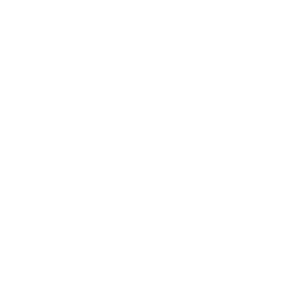}
\fbox{{\includegraphics[trim={2.9cm 0.6cm 3.3cm 0.5cm},clip,width=0.125\textwidth]{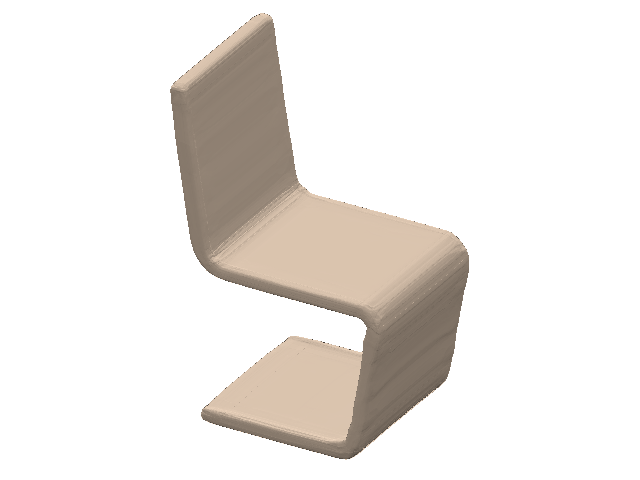}}}
\includegraphics[trim={2.9cm 0.6cm 3.3cm 0.5cm},clip,width=0.125\textwidth]{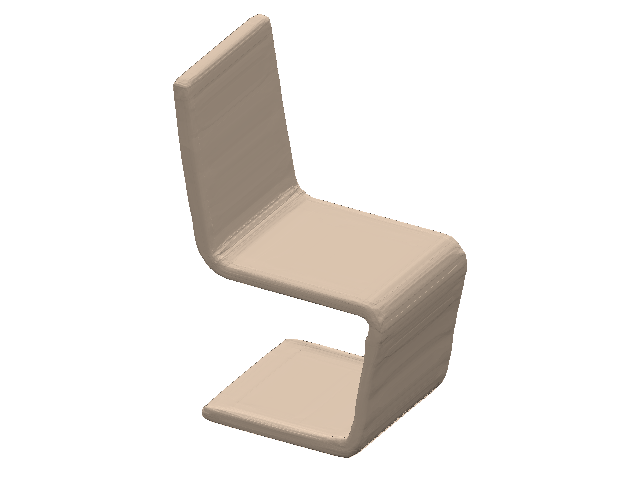}
\includegraphics[trim={2.9cm 0.6cm 3.3cm 0.5cm},clip,width=0.125\textwidth]{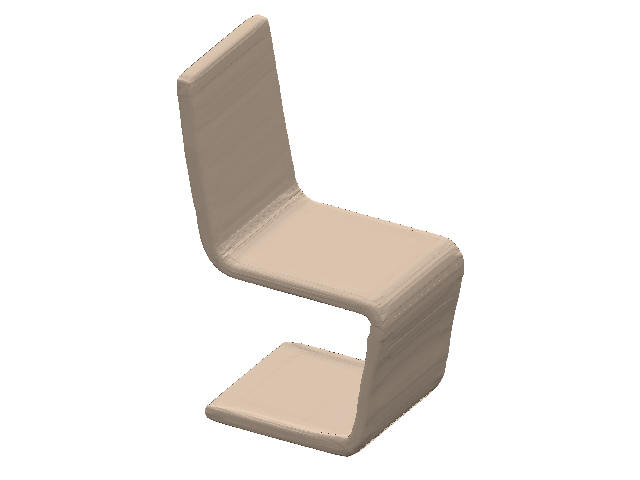}
\includegraphics[trim={2.9cm 0.6cm 3.3cm 0.5cm},clip,width=0.125\textwidth]{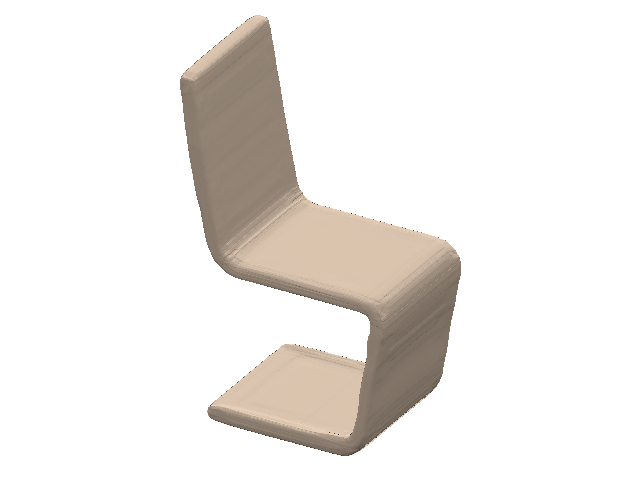} 
\includegraphics[trim={2.9cm 0.6cm 3.3cm 0.5cm},clip,width=0.125\textwidth]{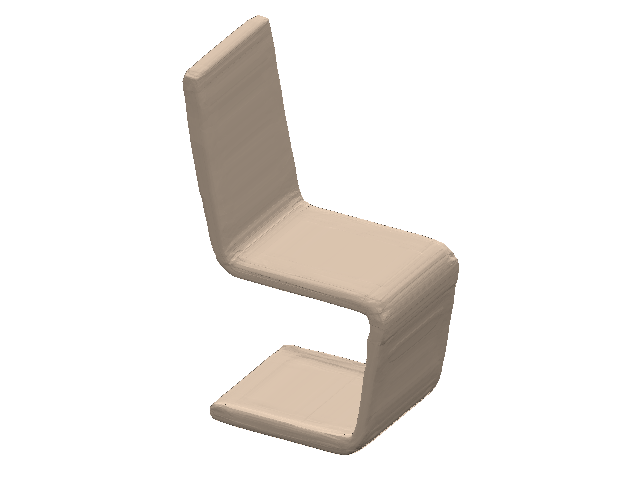}
\includegraphics[trim={2.9cm 0.6cm 3.3cm 0.5cm},clip,width=0.125\textwidth]{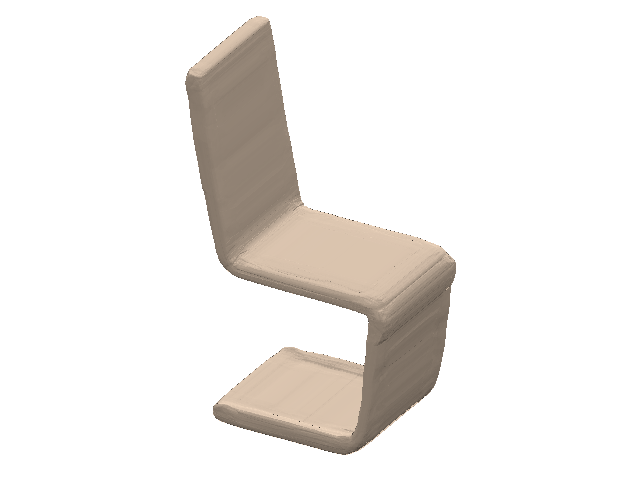}
\includegraphics[trim={2.9cm 0.6cm 3.3cm 0.5cm},clip,width=0.125\textwidth]{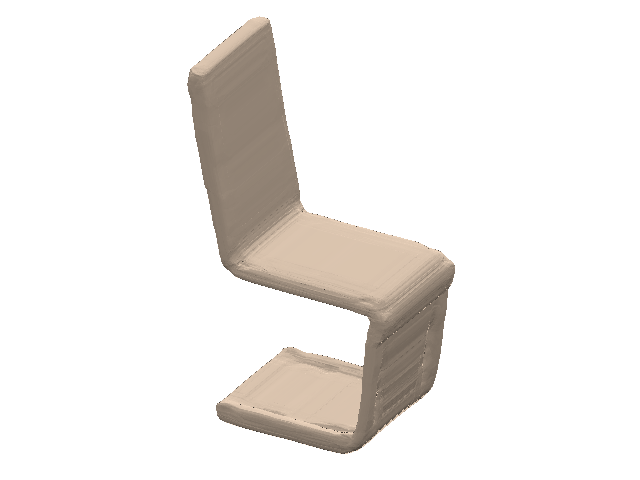}\\
\includegraphics[width=0.07\textwidth]{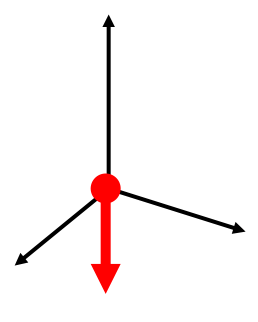}
\fbox{{\includegraphics[trim={2.9cm 0.6cm 3.3cm 0.5cm},clip,width=0.125\textwidth]{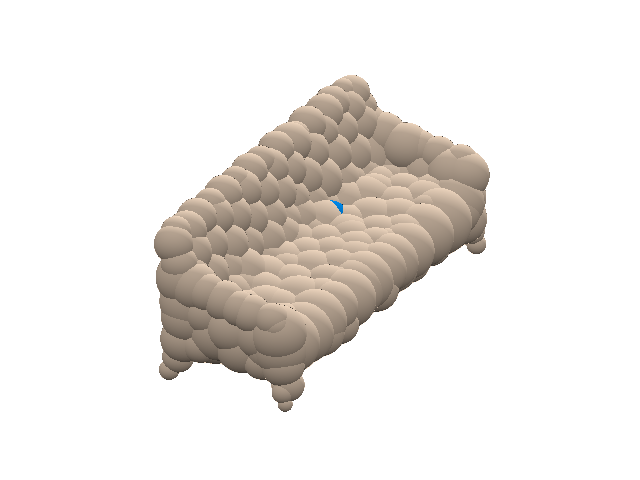}}}
\includegraphics[trim={2.9cm 0.6cm 3.3cm 0.5cm},clip,width=0.125\textwidth]{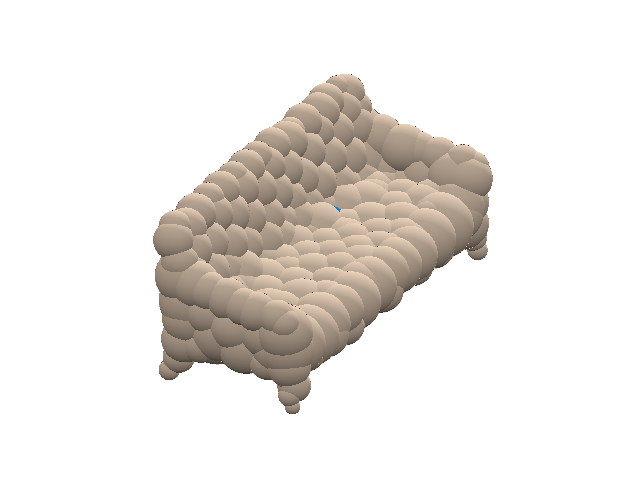}
\includegraphics[trim={2.9cm 0.6cm 3.3cm 0.5cm},clip,width=0.125\textwidth]{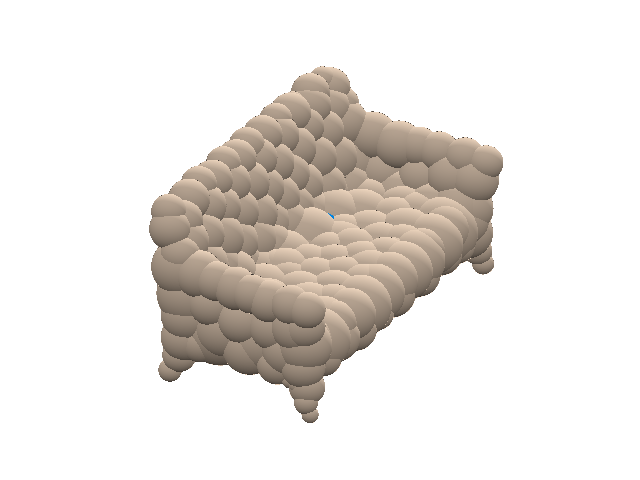}
\includegraphics[trim={2.9cm 0.6cm 3.3cm 0.5cm},clip,width=0.125\textwidth]{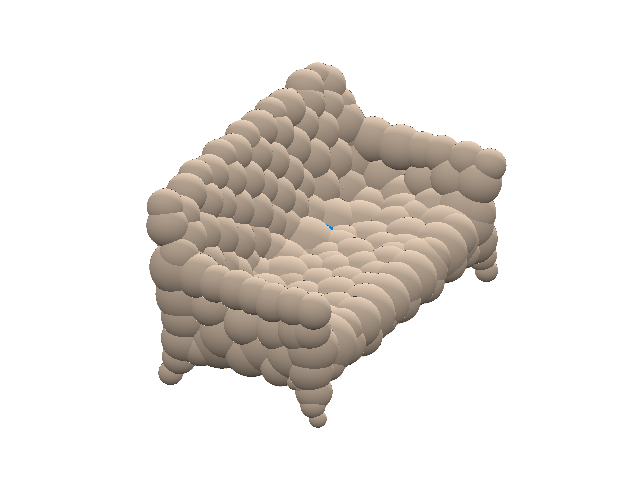} 
\includegraphics[trim={2.9cm 0.6cm 3.3cm 0.5cm},clip,width=0.125\textwidth]{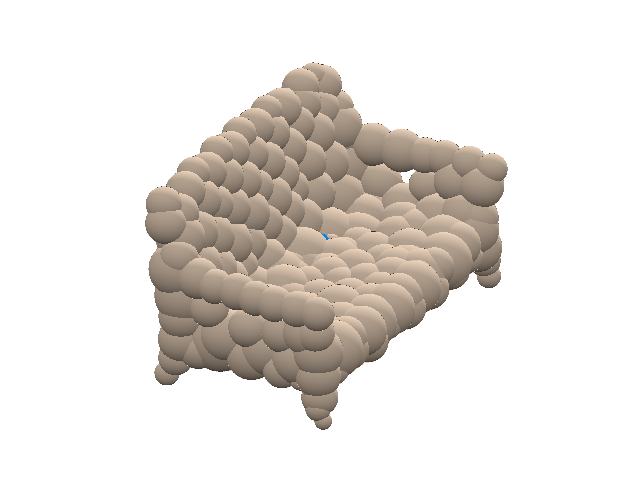}
\includegraphics[trim={2.9cm 0.6cm 3.3cm 0.5cm},clip,width=0.125\textwidth]{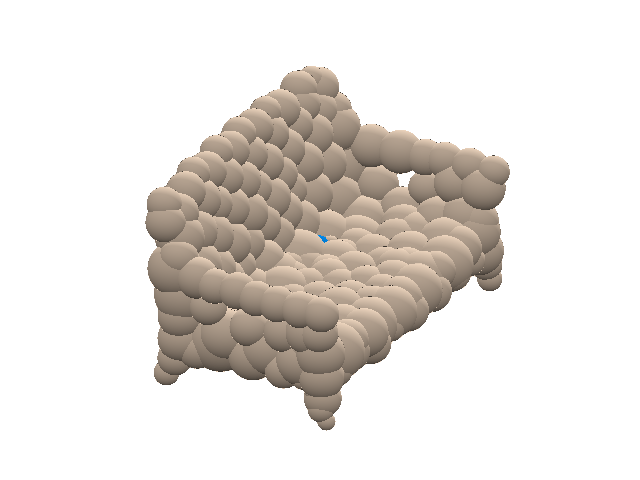}
\includegraphics[trim={2.9cm 0.6cm 3.3cm 0.5cm},clip,width=0.125\textwidth]{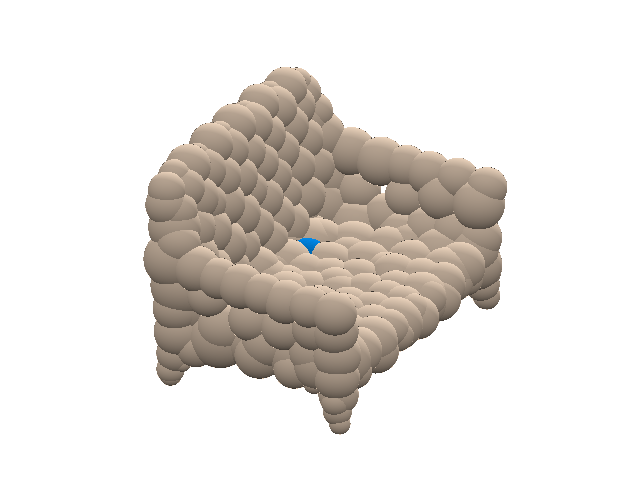} \\
\includegraphics[width=0.07\textwidth]{supp/figures/manip1/white.png}
\fbox{{\includegraphics[trim={2.9cm 0.6cm 3.3cm 0.5cm},clip,width=0.125\textwidth]{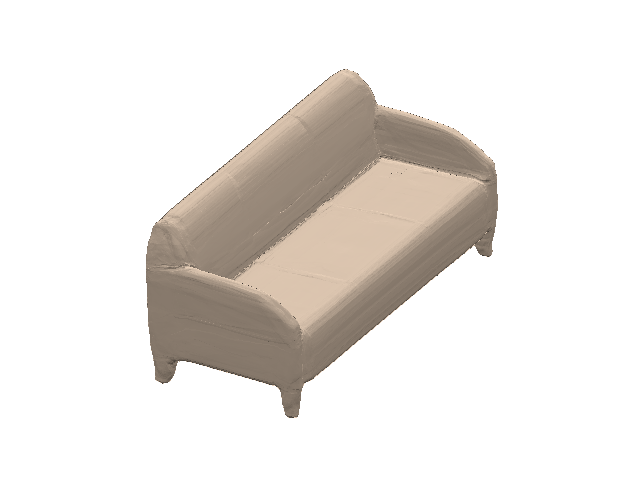}}}
\includegraphics[trim={2.9cm 0.6cm 3.3cm 0.5cm},clip,width=0.125\textwidth]{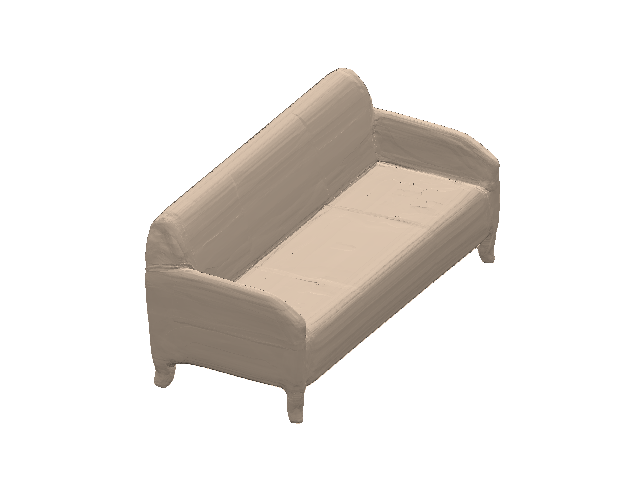}
\includegraphics[trim={2.9cm 0.6cm 3.3cm 0.5cm},clip,width=0.125\textwidth]{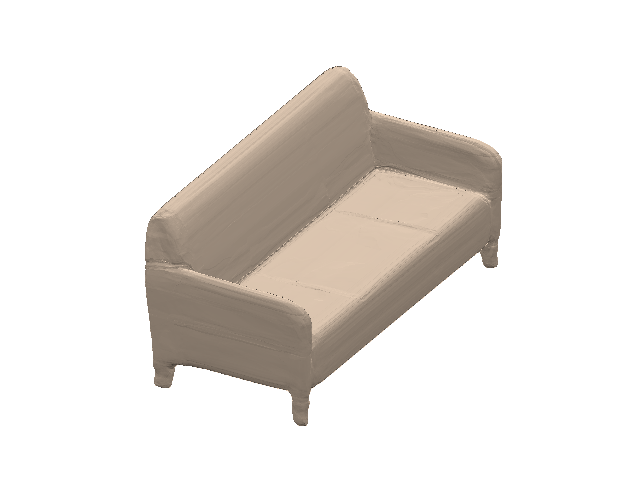}
\includegraphics[trim={2.9cm 0.6cm 3.3cm 0.5cm},clip,width=0.125\textwidth]{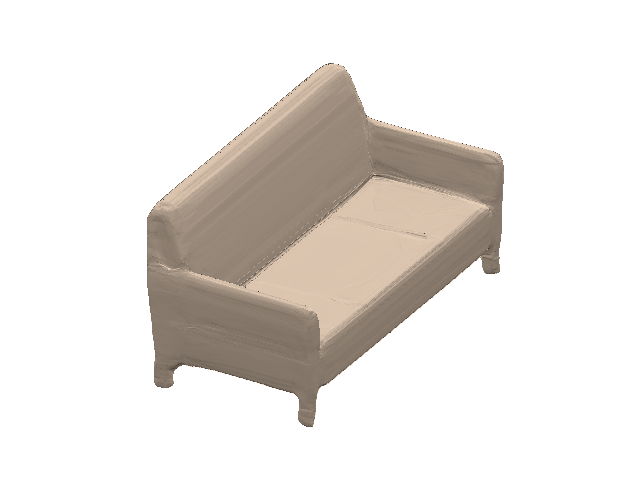} 
\includegraphics[trim={2.9cm 0.6cm 3.3cm 0.5cm},clip,width=0.125\textwidth]{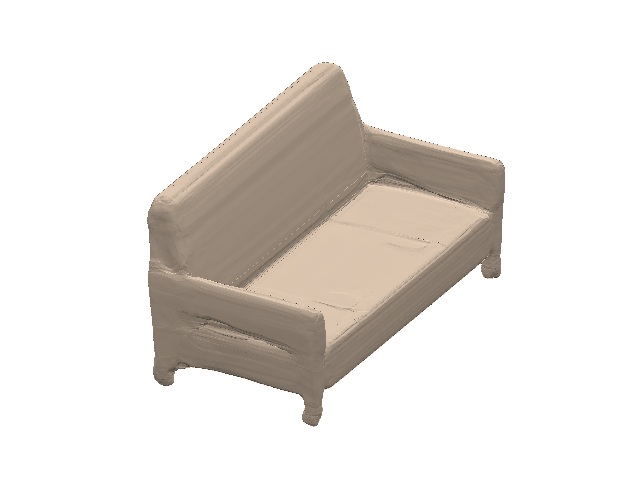}
\includegraphics[trim={2.9cm 0.6cm 3.3cm 0.5cm},clip,width=0.125\textwidth]{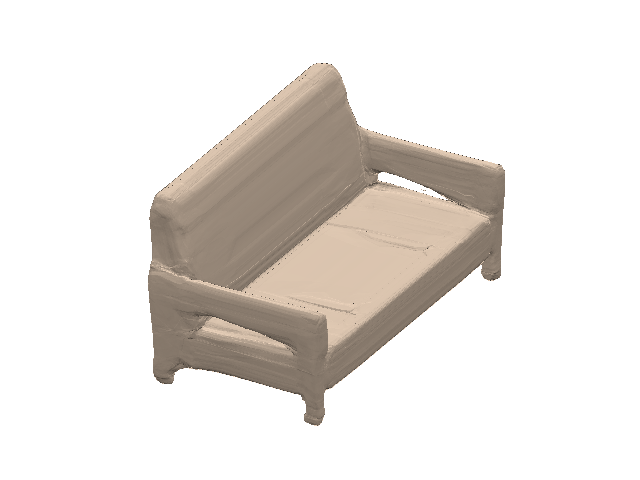}
\includegraphics[trim={2.9cm 0.6cm 3.3cm 0.5cm},clip,width=0.125\textwidth]{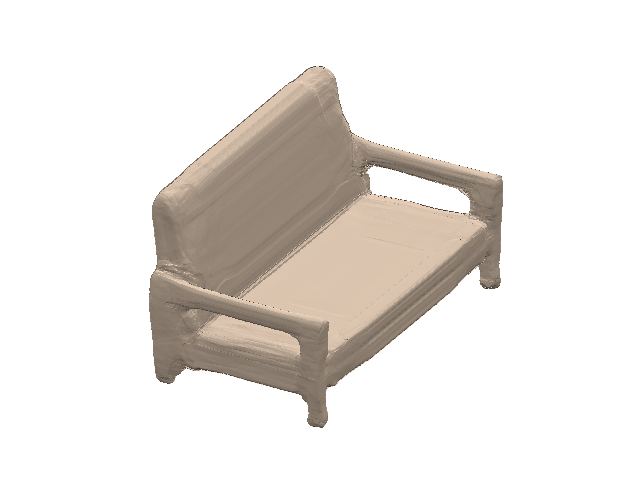}\\
\includegraphics[width=0.07\textwidth]{supp/figures/manip1/left.png}
\fbox{{\includegraphics[trim={2.9cm 0.6cm 3.3cm 0.5cm},clip,width=0.125\textwidth]{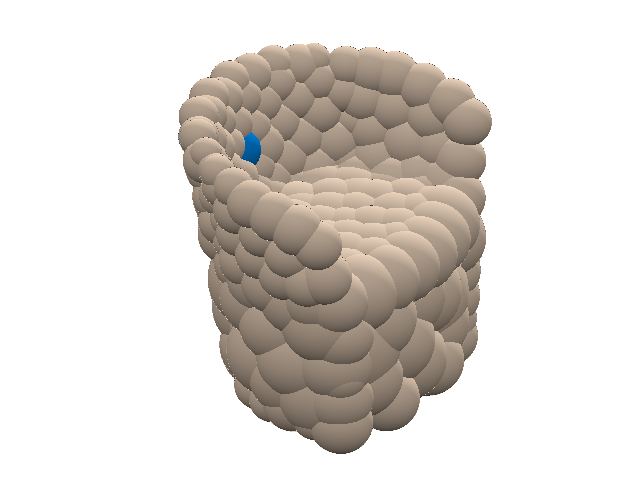}}}
\includegraphics[trim={2.9cm 0.6cm 3.3cm 0.5cm},clip,width=0.125\textwidth]{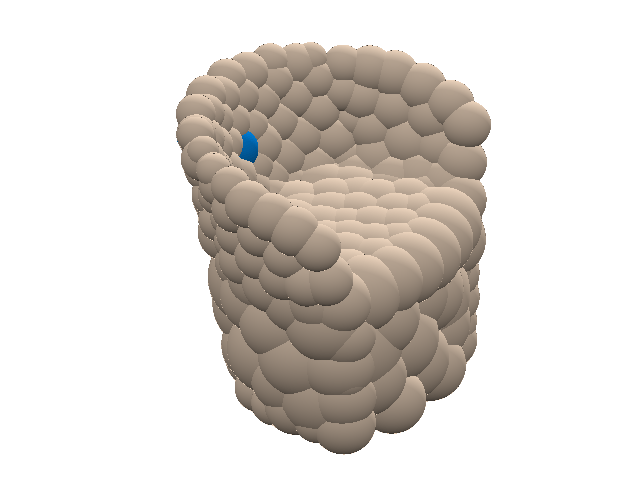}
\includegraphics[trim={2.9cm 0.6cm 3.3cm 0.5cm},clip,width=0.125\textwidth]{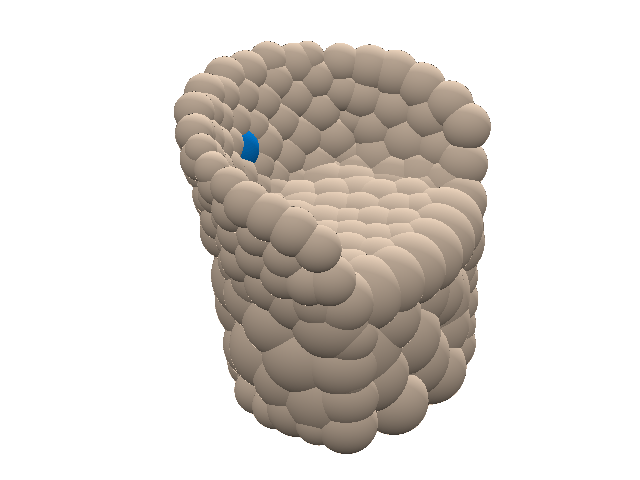}
\includegraphics[trim={2.9cm 0.6cm 3.3cm 0.5cm},clip,width=0.125\textwidth]{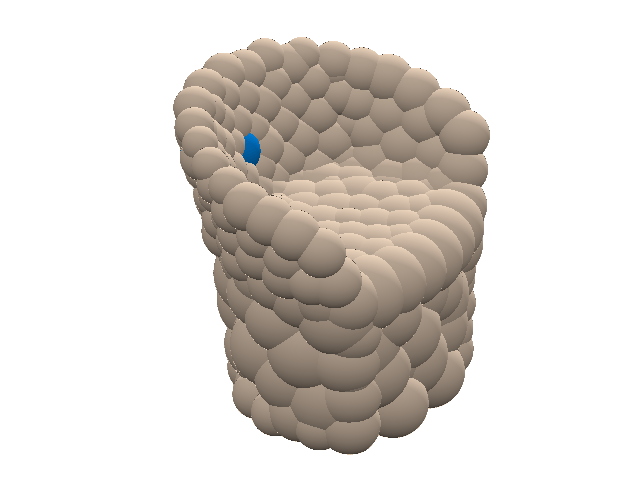} 
\includegraphics[trim={2.9cm 0.6cm 3.3cm 0.5cm},clip,width=0.125\textwidth]{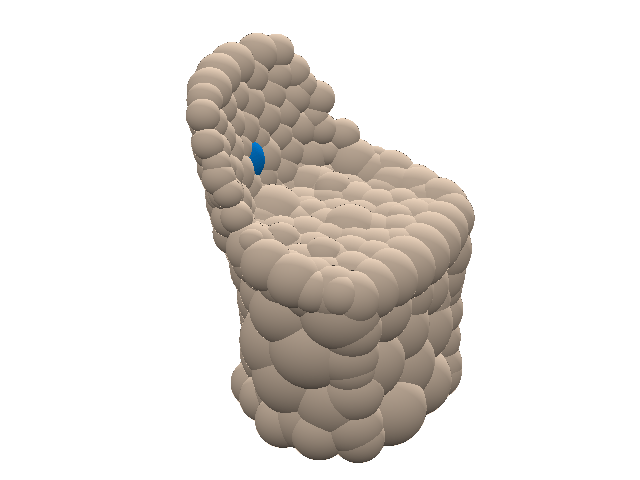}
\includegraphics[trim={2.9cm 0.6cm 3.3cm 0.5cm},clip,width=0.125\textwidth]{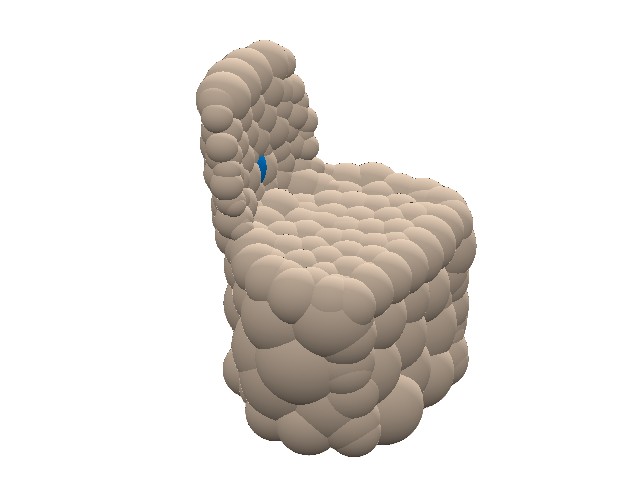}
\includegraphics[trim={2.9cm 0.6cm 3.3cm 0.5cm},clip,width=0.125\textwidth]{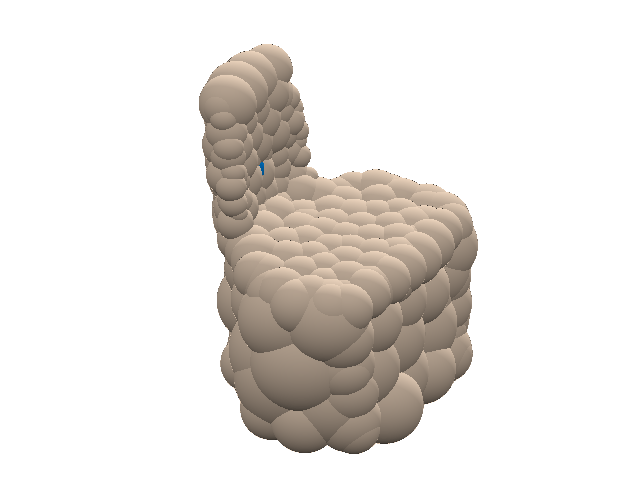} \\
\includegraphics[width=0.07\textwidth]{supp/figures/manip1/white.png}
\fbox{{\includegraphics[trim={2.9cm 0.6cm 3.3cm 0.5cm},clip,width=0.125\textwidth]{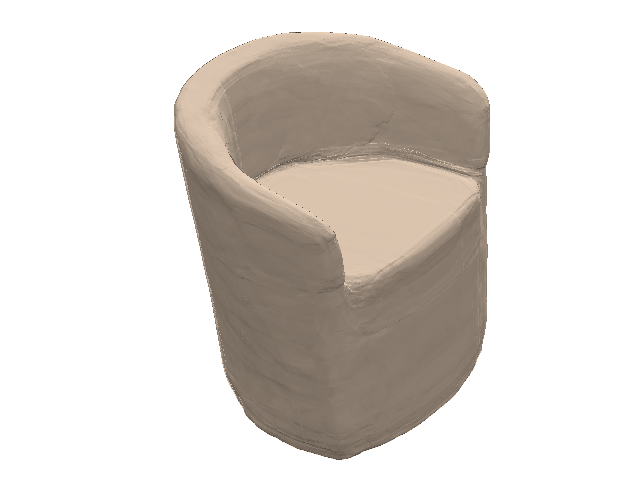}}}
\includegraphics[trim={2.9cm 0.6cm 3.3cm 0.5cm},clip,width=0.125\textwidth]{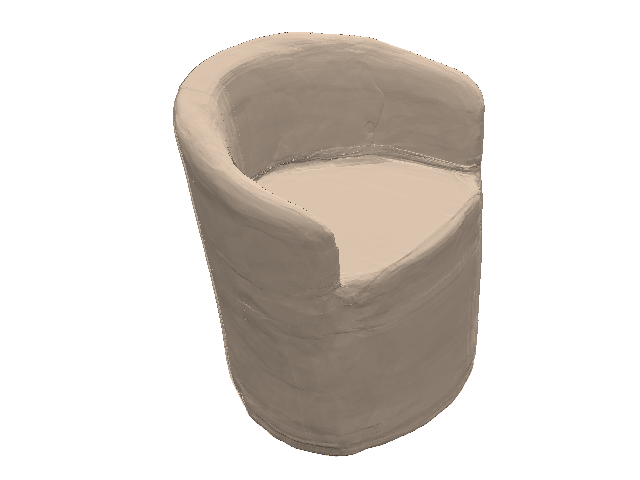}
\includegraphics[trim={2.9cm 0.6cm 3.3cm 0.5cm},clip,width=0.125\textwidth]{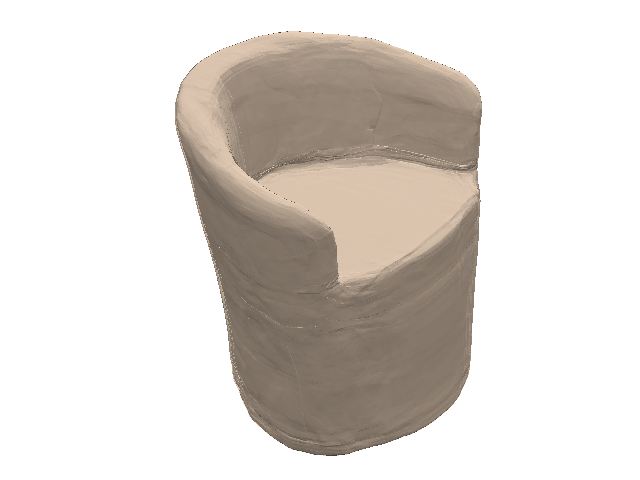}
\includegraphics[trim={2.9cm 0.6cm 3.3cm 0.5cm},clip,width=0.125\textwidth]{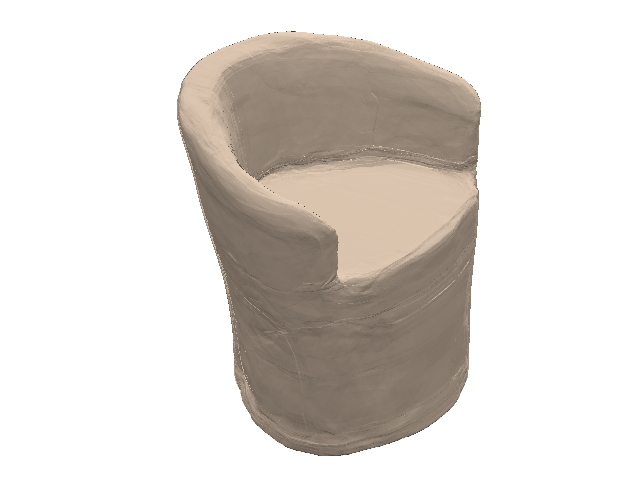} 
\includegraphics[trim={2.9cm 0.6cm 3.3cm 0.5cm},clip,width=0.125\textwidth]{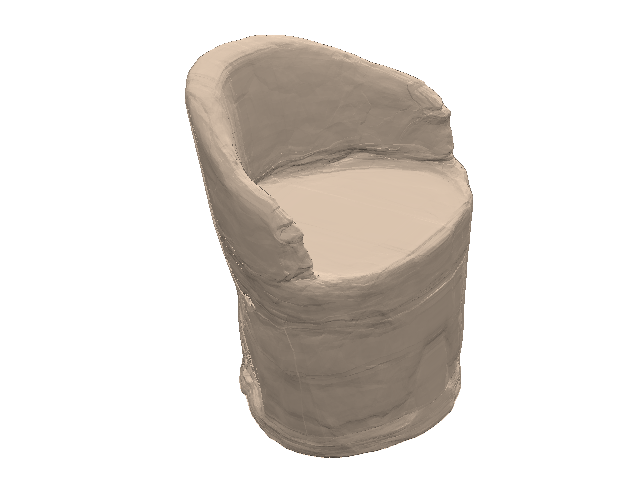}
\includegraphics[trim={2.9cm 0.6cm 3.3cm 0.5cm},clip,width=0.125\textwidth]{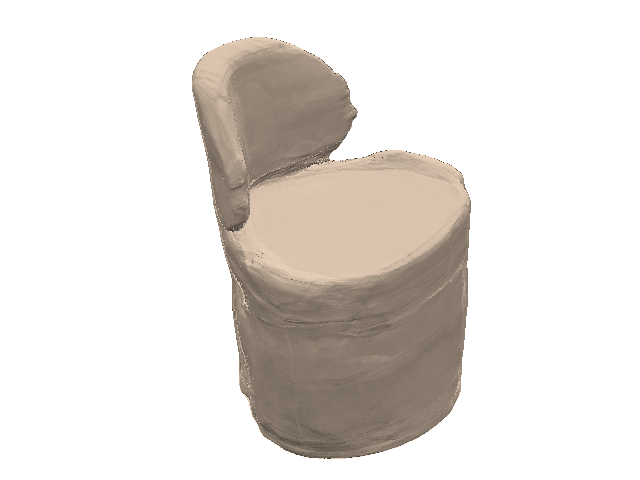}
\includegraphics[trim={2.9cm 0.6cm 3.3cm 0.5cm},clip,width=0.125\textwidth]{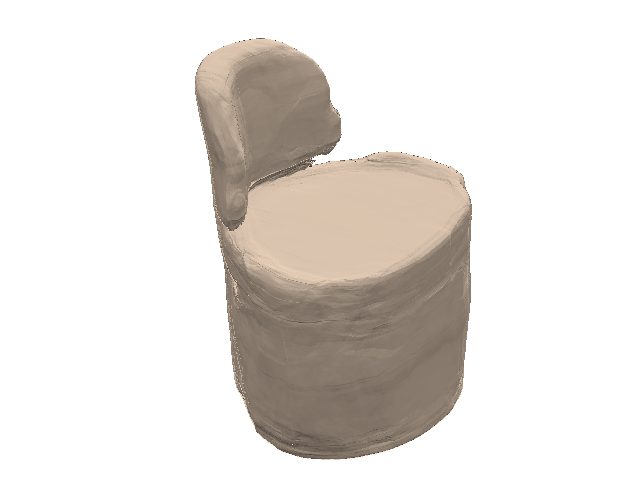}\\
\includegraphics[width=0.07\textwidth]{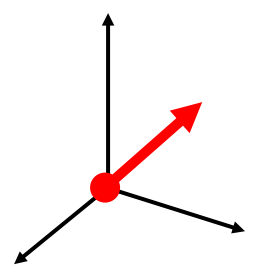}
\fbox{{\includegraphics[trim={2.9cm 3.6cm 3.3cm 3.5cm},clip,width=0.125\textwidth]{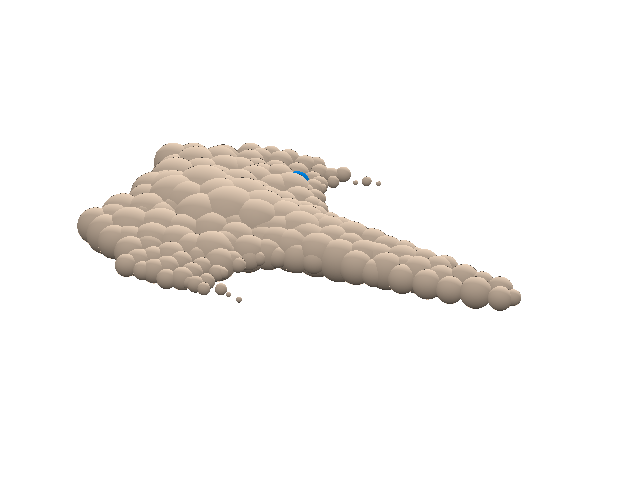}}}
\includegraphics[trim={2.9cm 3.6cm 3.3cm 3.5cm},clip,width=0.125\textwidth]{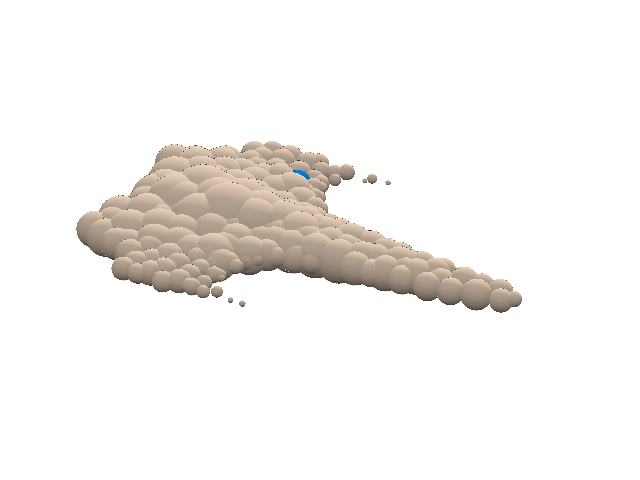}
\includegraphics[trim={2.9cm 3.6cm 3.3cm 3.5cm},clip,width=0.125\textwidth]{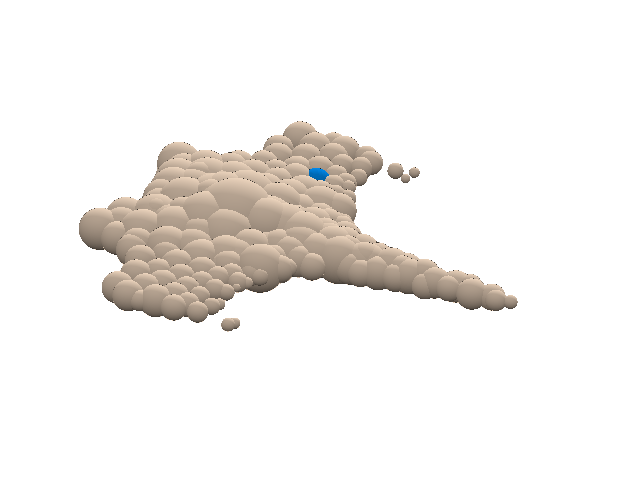}
\includegraphics[trim={2.9cm 3.6cm 3.3cm 3.5cm},clip,width=0.125\textwidth]{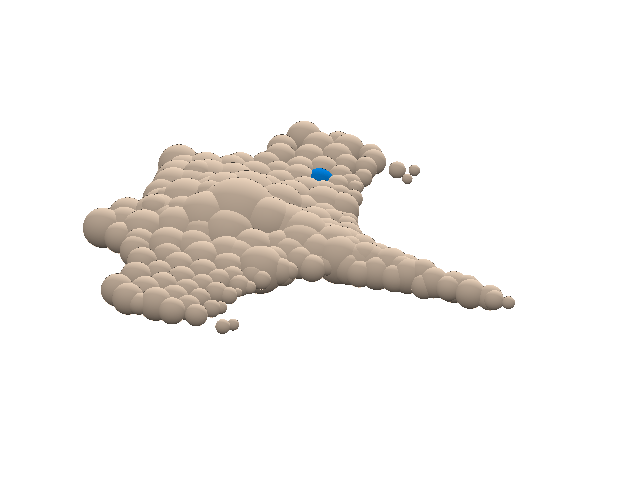} 
\includegraphics[trim={2.9cm 3.6cm 3.3cm 3.5cm},clip,width=0.125\textwidth]{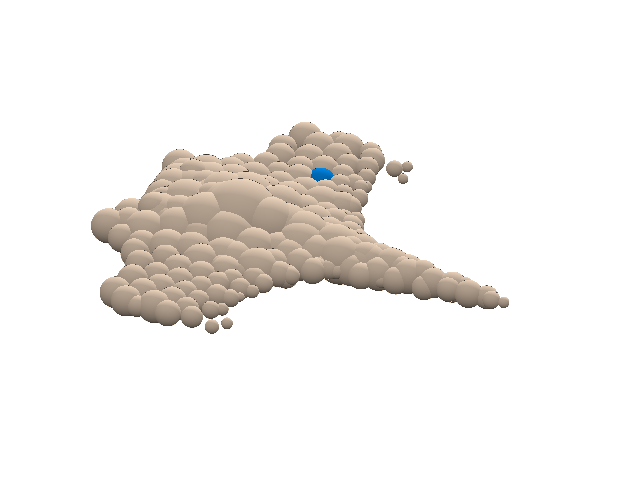}
\includegraphics[trim={2.9cm 3.6cm 3.3cm 3.5cm},clip,width=0.125\textwidth]{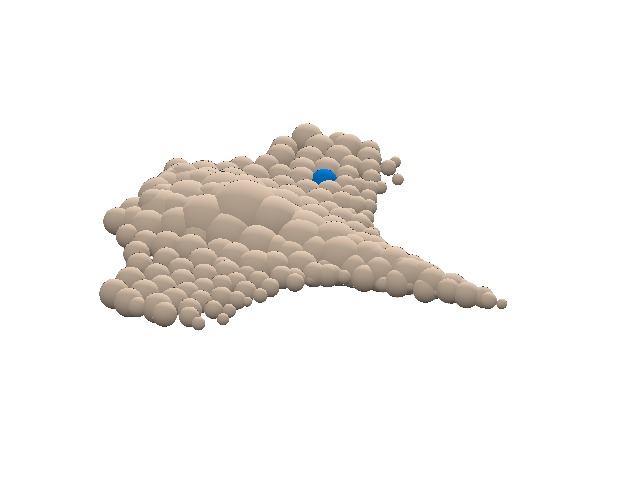}
\includegraphics[trim={2.9cm 3.6cm 3.3cm 3.5cm},clip,width=0.125\textwidth]{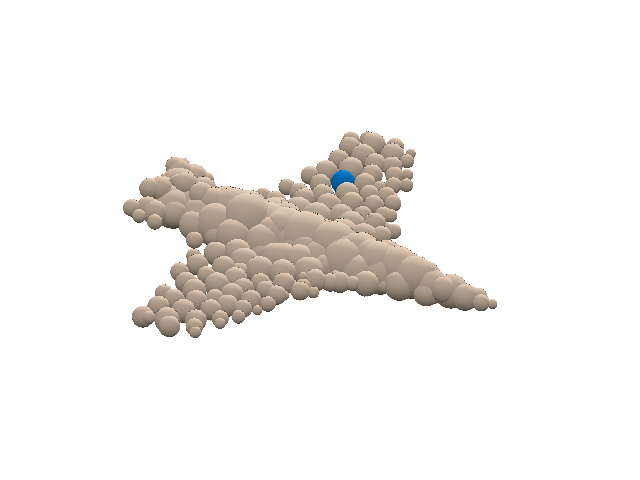}  \\
\includegraphics[width=0.07\textwidth]{supp/figures/manip1/white.png}
\fbox{{\includegraphics[trim={2.9cm 3.6cm 3.3cm 3.5cm},clip,width=0.125\textwidth]{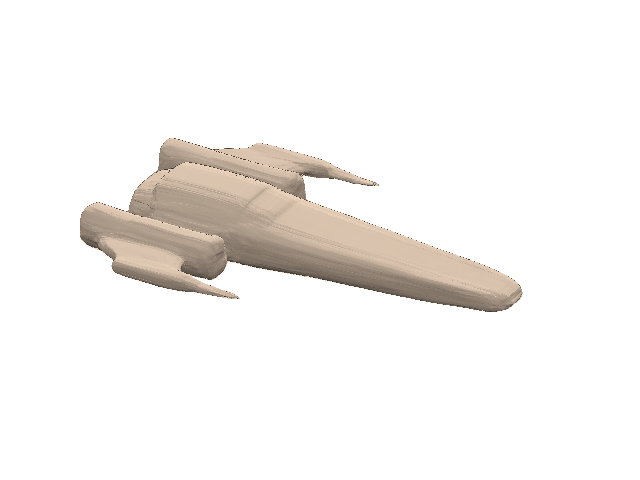}}}
\includegraphics[trim={2.9cm 3.6cm 3.3cm 3.5cm},clip,width=0.125\textwidth]{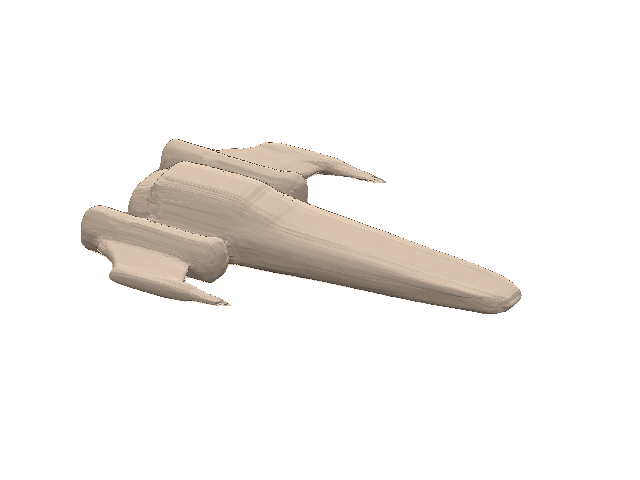}
\includegraphics[trim={2.9cm 3.6cm 3.3cm 3.5cm},clip,width=0.125\textwidth]{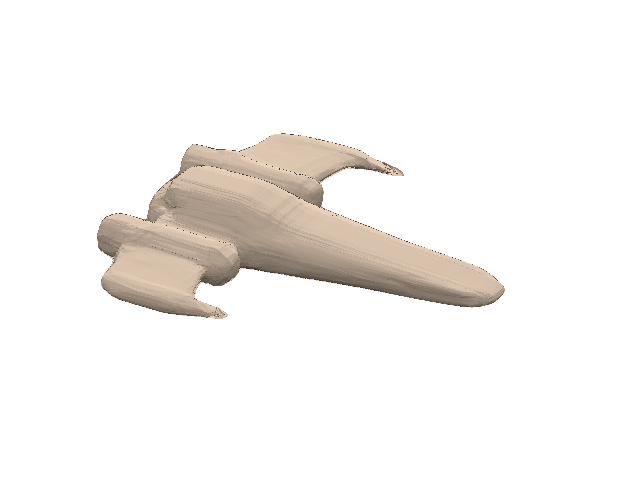}
\includegraphics[trim={2.9cm 3.6cm 3.3cm 3.5cm},clip,width=0.125\textwidth]{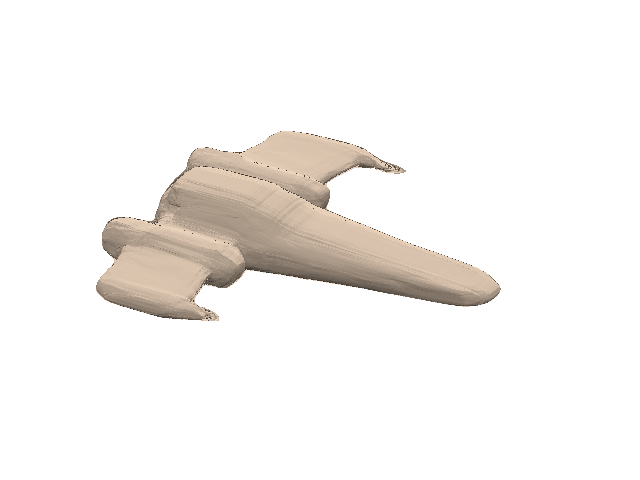} 
\includegraphics[trim={2.9cm 3.6cm 3.3cm 3.5cm},clip,width=0.125\textwidth]{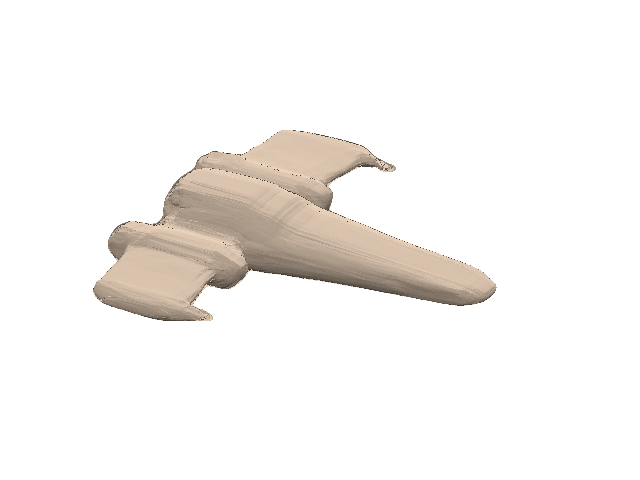}
\includegraphics[trim={2.9cm 3.6cm 3.3cm 3.5cm},clip,width=0.125\textwidth]{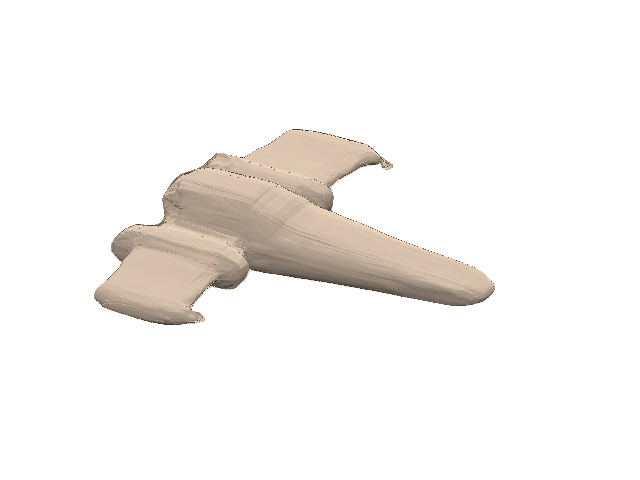}
\includegraphics[trim={2.9cm 3.6cm 3.3cm 3.5cm},clip,width=0.125\textwidth]{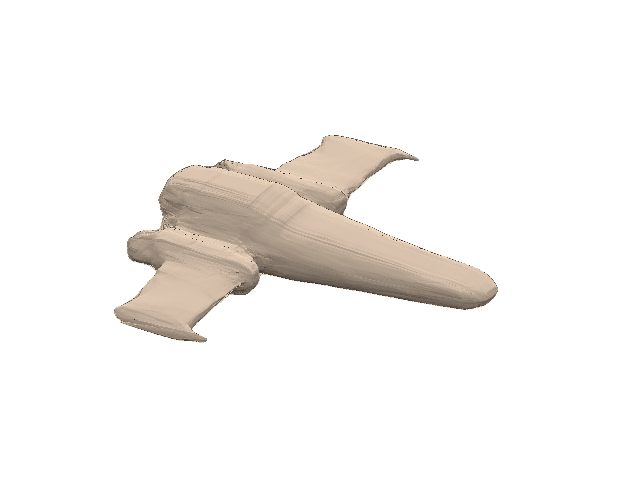}\\
\includegraphics[width=0.07\textwidth]{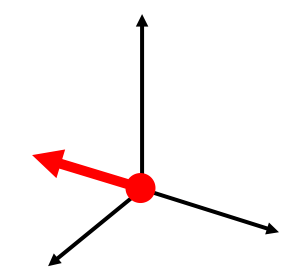}
\fbox{{\includegraphics[trim={2.9cm 3.6cm 3.3cm 3.5cm},clip,width=0.125\textwidth]{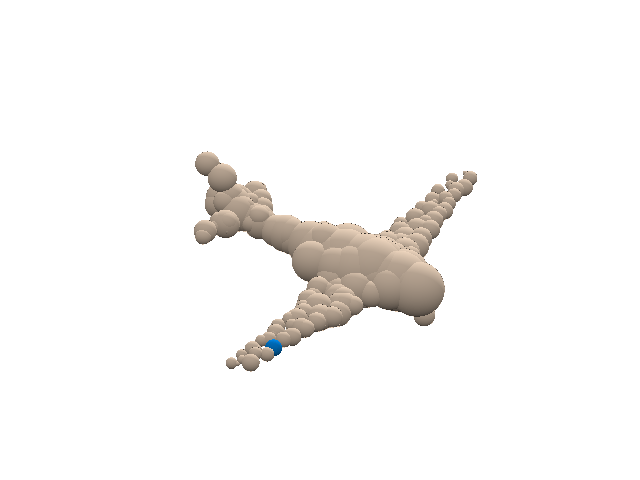}}}
\includegraphics[trim={2.9cm 3.6cm 3.3cm 3.5cm},clip,width=0.125\textwidth]{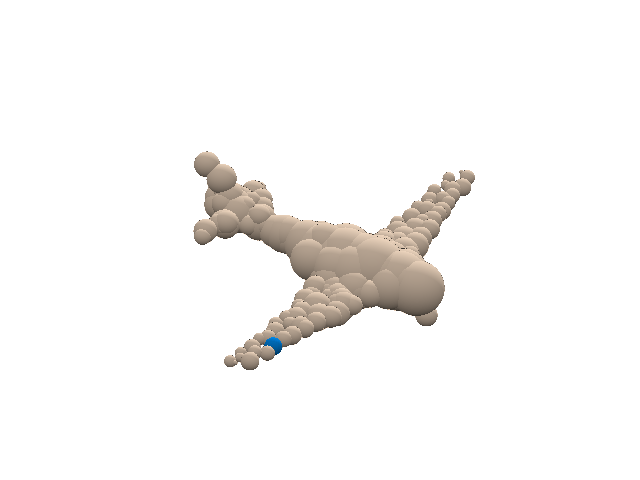}
\includegraphics[trim={2.9cm 3.6cm 3.3cm 3.5cm},clip,width=0.125\textwidth]{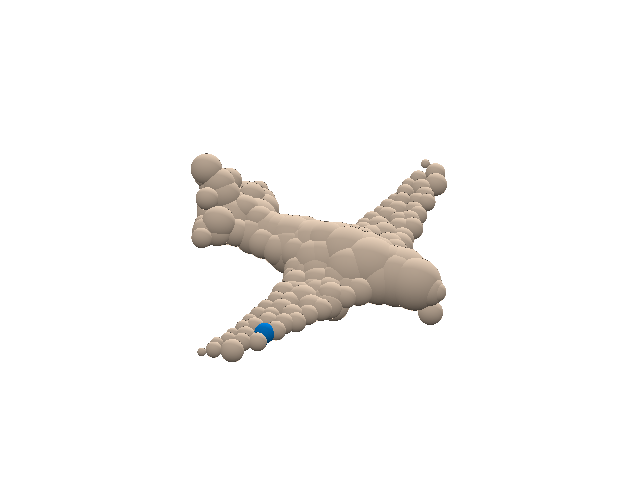}
\includegraphics[trim={2.9cm 3.6cm 3.3cm 3.5cm},clip,width=0.125\textwidth]{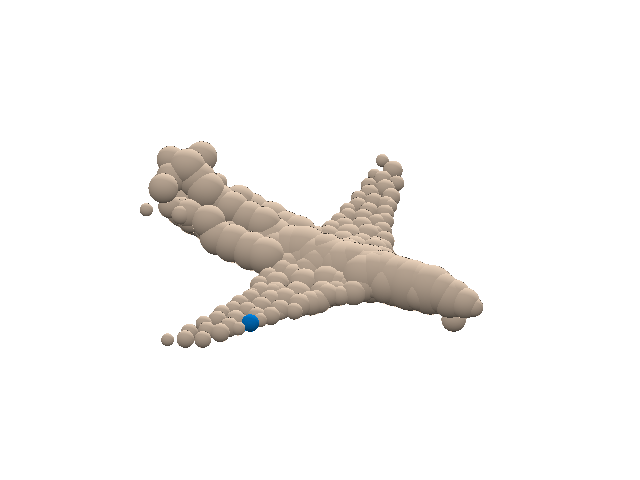} 
\includegraphics[trim={2.9cm 3.6cm 3.3cm 3.5cm},clip,width=0.125\textwidth]{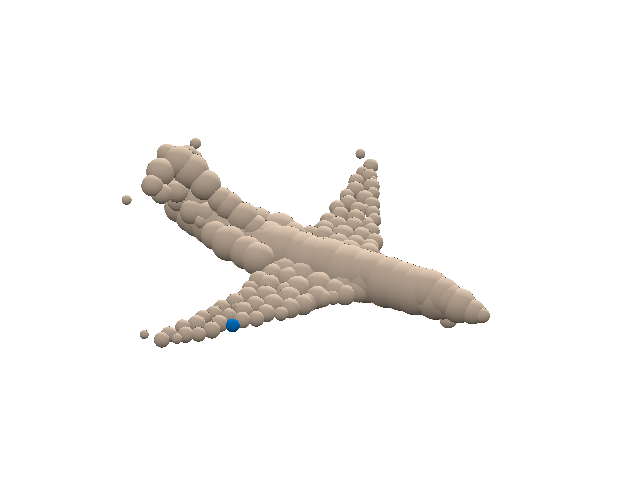}
\includegraphics[trim={2.9cm 3.6cm 3.3cm 3.5cm},clip,width=0.125\textwidth]{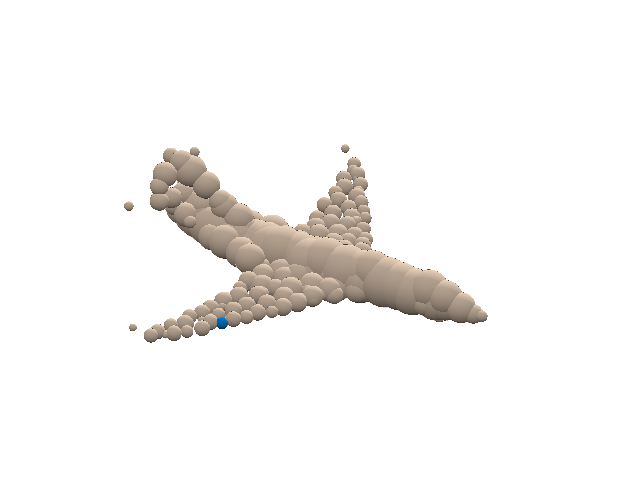}
\includegraphics[trim={2.9cm 3.6cm 3.3cm 3.5cm},clip,width=0.125\textwidth]{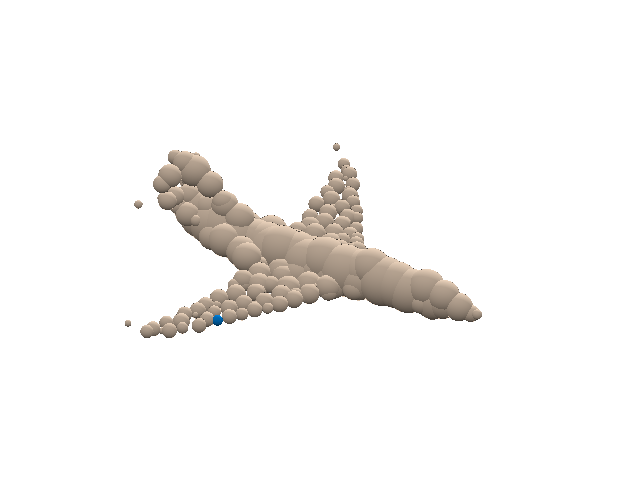} \\
\includegraphics[width=0.07\textwidth]{supp/figures/manip1/white.png}
\fbox{{\includegraphics[trim={2.9cm 3.6cm 3.3cm 3.5cm},clip,width=0.125\textwidth]{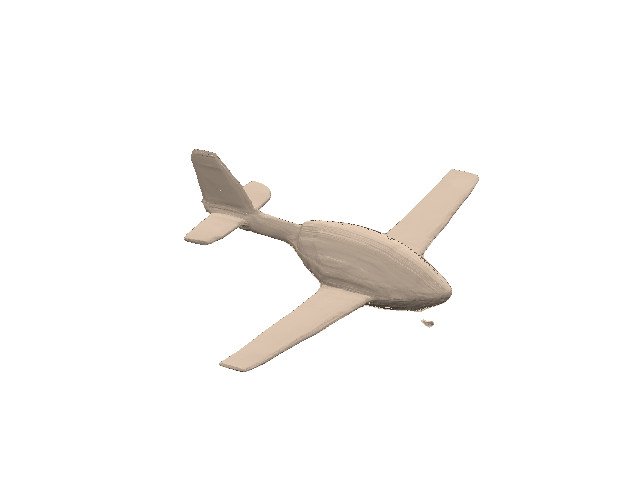}}}
\includegraphics[trim={2.9cm 3.6cm 3.3cm 3.5cm},clip,width=0.125\textwidth]{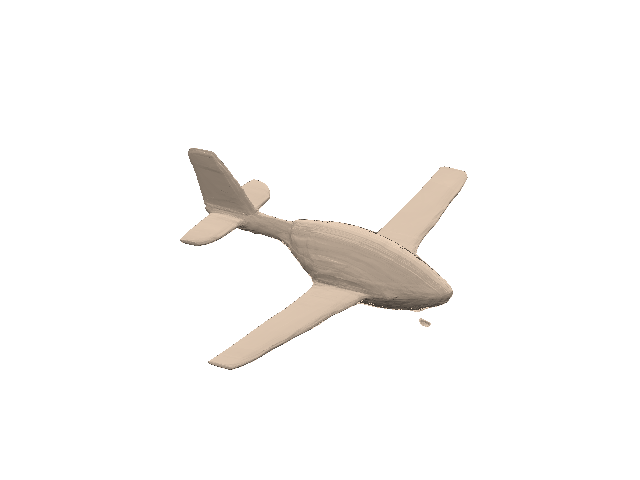}
\includegraphics[trim={2.9cm 3.6cm 3.3cm 3.5cm},clip,width=0.125\textwidth]{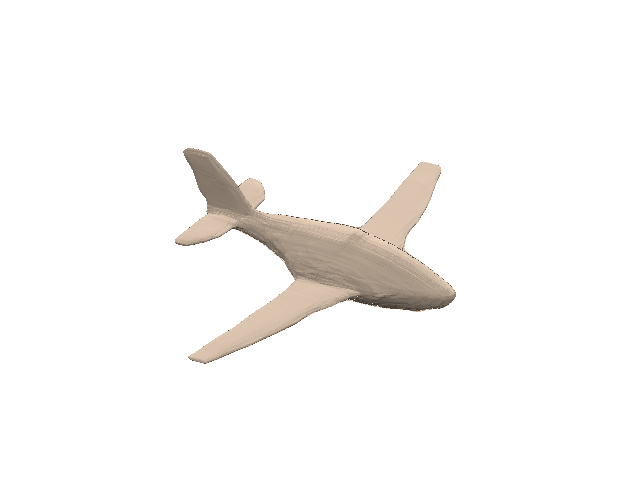}
\includegraphics[trim={2.9cm 3.6cm 3.3cm 3.5cm},clip,width=0.125\textwidth]{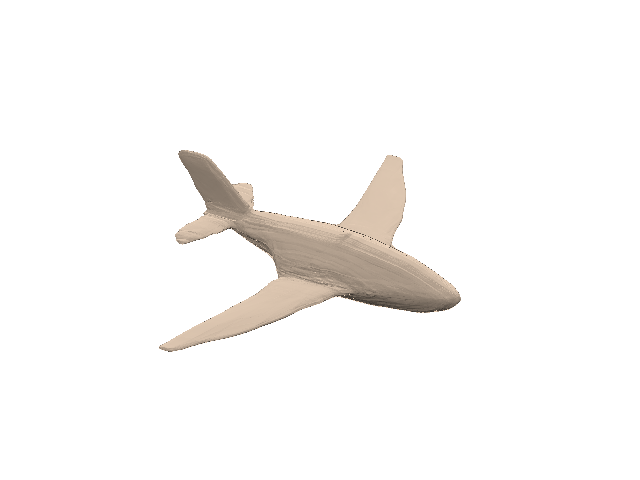} 
\includegraphics[trim={2.9cm 3.6cm 3.3cm 3.5cm},clip,width=0.125\textwidth]{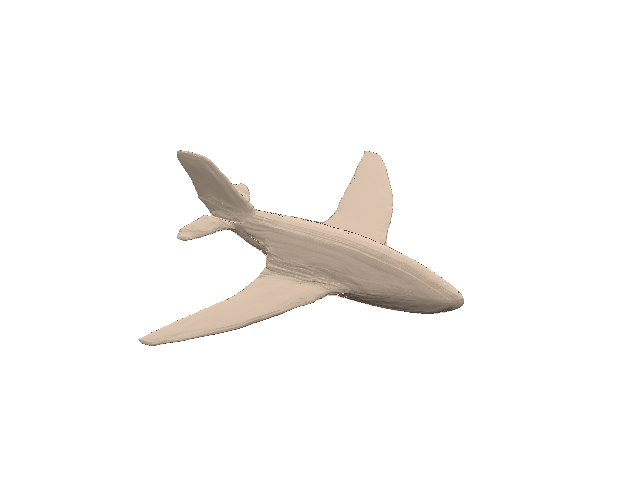}
\includegraphics[trim={2.9cm 3.6cm 3.3cm 3.5cm},clip,width=0.125\textwidth]{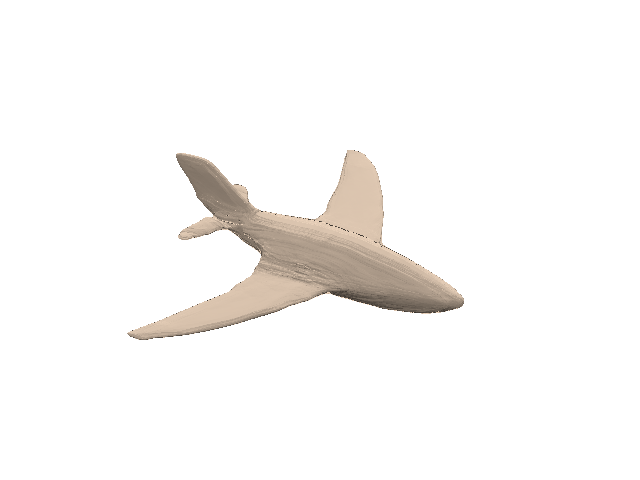}
\includegraphics[trim={2.9cm 3.6cm 3.3cm 3.5cm},clip,width=0.125\textwidth]{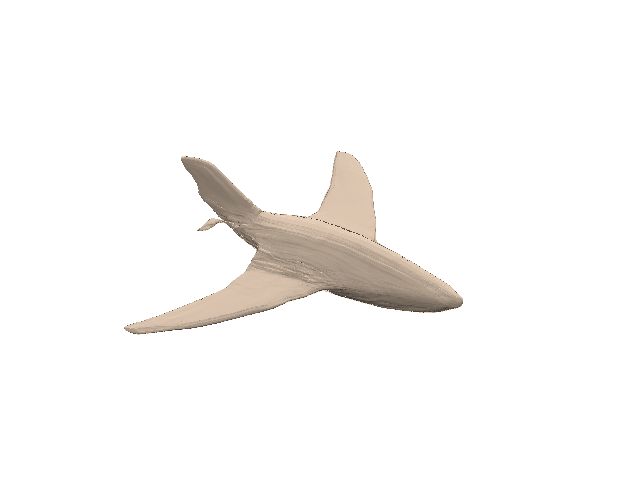}\\
\includegraphics[width=0.07\textwidth]{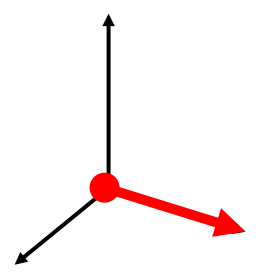}
\fbox{{\includegraphics[trim={2.9cm 3.6cm 3.3cm 3.5cm},clip,width=0.125\textwidth]{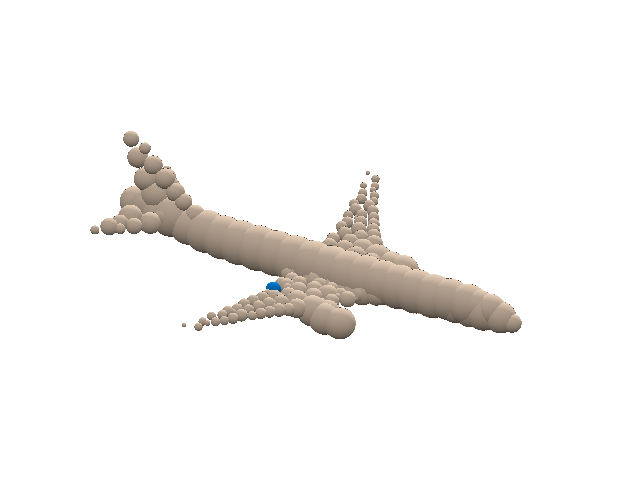}}}
\includegraphics[trim={2.9cm 3.6cm 3.3cm 3.5cm},clip,width=0.125\textwidth]{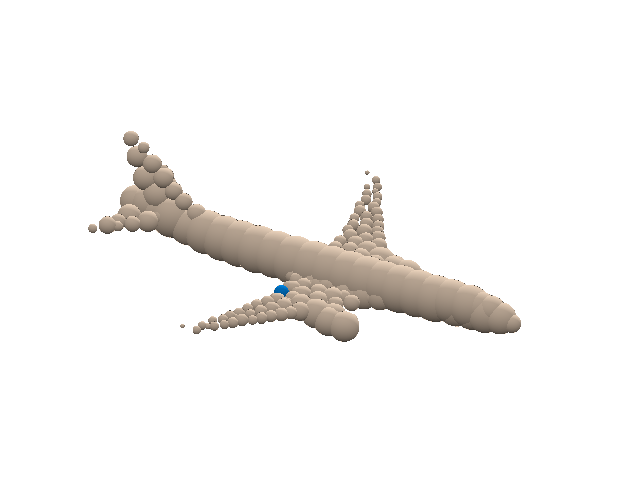}
\includegraphics[trim={2.9cm 3.6cm 3.3cm 3.5cm},clip,width=0.125\textwidth]{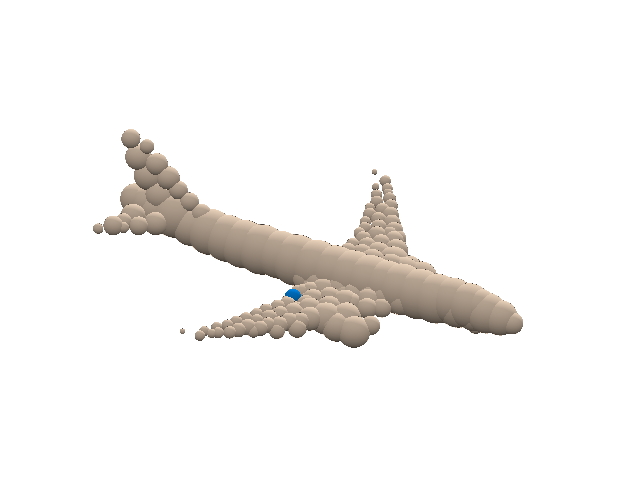}
\includegraphics[trim={2.9cm 3.6cm 3.3cm 3.5cm},clip,width=0.125\textwidth]{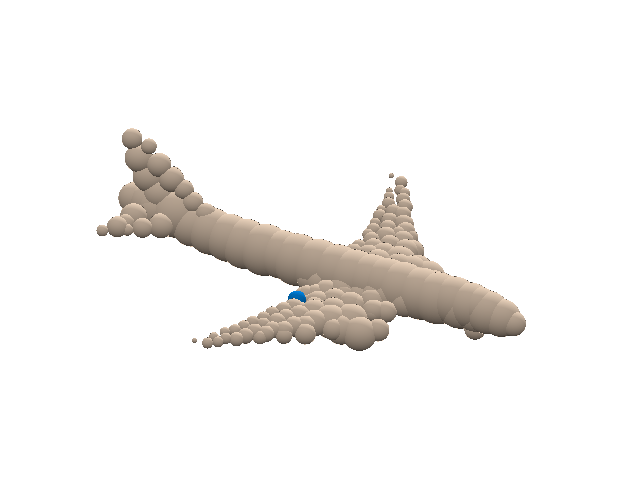} 
\includegraphics[trim={2.9cm 3.6cm 3.3cm 3.5cm},clip,width=0.125\textwidth]{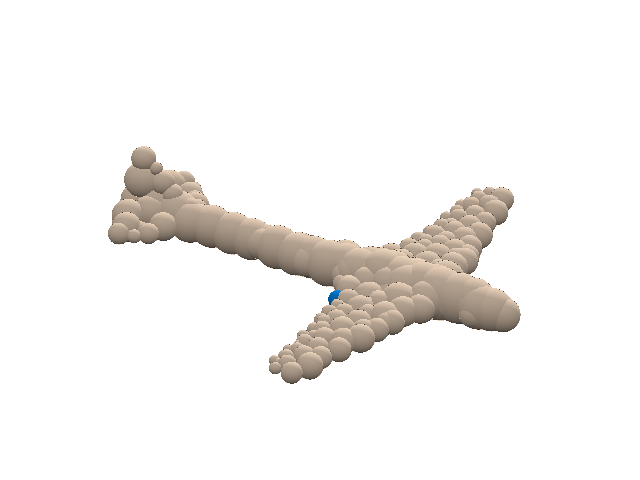}
\includegraphics[trim={2.9cm 3.6cm 3.3cm 3.5cm},clip,width=0.125\textwidth]{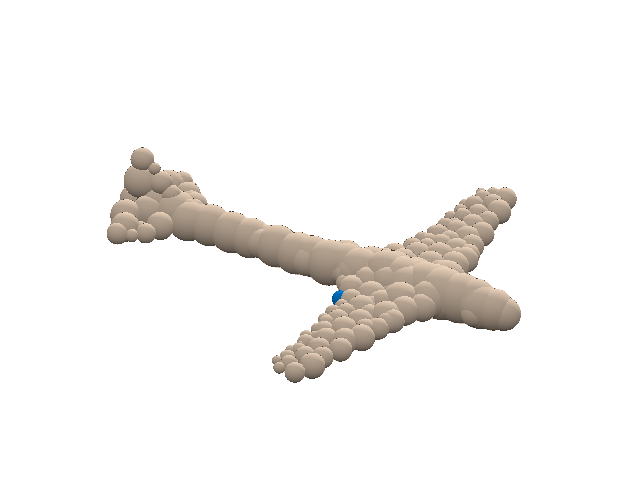}
\includegraphics[trim={2.9cm 3.6cm 3.3cm 3.5cm},clip,width=0.125\textwidth]{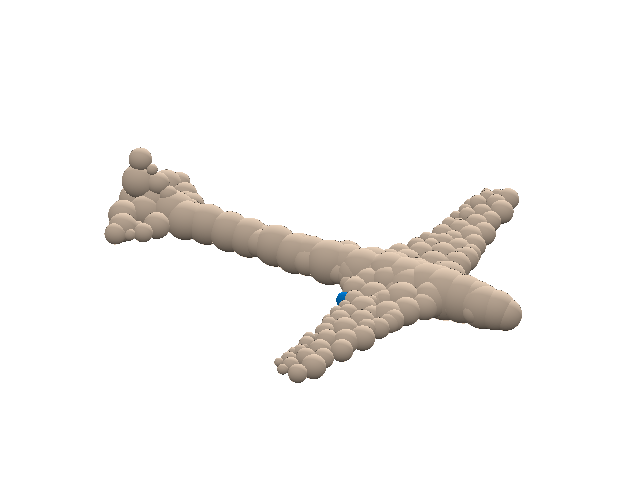} \\
\includegraphics[width=0.07\textwidth]{supp/figures/manip1/white.png}
\fbox{{\includegraphics[trim={2.9cm 3.6cm 3.3cm 3.5cm},clip,width=0.125\textwidth]{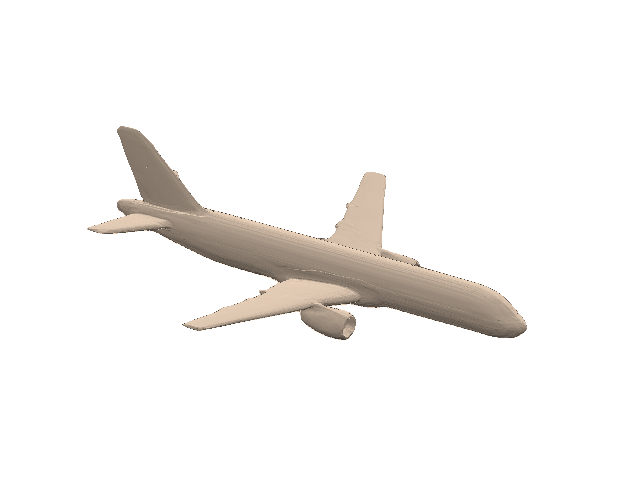}}}
\includegraphics[trim={2.9cm 3.6cm 3.3cm 3.5cm},clip,width=0.125\textwidth]{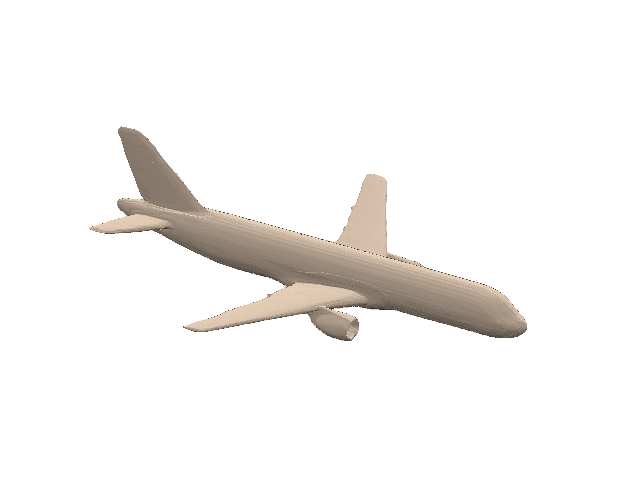}
\includegraphics[trim={2.9cm 3.6cm 3.3cm 3.5cm},clip,width=0.125\textwidth]{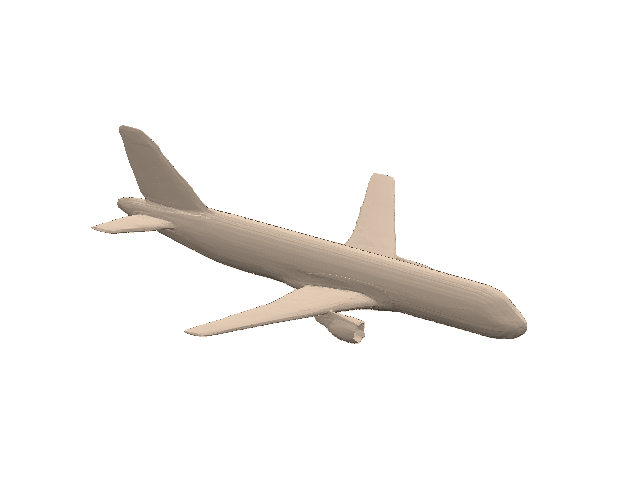}
\includegraphics[trim={2.9cm 3.6cm 3.3cm 3.5cm},clip,width=0.125\textwidth]{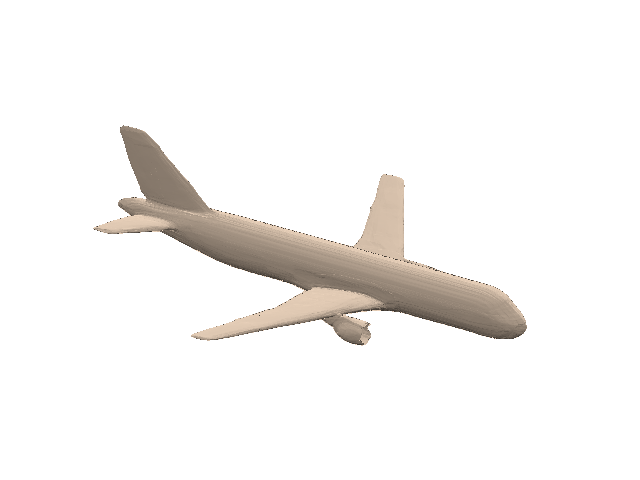} 
\includegraphics[trim={2.9cm 3.6cm 3.3cm 3.5cm},clip,width=0.125\textwidth]{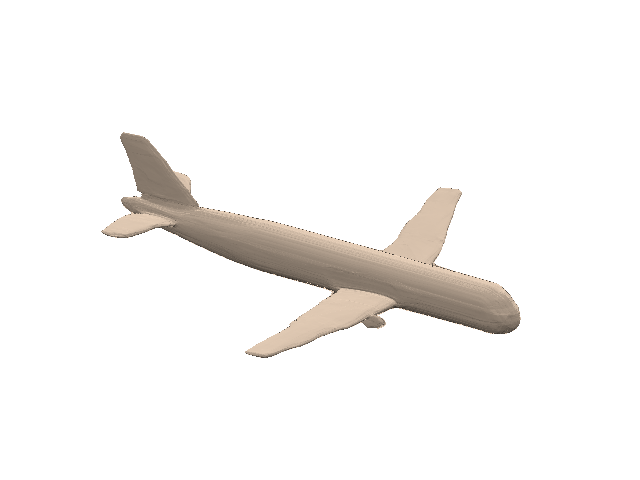}
\includegraphics[trim={2.9cm 3.6cm 3.3cm 3.5cm},clip,width=0.125\textwidth]{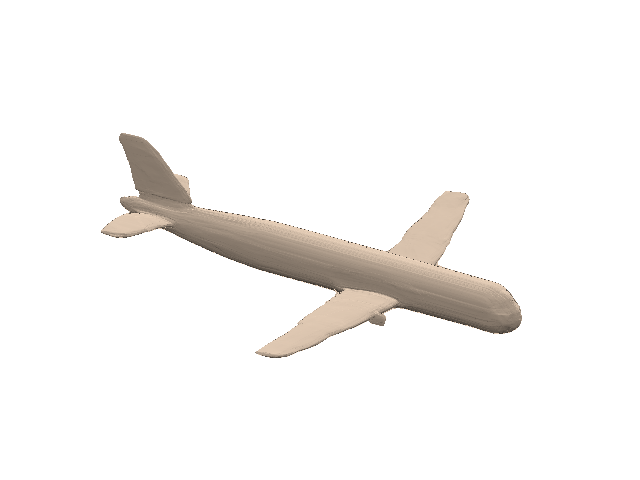}
\includegraphics[trim={2.9cm 3.6cm 3.3cm 3.5cm},clip,width=0.125\textwidth]{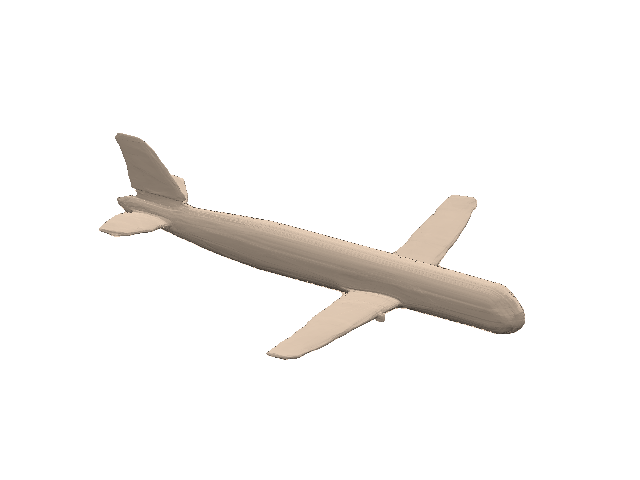}\\
\vspace{-7pt}
  \caption{Manipulating shapes (on the left) by dragging a single primitive (colored in blue) along a specified direction (red arrow on the left). Please refer to the accompanying video for full manipulated sequences.
  }
      \label{fig:manipulation}
\end{figure*}
\begin{figure*}
  \centering
  \setlength{\fboxrule}{2.0pt}
\fbox{{\includegraphics[trim={2.9cm 0.6cm 3.3cm 0.5cm},clip,width=0.116\textwidth]{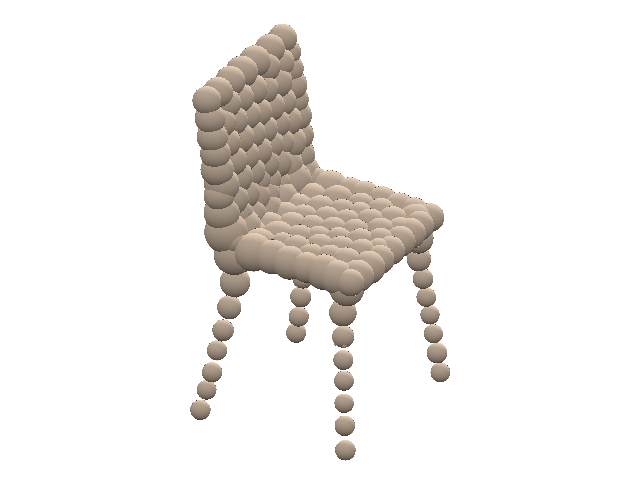}}}
\includegraphics[trim={2.9cm 0.6cm 3.3cm 0.5cm},clip,width=0.116\textwidth]{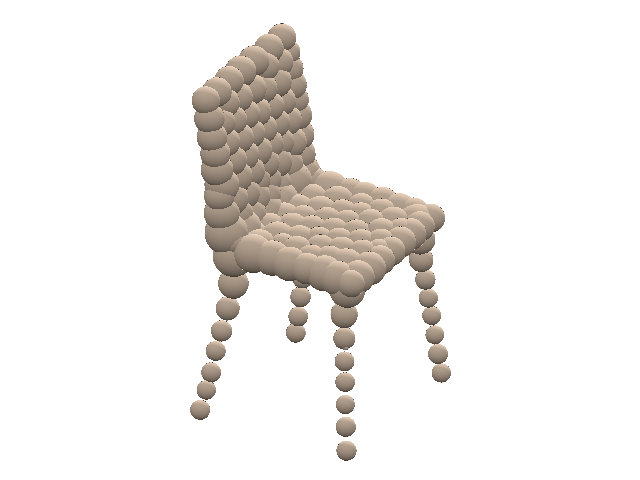}
\includegraphics[trim={2.9cm 0.6cm 3.3cm 0.5cm},clip,width=0.116\textwidth]{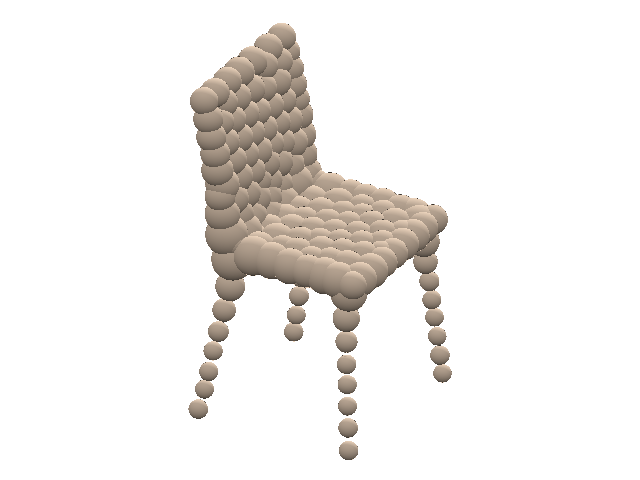}
\includegraphics[trim={2.9cm 0.6cm 3.3cm 0.5cm},clip,width=0.116\textwidth]{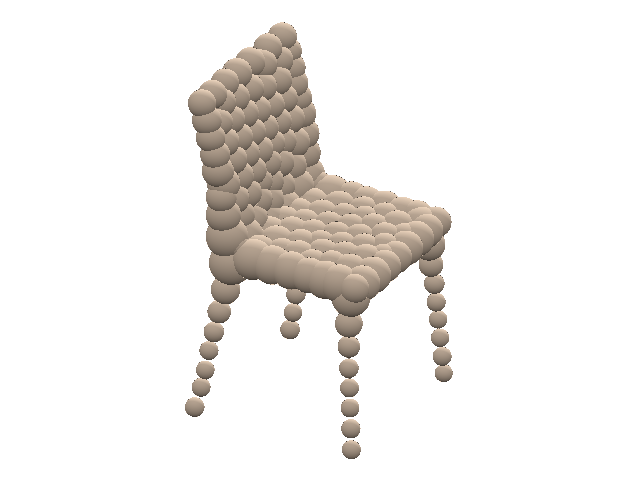}
\includegraphics[trim={2.9cm 0.6cm 3.3cm 0.5cm},clip,width=0.116\textwidth]{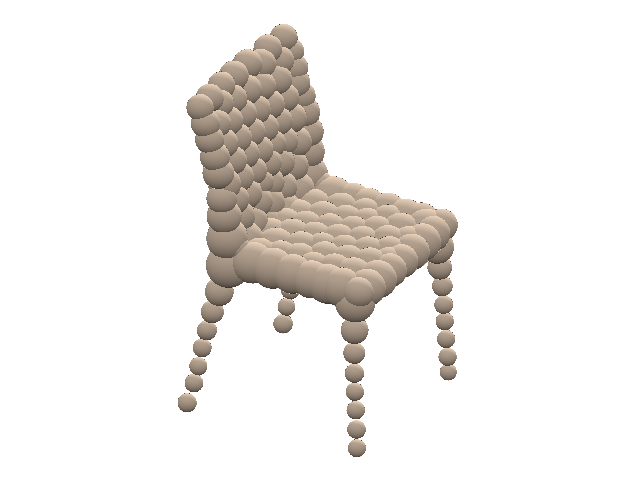}
\includegraphics[trim={2.9cm 0.6cm 3.3cm 0.5cm},clip,width=0.116\textwidth]{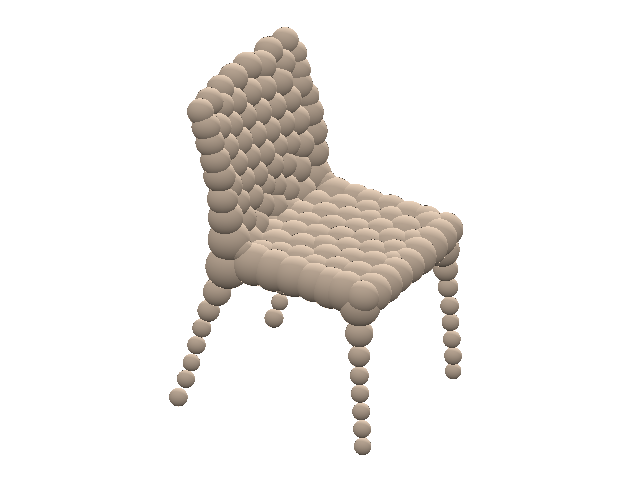}
\includegraphics[trim={2.9cm 0.6cm 3.3cm 0.5cm},clip,width=0.116\textwidth]{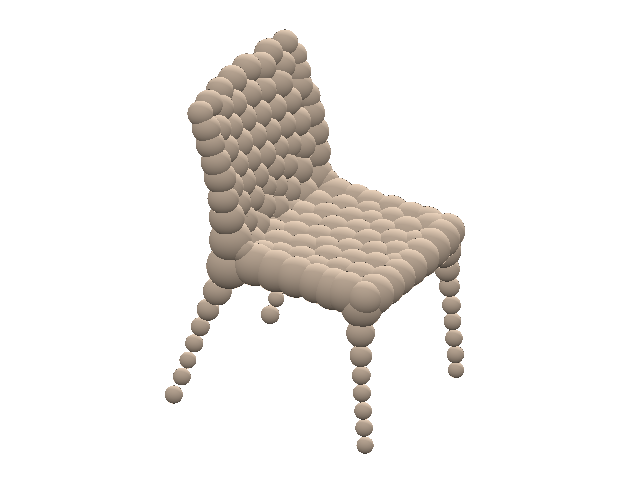}
\fbox{\includegraphics[trim={2.9cm 0.6cm 3.3cm 0.5cm},clip,width=0.116\textwidth]{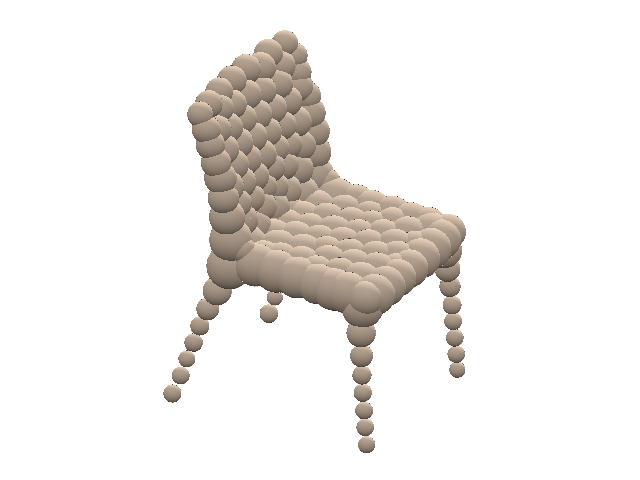}} \\
\fbox{{\includegraphics[trim={2.9cm 0.6cm 3.3cm 0.5cm},clip,width=0.116\textwidth]{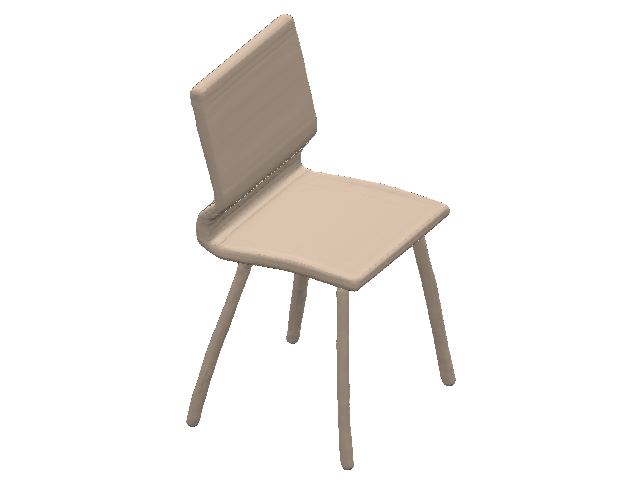}}}
\includegraphics[trim={2.9cm 0.6cm 3.3cm 0.5cm},clip,width=0.116\textwidth]{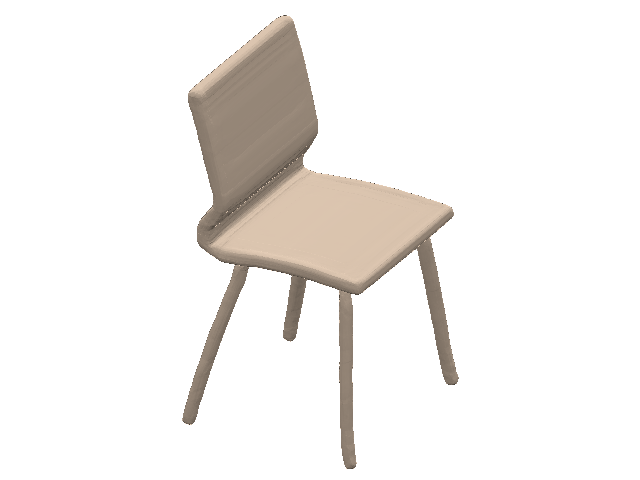}
\includegraphics[trim={2.9cm 0.6cm 3.3cm 0.5cm},clip,width=0.116\textwidth]{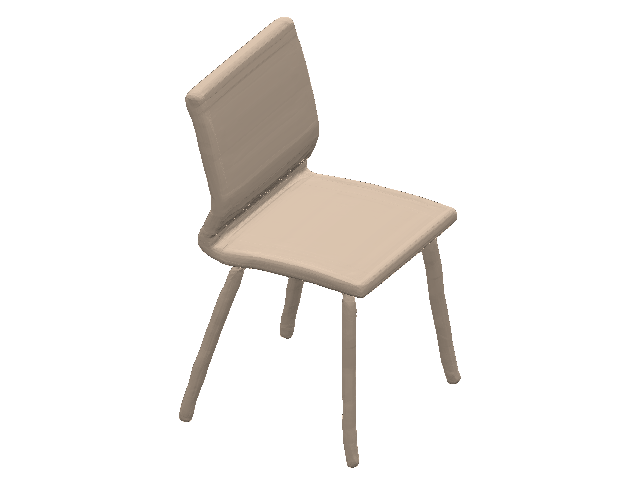}
\includegraphics[trim={2.9cm 0.6cm 3.3cm 0.5cm},clip,width=0.116\textwidth]{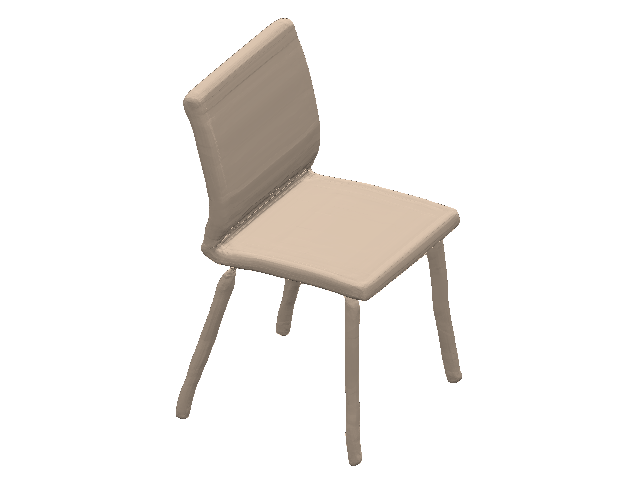}
\includegraphics[trim={2.9cm 0.6cm 3.3cm 0.5cm},clip,width=0.116\textwidth]{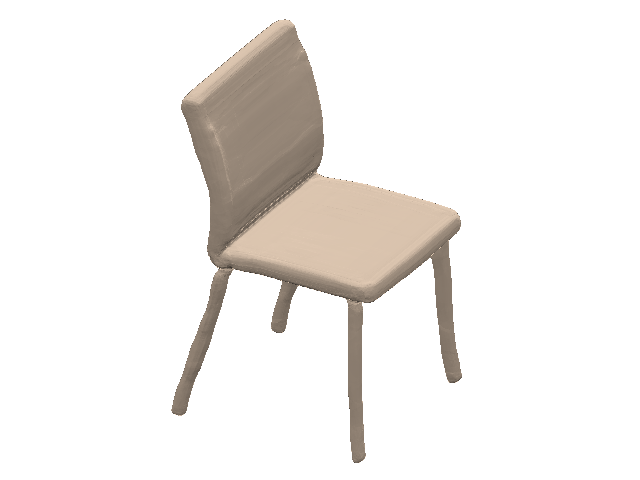}
\includegraphics[trim={2.9cm 0.6cm 3.3cm 0.5cm},clip,width=0.116\textwidth]{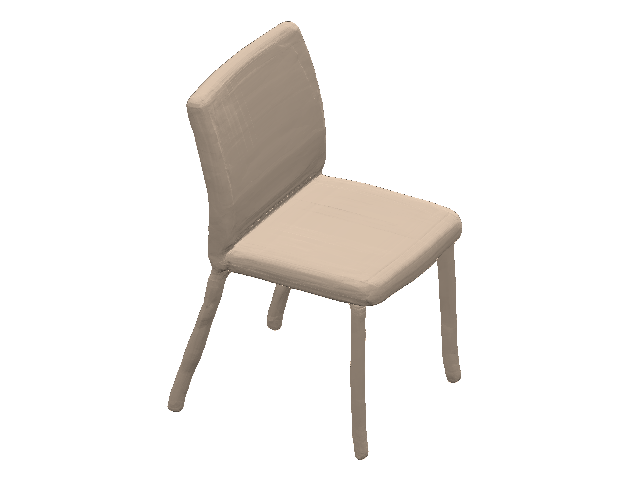}
\includegraphics[trim={2.9cm 0.6cm 3.3cm 0.5cm},clip,width=0.116\textwidth]{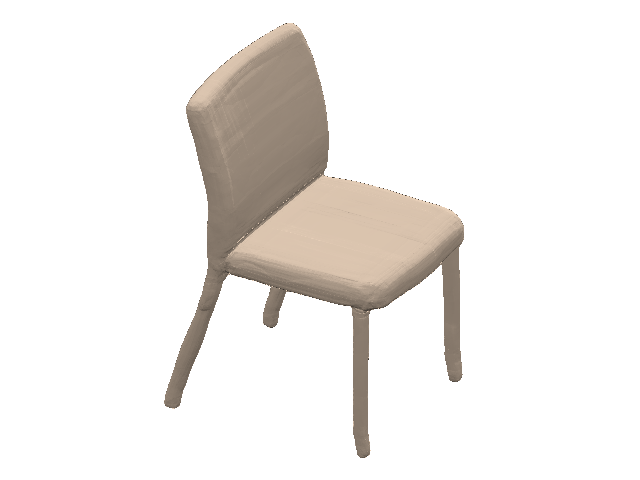}
\fbox{\includegraphics[trim={2.9cm 0.6cm 3.3cm 0.5cm},clip,width=0.116\textwidth]{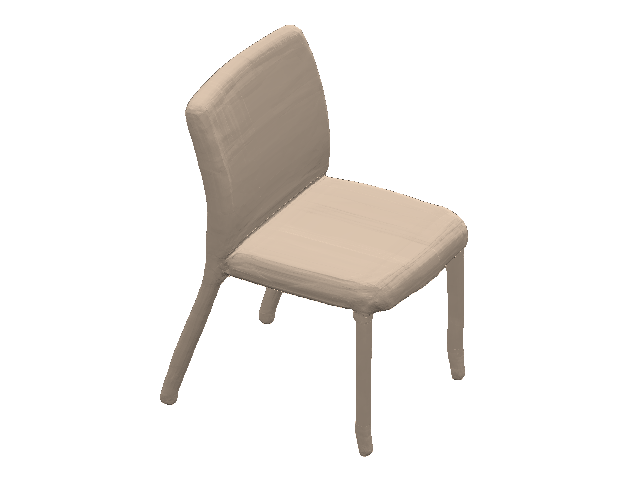}} \\
\fbox{{\includegraphics[trim={2.9cm 0.6cm 3.3cm 0.5cm},clip,width=0.116\textwidth]{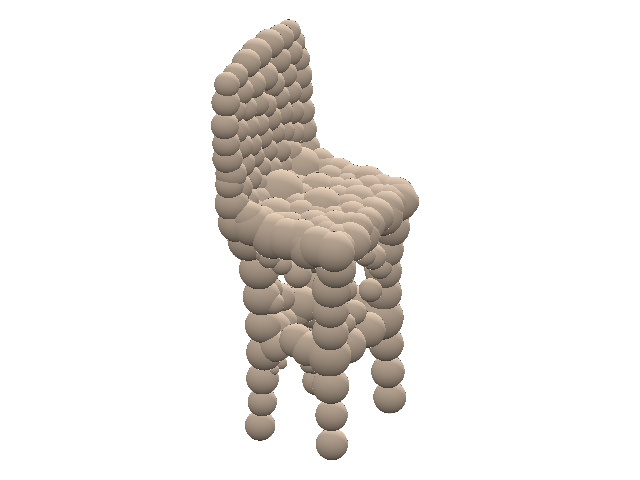}}}
\includegraphics[trim={2.9cm 0.6cm 3.3cm 0.5cm},clip,width=0.116\textwidth]{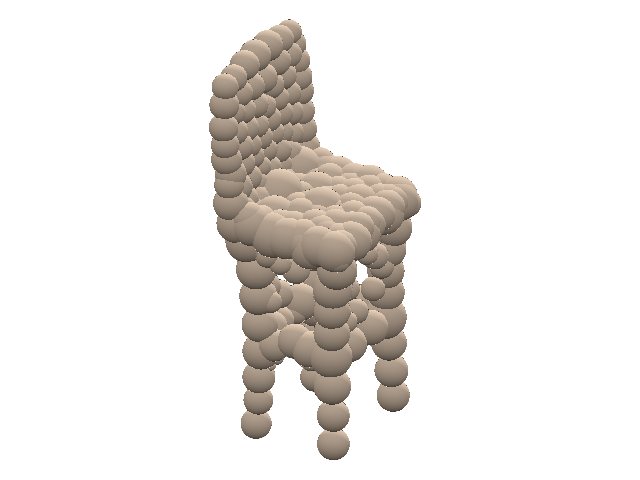}
\includegraphics[trim={2.9cm 0.6cm 3.3cm 0.5cm},clip,width=0.116\textwidth]{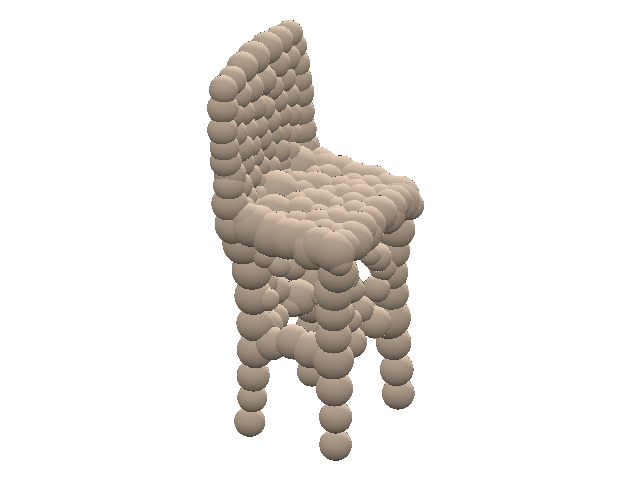}
\includegraphics[trim={2.9cm 0.6cm 3.3cm 0.5cm},clip,width=0.116\textwidth]{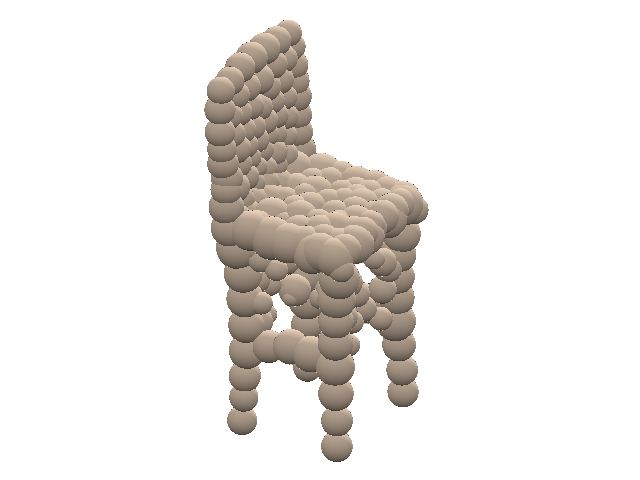}
\includegraphics[trim={2.9cm 0.6cm 3.3cm 0.5cm},clip,width=0.116\textwidth]{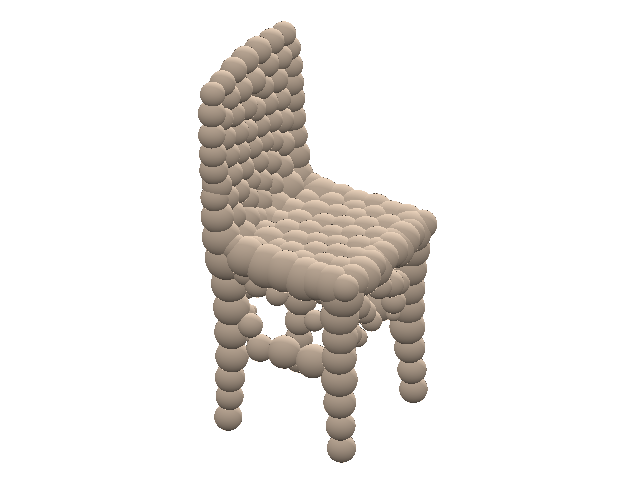}
\includegraphics[trim={2.9cm 0.6cm 3.3cm 0.5cm},clip,width=0.116\textwidth]{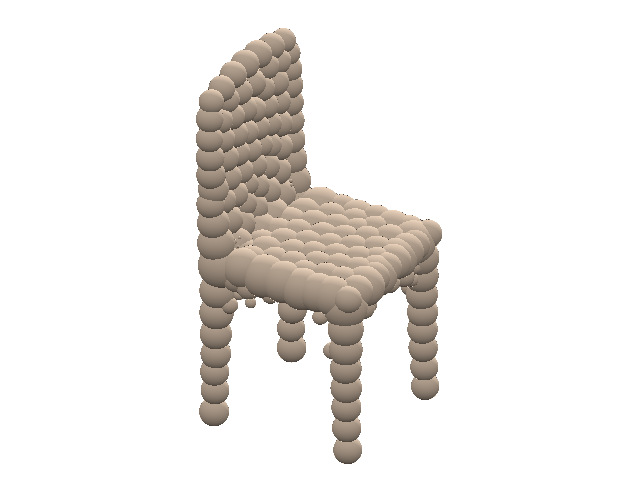}
\includegraphics[trim={2.9cm 0.6cm 3.3cm 0.5cm},clip,width=0.116\textwidth]{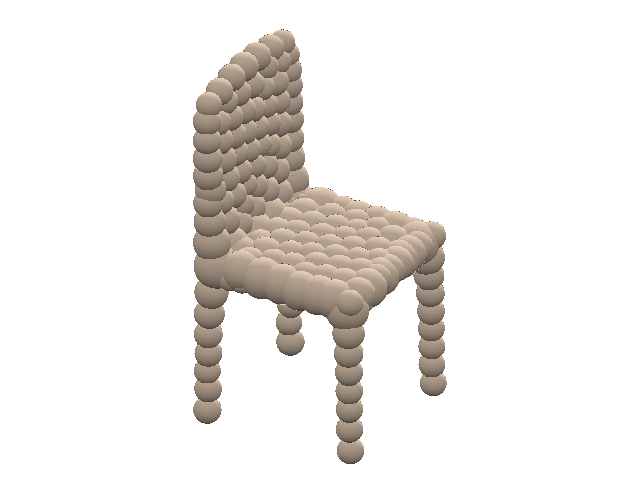}
\fbox{\includegraphics[trim={2.9cm 0.6cm 3.3cm 0.5cm},clip,width=0.116\textwidth]{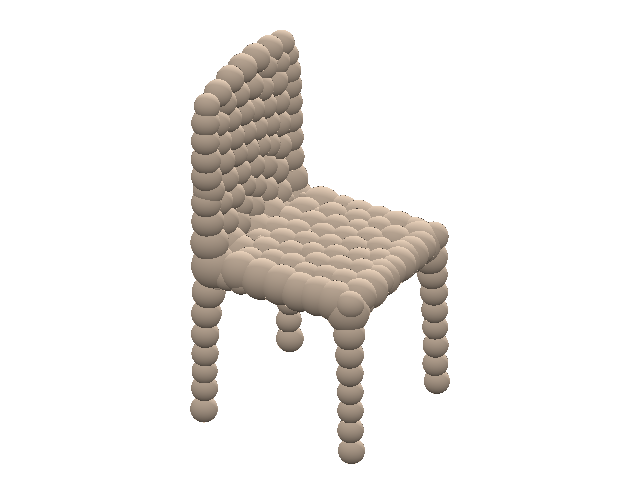}}
\\
\fbox{{\includegraphics[trim={2.9cm 0.6cm 3.3cm 0.5cm},clip,width=0.116\textwidth]{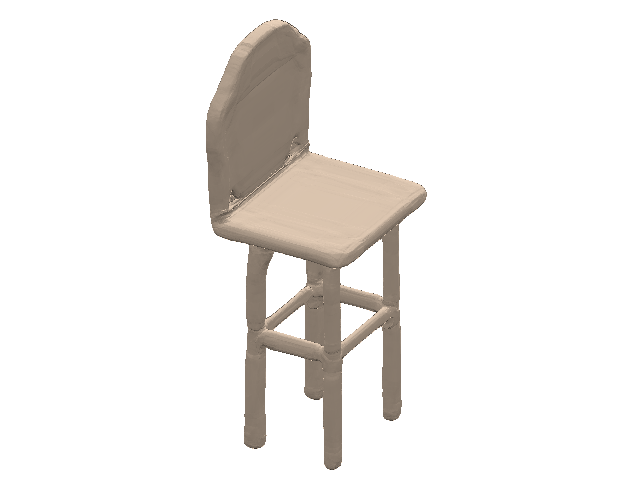}}}
\includegraphics[trim={2.9cm 0.6cm 3.3cm 0.5cm},clip,width=0.116\textwidth]{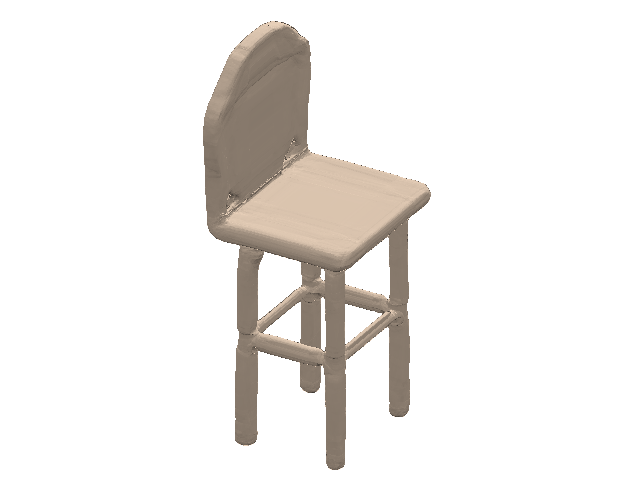}
\includegraphics[trim={2.9cm 0.6cm 3.3cm 0.5cm},clip,width=0.116\textwidth]{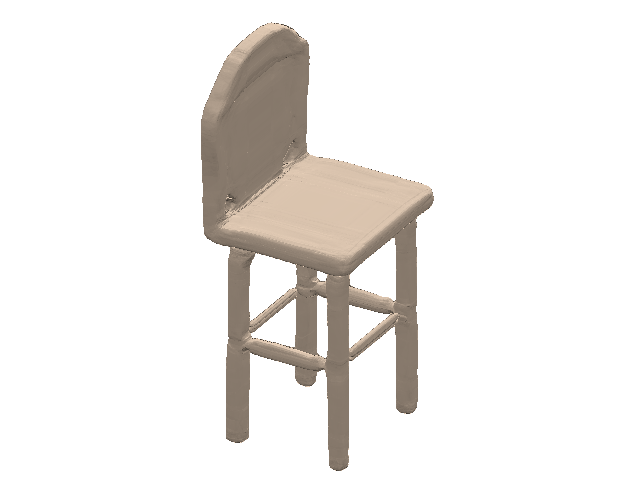}
\includegraphics[trim={2.9cm 0.6cm 3.3cm 0.5cm},clip,width=0.116\textwidth]{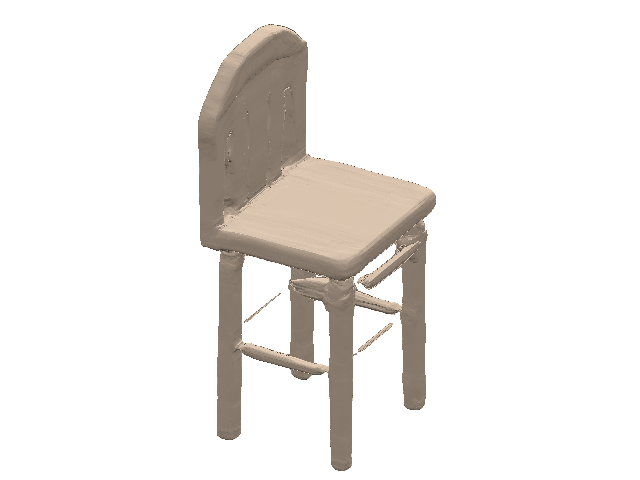}
\includegraphics[trim={2.9cm 0.6cm 3.3cm 0.5cm},clip,width=0.116\textwidth]{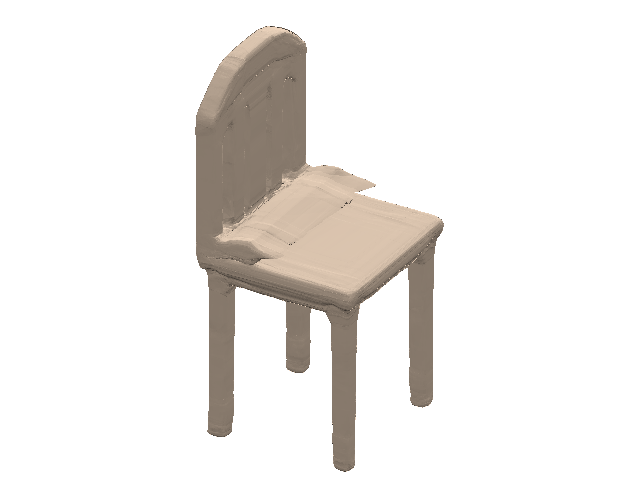}
\includegraphics[trim={2.9cm 0.6cm 3.3cm 0.5cm},clip,width=0.116\textwidth]{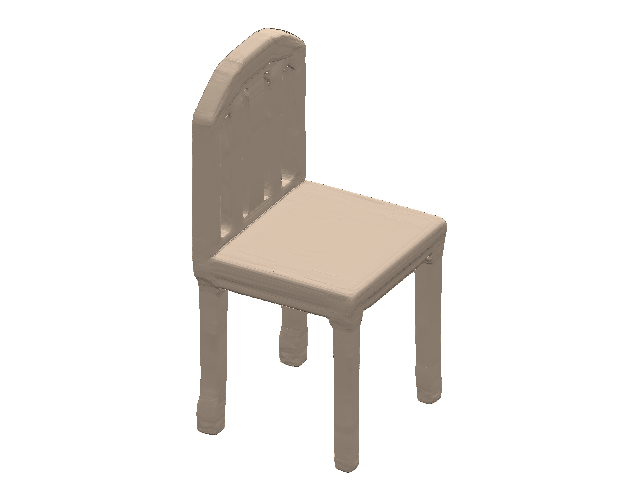}
\includegraphics[trim={2.9cm 0.6cm 3.3cm 0.5cm},clip,width=0.116\textwidth]{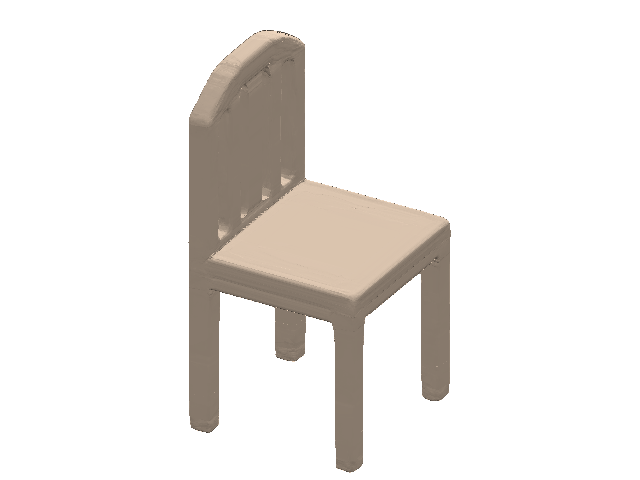}
\fbox{\includegraphics[trim={2.9cm 0.6cm 3.3cm 0.5cm},clip,width=0.116\textwidth]{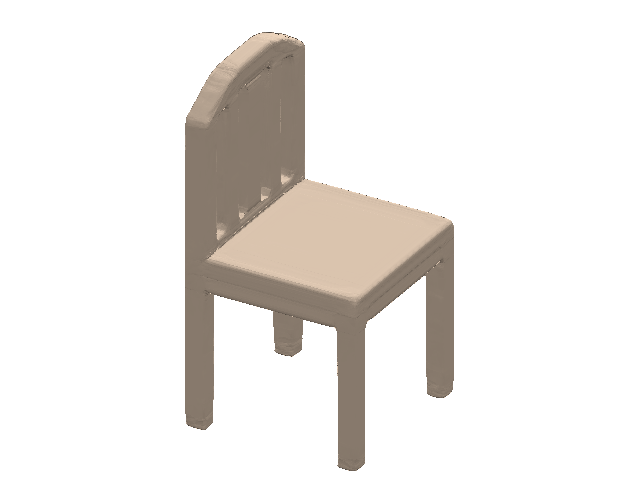}}
  \caption{Linear interpolation in the latent space between pairs of shapes. The original shape pairs are marked with red boxes while the shapes in between are generated from interpolated latent code. Accompanying each shape is its corresponding primitive-based representation.
  }
      \label{fig:interp_reg}
\end{figure*}
\begin{figure*}
  \centering
  \setlength{\fboxrule}{2.0pt}
\fbox{{\includegraphics[trim={2.9cm 0.6cm 3.3cm 0.5cm},clip,width=0.116\textwidth]{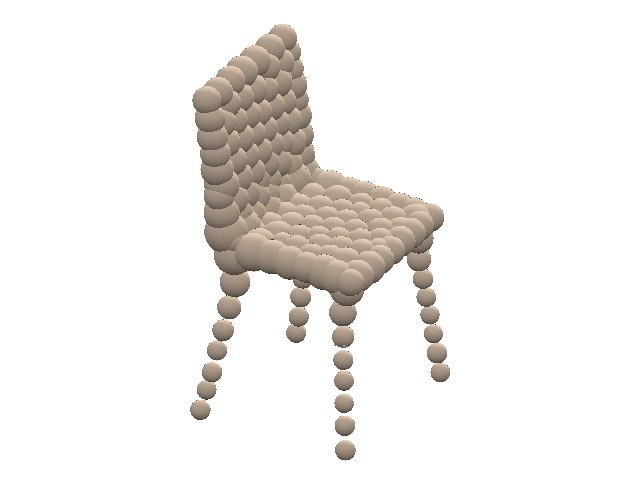}}}
\includegraphics[trim={2.9cm 0.6cm 3.3cm 0.5cm},clip,width=0.116\textwidth]{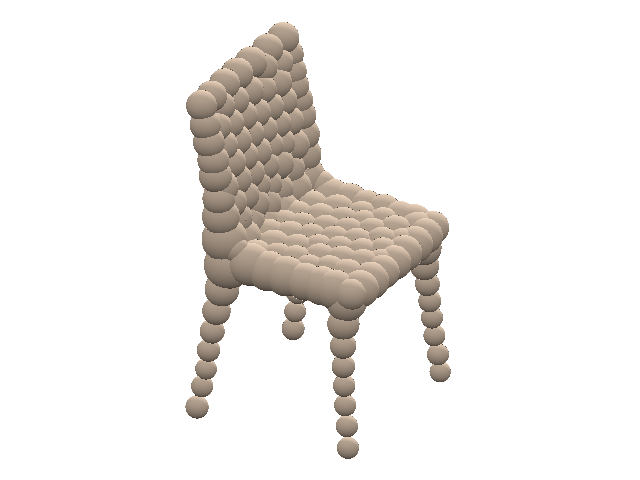}
\includegraphics[trim={2.9cm 0.6cm 3.3cm 0.5cm},clip,width=0.116\textwidth]{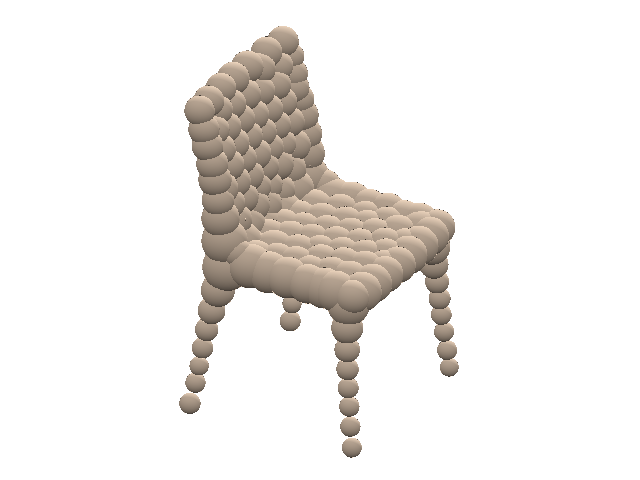}
\includegraphics[trim={2.9cm 0.6cm 3.3cm 0.5cm},clip,width=0.116\textwidth]{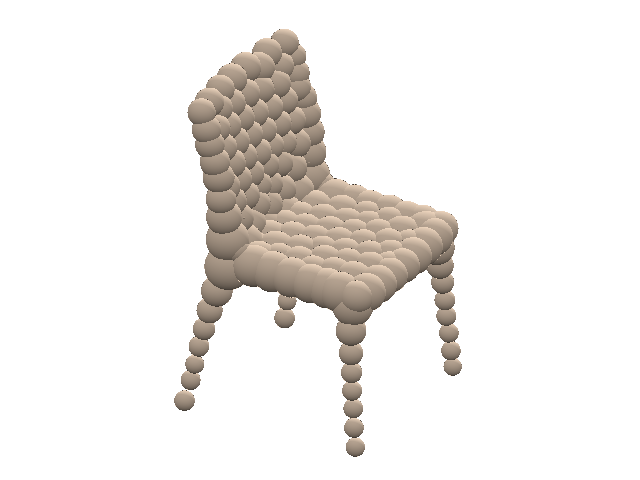}
\includegraphics[trim={2.9cm 0.6cm 3.3cm 0.5cm},clip,width=0.116\textwidth]{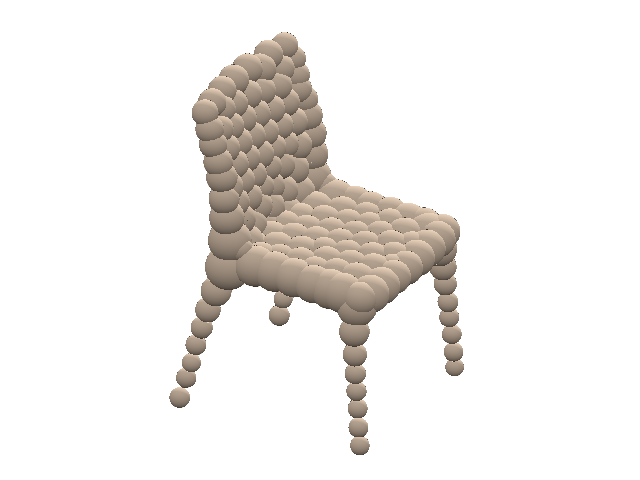}
\includegraphics[trim={2.9cm 0.6cm 3.3cm 0.5cm},clip,width=0.116\textwidth]{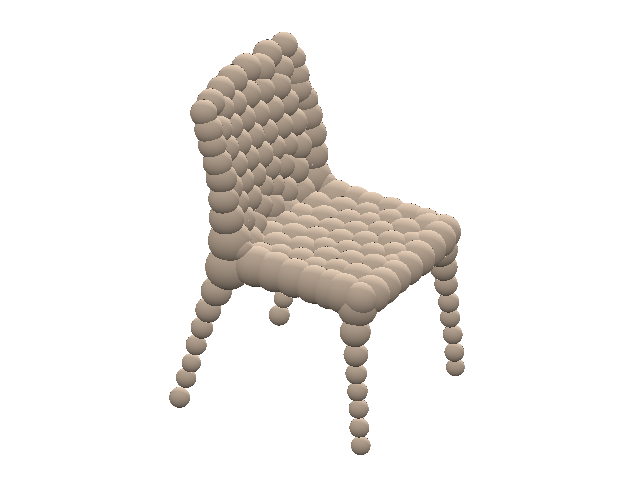}
\includegraphics[trim={2.9cm 0.6cm 3.3cm 0.5cm},clip,width=0.116\textwidth]{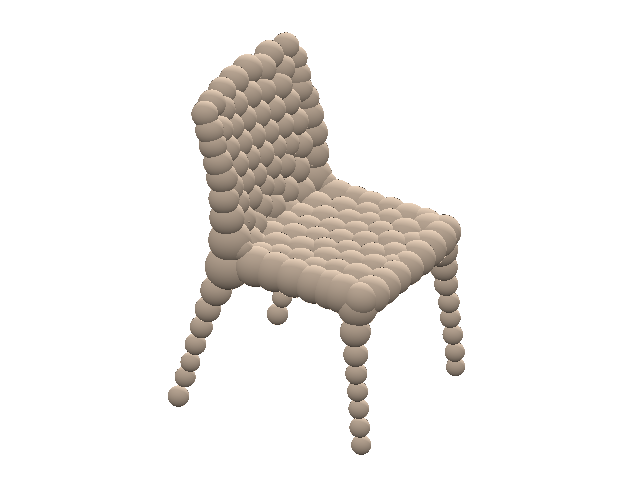}
\fbox{{\includegraphics[trim={2.9cm 0.6cm 3.3cm 0.5cm},clip,width=0.116\textwidth]{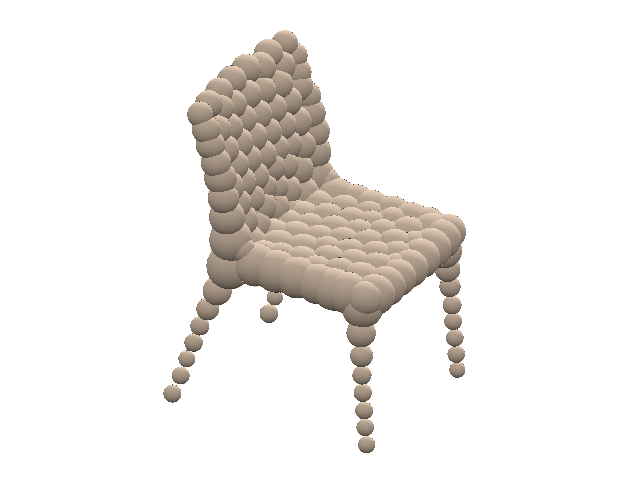}}}\\
\fbox{{\includegraphics[trim={2.9cm 0.6cm 3.3cm 0.5cm},clip,width=0.116\textwidth]{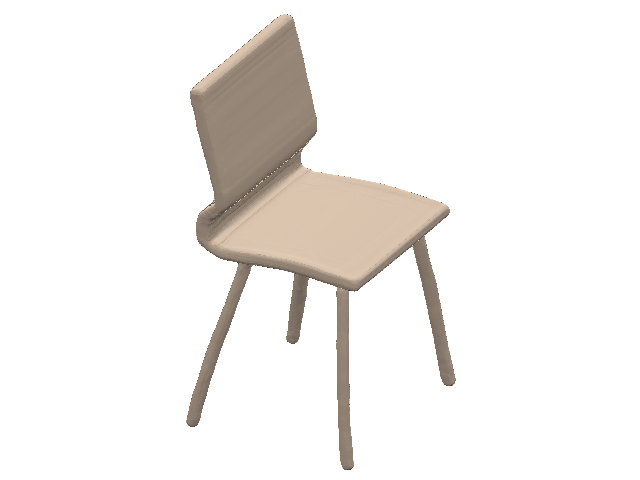}}}
\includegraphics[trim={2.9cm 0.6cm 3.3cm 0.5cm},clip,width=0.116\textwidth]{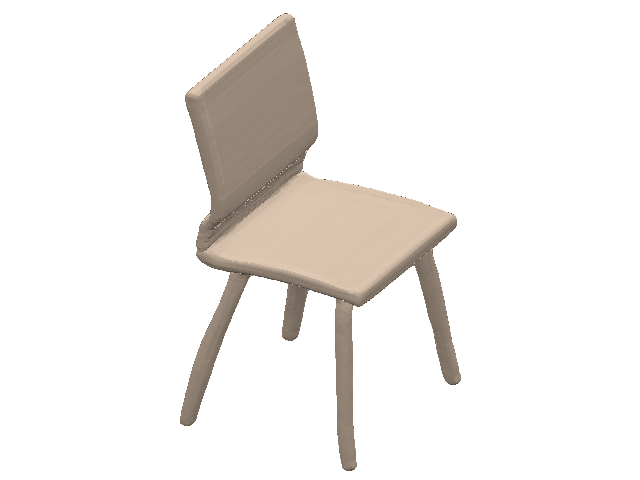}
\includegraphics[trim={2.9cm 0.6cm 3.3cm 0.5cm},clip,width=0.116\textwidth]{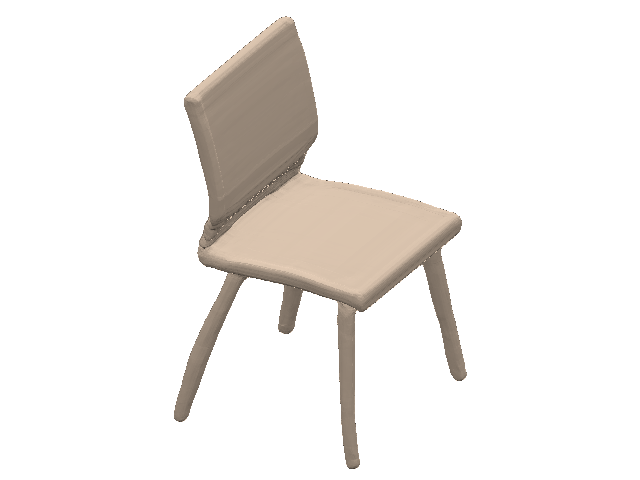}
\includegraphics[trim={2.9cm 0.6cm 3.3cm 0.5cm},clip,width=0.116\textwidth]{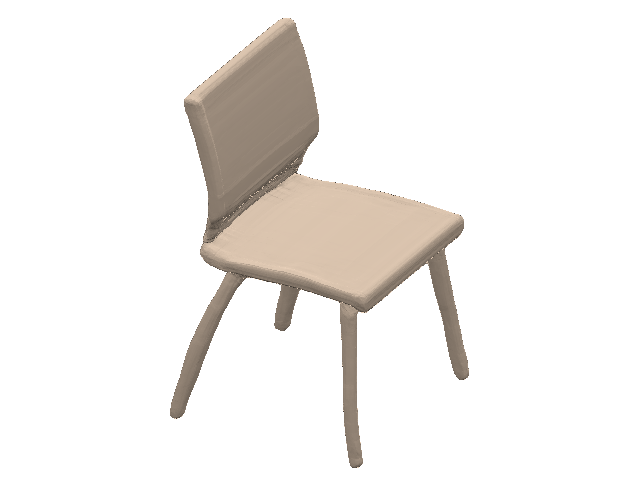}
\includegraphics[trim={2.9cm 0.6cm 3.3cm 0.5cm},clip,width=0.116\textwidth]{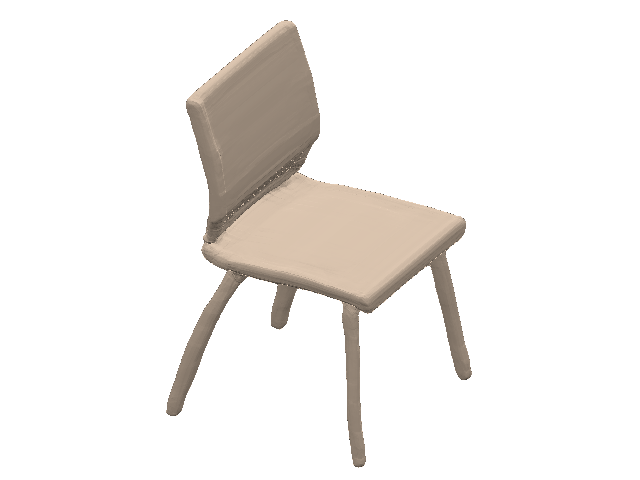}
\includegraphics[trim={2.9cm 0.6cm 3.3cm 0.5cm},clip,width=0.116\textwidth]{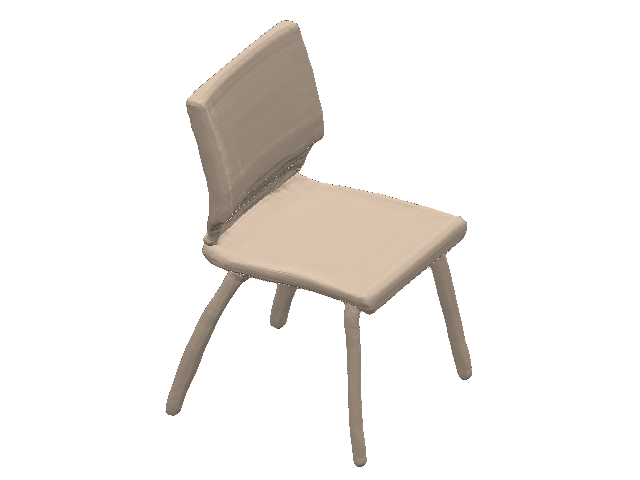}
\includegraphics[trim={2.9cm 0.6cm 3.3cm 0.5cm},clip,width=0.116\textwidth]{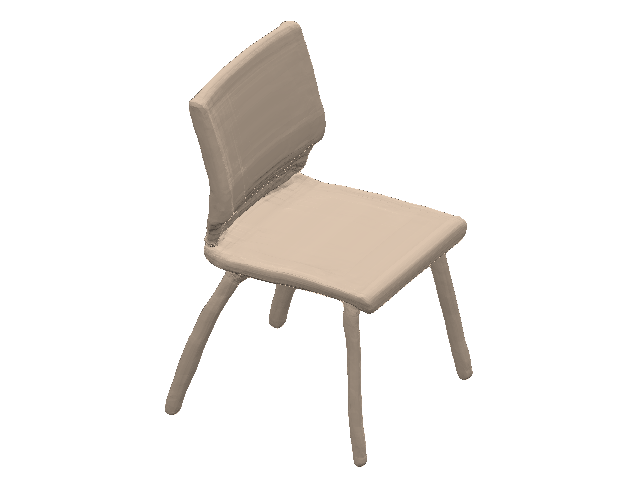} \fbox{{\includegraphics[trim={2.9cm 0.6cm 3.3cm 0.5cm},clip,width=0.116\textwidth]{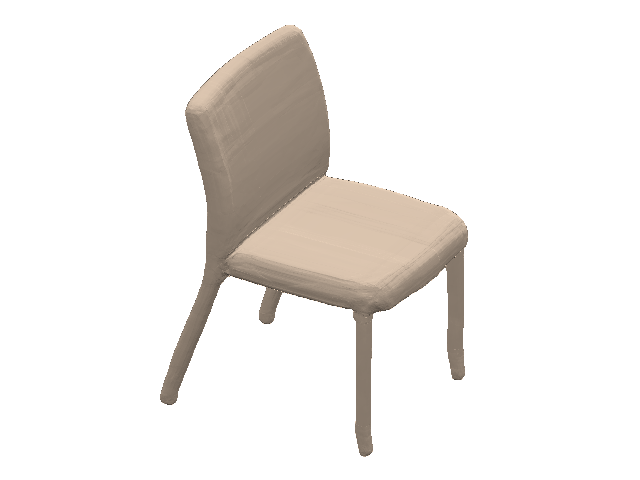}}}\\
\fbox{{\includegraphics[trim={2.9cm 0.6cm 3.3cm 0.5cm},clip,width=0.116\textwidth]{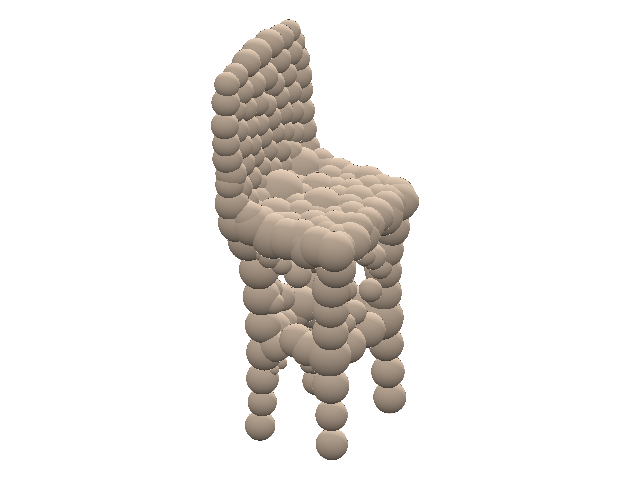}}}
\includegraphics[trim={2.9cm 0.6cm 3.3cm 0.5cm},clip,width=0.116\textwidth]{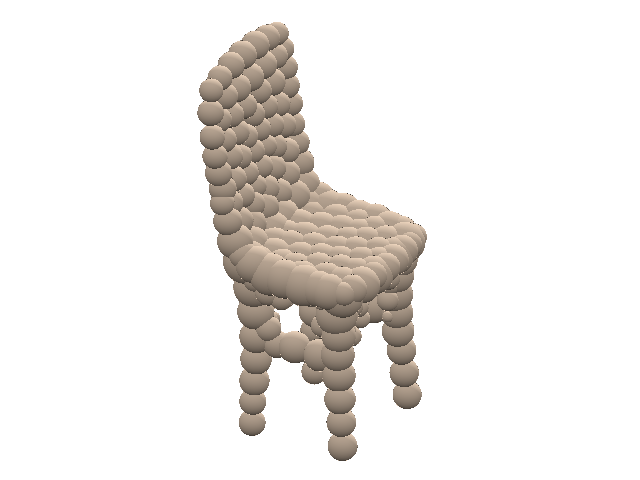}
\includegraphics[trim={2.9cm 0.6cm 3.3cm 0.5cm},clip,width=0.116\textwidth]{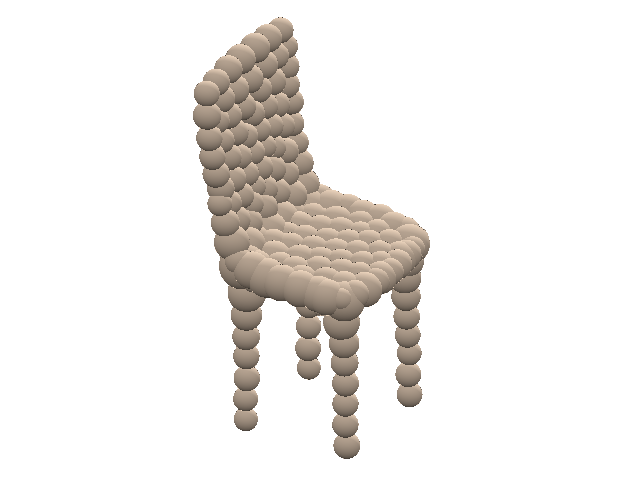}
\includegraphics[trim={2.9cm 0.6cm 3.3cm 0.5cm},clip,width=0.116\textwidth]{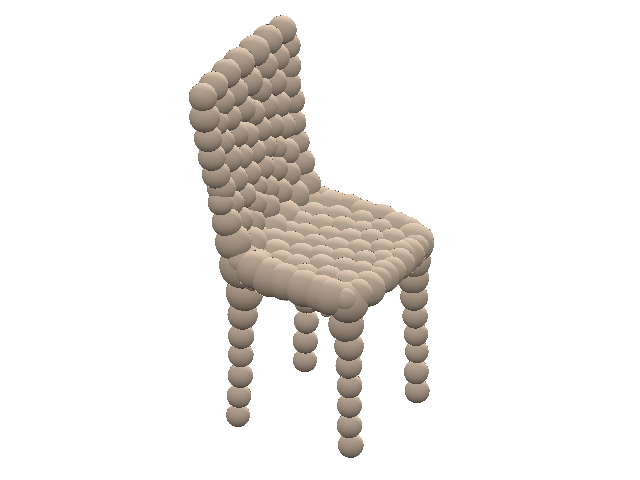}
\includegraphics[trim={2.9cm 0.6cm 3.3cm 0.5cm},clip,width=0.116\textwidth]{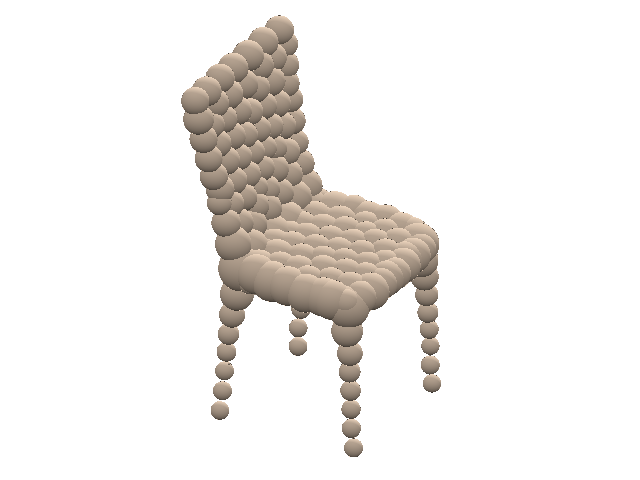}
\includegraphics[trim={2.9cm 0.6cm 3.3cm 0.5cm},clip,width=0.116\textwidth]{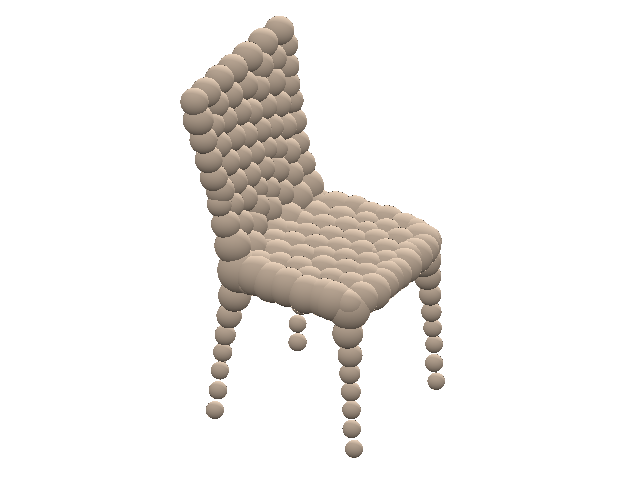}
\includegraphics[trim={2.9cm 0.6cm 3.3cm 0.5cm},clip,width=0.116\textwidth]{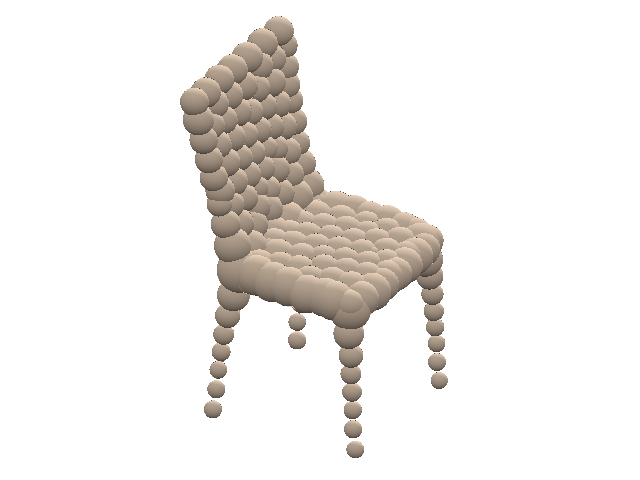}
\fbox{{\includegraphics[trim={2.9cm 0.6cm 3.3cm 0.5cm},clip,width=0.116\textwidth]{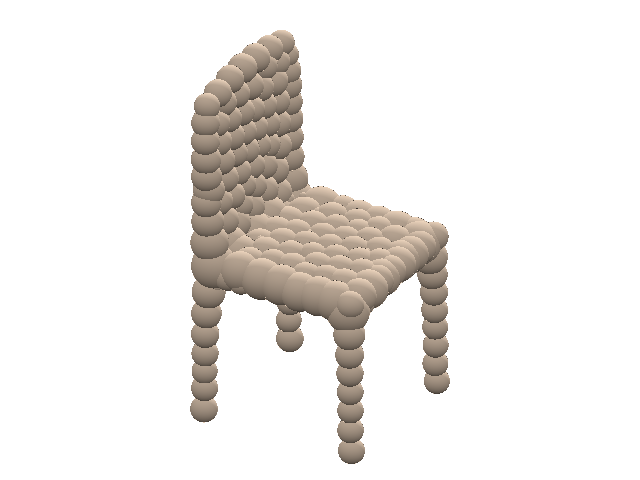}}}\\
\fbox{{\includegraphics[trim={2.9cm 0.6cm 3.3cm 0.5cm},clip,width=0.116\textwidth]{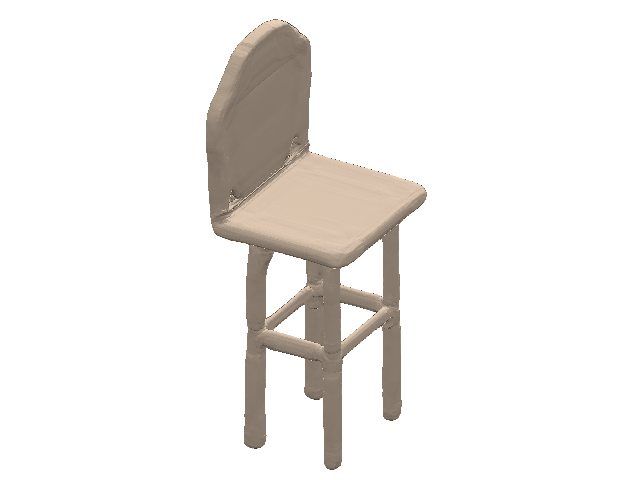}}}
\includegraphics[trim={2.9cm 0.6cm 3.3cm 0.5cm},clip,width=0.116\textwidth]{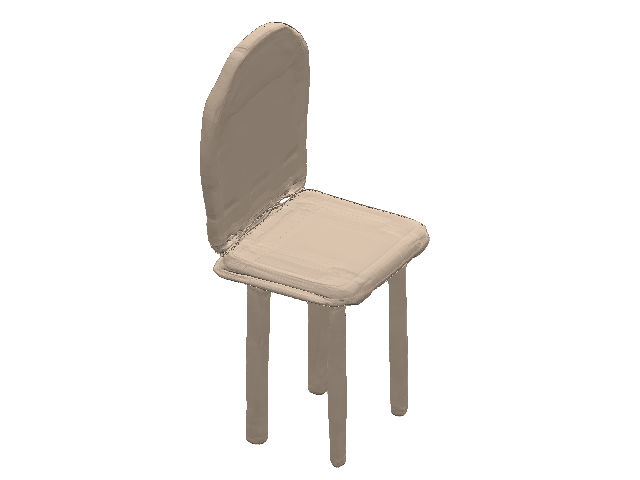}
\includegraphics[trim={2.9cm 0.6cm 3.3cm 0.5cm},clip,width=0.116\textwidth]{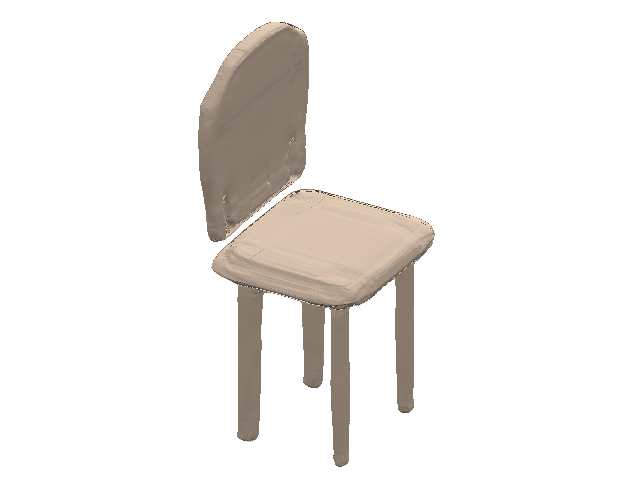}
\includegraphics[trim={2.9cm 0.6cm 3.3cm 0.5cm},clip,width=0.116\textwidth]{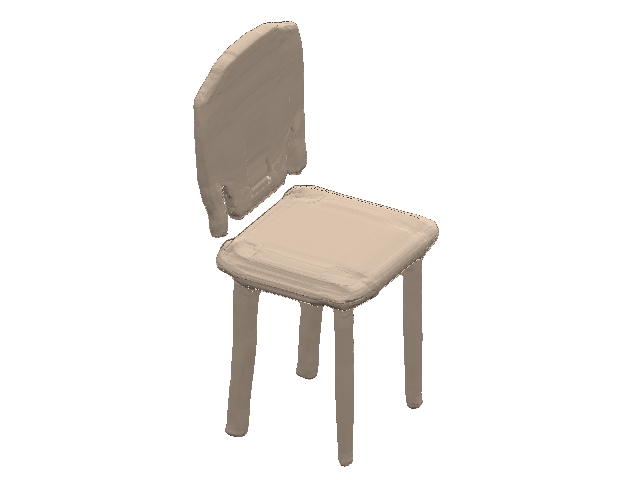}
\includegraphics[trim={2.9cm 0.6cm 3.3cm 0.5cm},clip,width=0.116\textwidth]{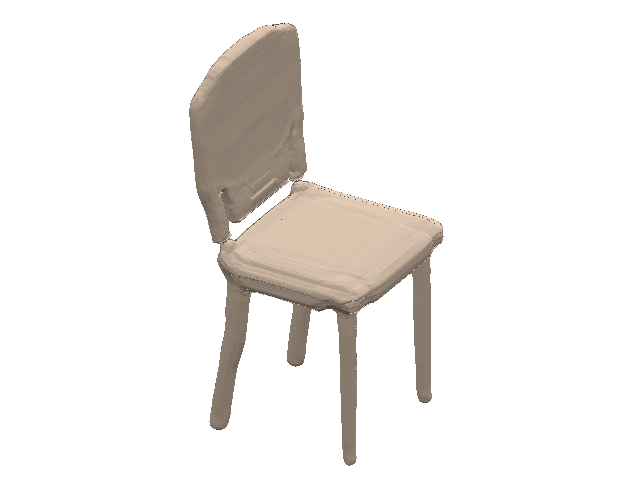}
\includegraphics[trim={2.9cm 0.6cm 3.3cm 0.5cm},clip,width=0.116\textwidth]{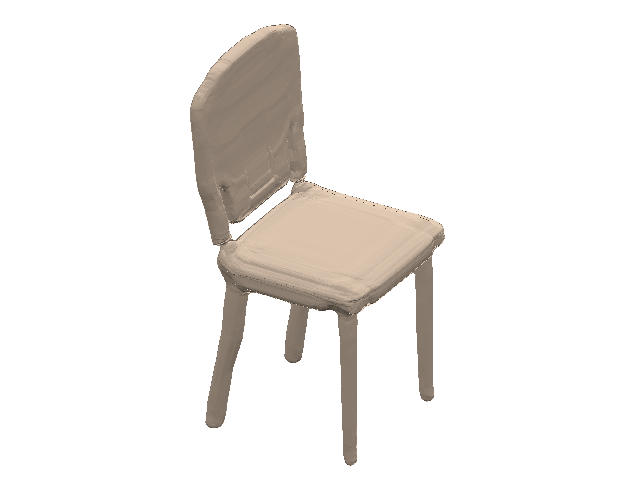}
\includegraphics[trim={2.9cm 0.6cm 3.3cm 0.5cm},clip,width=0.116\textwidth]{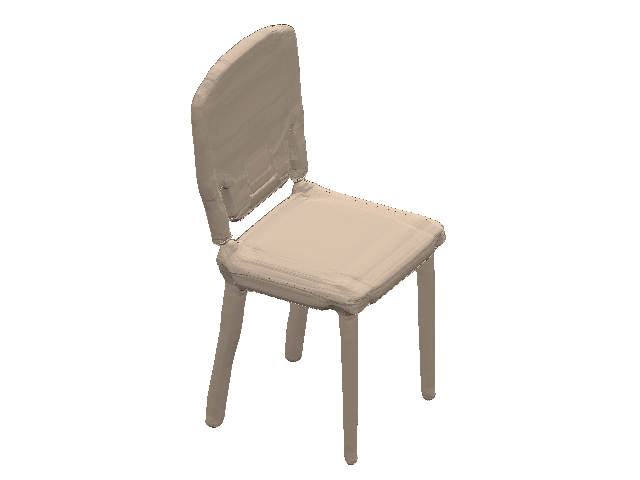}
\fbox{{\includegraphics[trim={2.9cm 0.6cm 3.3cm 0.5cm},clip,width=0.116\textwidth]{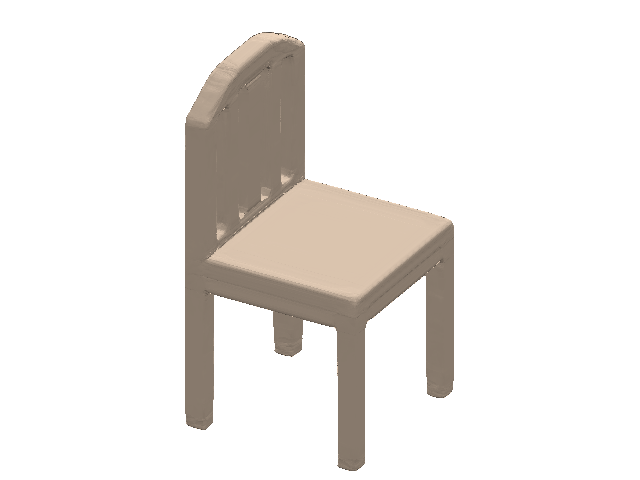}}}
  \caption{Selective primitive-space interpolation on the same pairs of shapes shown in Figure \ref{fig:interp_reg}. This type of interpolation supports selectively interpolating some of the attributes. For the first two rows, the outlines of the chairs are interpolated by gradually matching the coarse representation of the left chair to the right chair. Note how the fine details of the left chair are preserved. Similarly, for the bottom two rows, the heights of the primitives of the left chair are gradually matched to the right chair, while everything else is allowed to change freely. Many key features of the left chair are preserved in this process.
  }
      \label{fig:interp}
\end{figure*}
\section{Additional Results}

In Figure \ref{fig:manipulation}, we show additional shape manipulation results on Chair and Airplane collections. Please refer to the accompanying video for \emph{full} sequences on additional shapes. Note that while these results are obtained using simple editing operations (such as dragging a single primitive along one axis), more complicated operations on multiple primitives can be achieved in a similar manner (as we show in the main paper).

We demonstrate shape interpolation results obtained in two ways. Figure \ref{fig:interp_reg} shows results on linearly interpolating the latent code, while Figure \ref{fig:interp} illustrates a novel way of partially interpolating between two shapes by optimizing the primitive parameters, as we propose in the main paper. The latter method allows selectively interpolating certain characteristics of the shapes, such as the outline of the shape.

As illustrated in Figure \ref{fig:interp}, in the top two rows, we perform optimization on the primitive attributes to encourage the coarse representation of the left chair to match the coarse representation of the right chair. This effectively interpolates the outlines of the chairs while keeping the fine details on the left chair intact. This cannot be achieved with standard interpolation (Figure \ref{fig:interp_reg}, top two rows). 
Similarly, in the bottom two rows, we use L1 loss to encourage the heights (y-coordinates) of the primitives on the left chair to match the heights of the corresponding primitives on the right chair, while allowing other attributes to change freely during the optimization. 
Note that finding the correspondences of primitives between two shapes is trivial since each primitive generally stays at the same position across different shapes within a single class, as we illustrate in the main paper.

\end{document}